\documentclass{article} 
\usepackage{iclr2022_conference,times}

\usepackage{amsmath,amsfonts,bm}

\def\eqref#1{(\ref{#1})}

\def\1{\bm{1}}

\DeclareMathAlphabet{\mathsfit}{\encodingdefault}{\sfdefault}{m}{sl}
\SetMathAlphabet{\mathsfit}{bold}{\encodingdefault}{\sfdefault}{bx}{n}

\usepackage{hyperref}
\usepackage{url}
\usepackage{nicefrac}

\usepackage{booktabs}
\usepackage{epsfig}
\usepackage{amssymb}
\usepackage{mathrsfs}
\usepackage{mathtools}
\usepackage{resizegather}
\usepackage[title]{appendix}
\usepackage{multicol}
\usepackage{subcaption}
\usepackage{caption}
\usepackage[ruled,vlined,linesnumbered]{algorithm2e}
\usepackage{wrapfig}
\usepackage{pifont}
\usepackage{ulem}
\usepackage{wasysym}

\usepackage{amsthm}
\newtheorem{theorem}{Theorem}[section]

\newtheorem{lemma}[theorem]{Lemma}

\DeclareMathOperator*{\argsup}{arg\,sup}

\allowdisplaybreaks

\usepackage{enumitem}
\setlist{leftmargin=*,itemsep=0pt}

\newcommand{\rebut}[1]{{\color{black} #1}}

\title{Generative Modeling \\ with Optimal Transport Maps}

\author{Litu Rout \\
	Space Applications Centre\\
	Indian Space Research Organisation\\
	\texttt{lr@sac.isro.gov.in} \\
	\And
	Alexander Korotin \\
	Skolkovo Institute of Science and Technology\\
	Artificial Intelligence Research Institute (AIRI)\\
	\texttt{a.korotin@skoltech.ru}
	\And
	Evgeny Burnaev\\
	Skolkovo Institute of Science and Technology\\
	Artificial Intelligence Research Institute (AIRI)\\
	\texttt{e.burnaev@skoltech.ru}
}

\iclrfinalcopy 
\begin{document}

	\maketitle
	
	\begin{abstract}


With the discovery of Wasserstein GANs, Optimal Transport (OT) has become a powerful tool for large-scale generative modeling tasks. In these tasks, OT cost is typically used as the loss for training GANs. In contrast to this approach, we show that the OT map itself can be used as a~generative model, providing comparable performance. Previous analogous approaches consider OT maps as generative models only in the latent spaces due to their poor performance in the original high-dimensional ambient space. In contrast, we apply OT maps directly in the ambient space, e.g., a space of high-dimensional images. First, we derive a min-max optimization algorithm to efficiently compute OT maps for the quadratic cost (Wasserstein-2 distance). Next, we extend the approach to the case when the input and output distributions are located in the spaces of different dimensions and derive error bounds for the computed OT map. We evaluate the algorithm on image generation and unpaired image restoration tasks. In particular, we consider denoising, colorization, and inpainting, where the optimality of the restoration map is a desired attribute, since the output (restored) image is expected to be close to the input (degraded) one.
		
	\end{abstract}
	\section{Introduction}
	\label{intro}
	Since the discovery of Generative Adversarial Networks (GANs, \cite{goodfellow2014generative}), there has been a surge in generative modeling \citep{radford2016unsupervised,arjovsky2017wasserstein,brock2018large,karras2019style}. In the past few years, Optimal Transport (OT, \cite{villani2008optimal}) theory has been pivotal in addressing important issues of generative models. In particular, the usage of Wasserstein distance has improved diversity~\citep{arjovsky2017wasserstein,gulrajani2017improved}, convergence~\citep{sanjabi2018convergence}, and stability~\citep{miyato2018spectral,kim2021local} of GANs.
	
\begin{figure}[!t]
		\centering
		\begin{subfigure}[b]{0.57\columnwidth}
			\centering
			\includegraphics[width=0.99\linewidth]{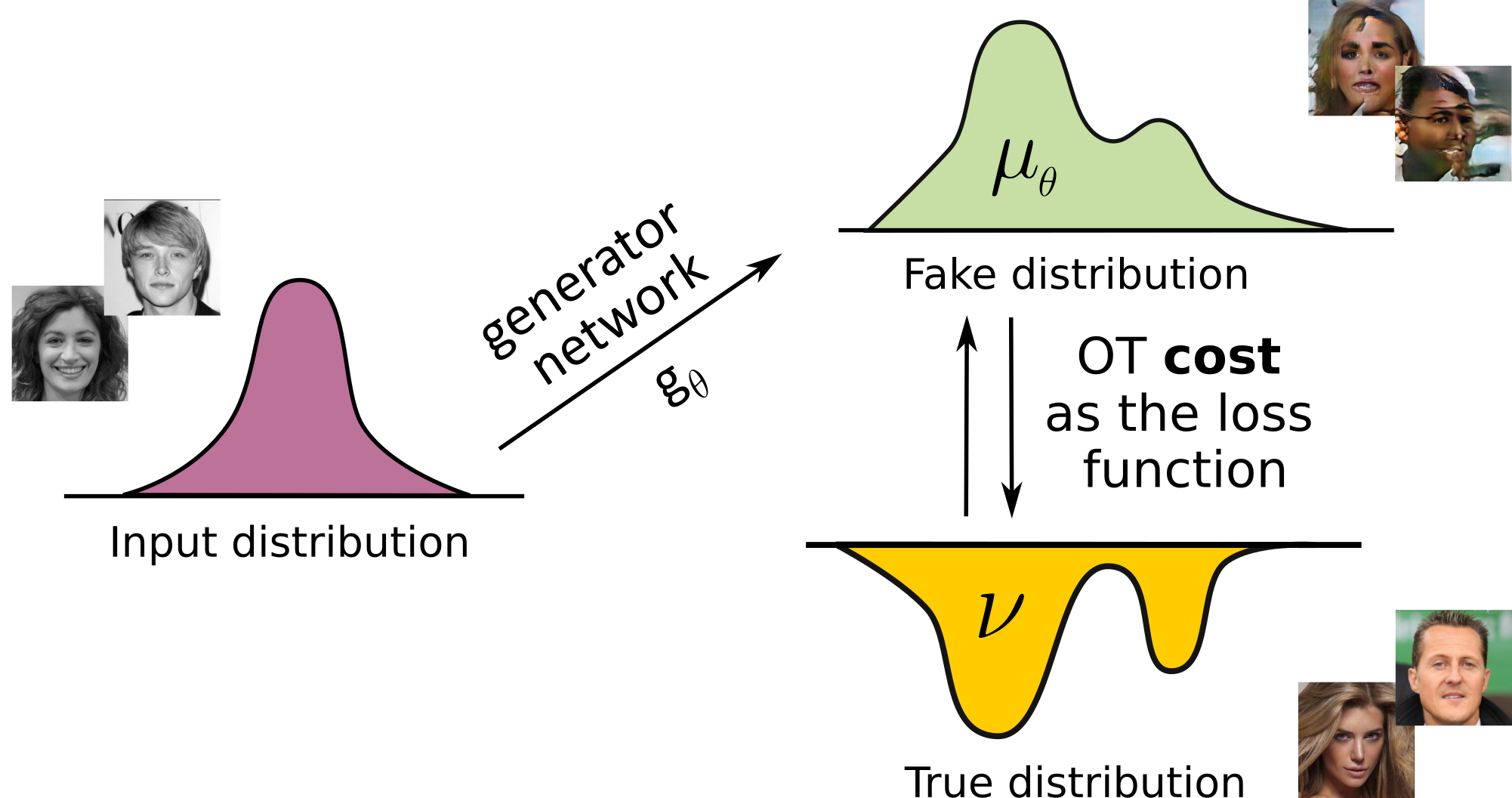}
			\caption{\centering OT cost as the loss for the generative model.}
			\label{fig:pipeline-OT-cost}
		\end{subfigure}
		\hspace{3mm}
		\vrule
		\hspace{3mm}
		\begin{subfigure}[b]{0.33\columnwidth}
			\centering
			\includegraphics[width=0.99\linewidth]{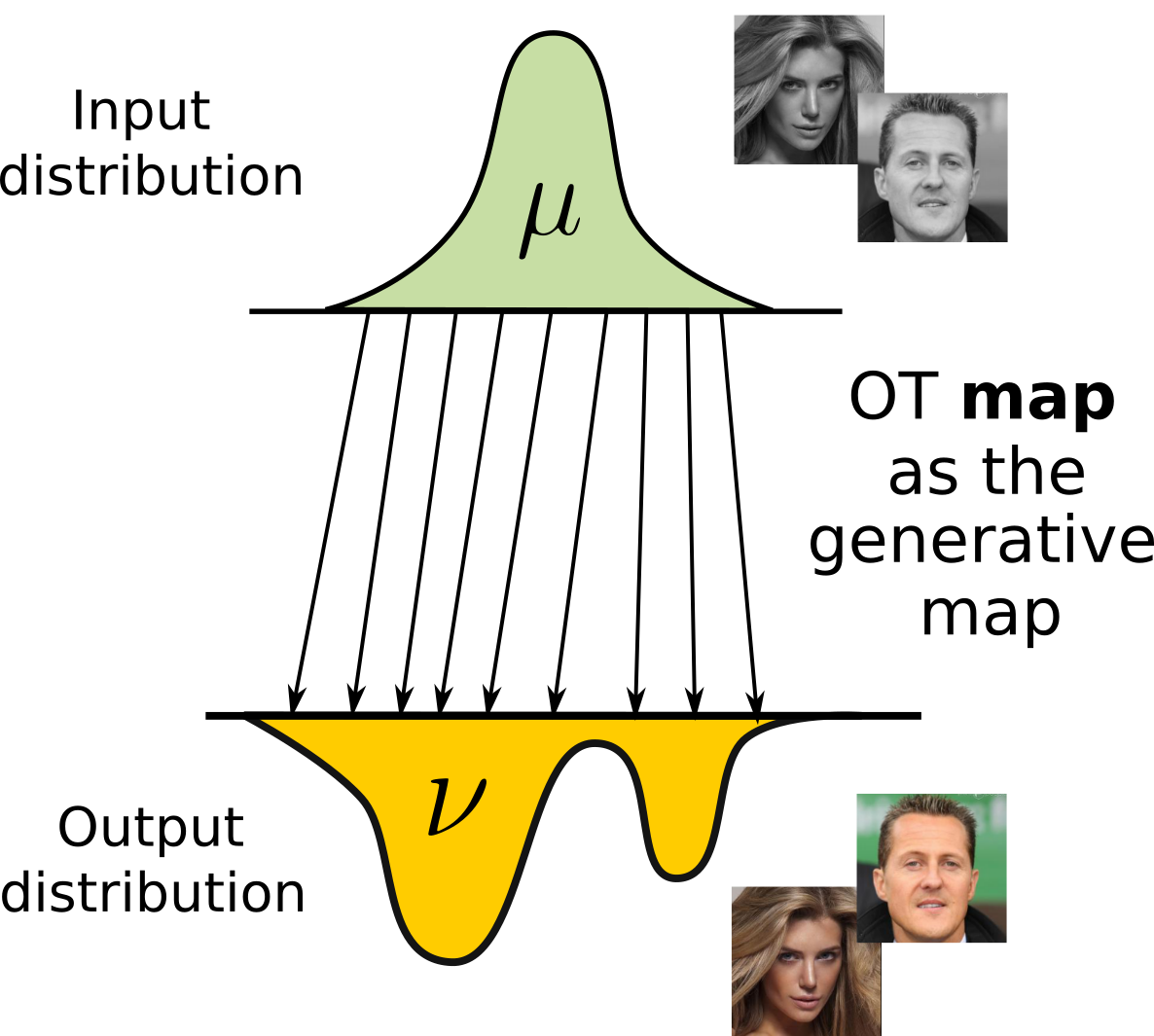}
			\caption{\centering OT map as the generative model.}
			\label{fig:pipeline-OT-map}
		\end{subfigure}
		\caption{Two existing approaches to use optimal transport in generative models.}
	\end{figure}

	Generative models based on OT can be split into two classes depending on what OT is used for. First, the \textbf{optimal transport cost serves as the loss} for generative models, see Figure \ref{fig:pipeline-OT-cost}. This is the most prevalent class of methods which includes WGAN \citep{arjovsky2017wasserstein} and its modifications: WGAN-GP \citep{gulrajani2017improved}, WGAN-LP \citep{petzka2018regularization}, and WGAN-QC \citep{Liu_2019_ICCV}. Second, the \textbf{optimal transport map is used as a~generative model} itself, see Figure \ref{fig:pipeline-OT-map}. Such approaches include LSOT \citep{seguy2018large}, AE-OT \citep{An2020AE-OT:}, ICNN-OT \citep{makkuva2020optimal}, W2GN \citep{korotin2021wasserstein}. Models of the first class have been well-studied, but limited attention has been paid to the second class. Existing approaches of the second class primarily consider \textbf{OT maps in latent spaces} of pre-trained autoencoders (AE), see Figure~\ref{fig:pipeline-OT-latent}. 
	The performance of such generative models depends on the underlying AEs, in which decoding transformations are often  not accurate; as a result this deficiency limits practical applications in high-dimensional ambient spaces.  For this reason, using OT in the latent space does not necessarily guarantee superior performance in generative modeling.
	
	The focus of our paper is the second class of OT-based models using OT map as the generative map. Finding an optimal mapping is motivated by its ability to preserve specific attributes of the input samples, a desired property in unpaired learning. For example, in unpaired image-to-image translation, the learner has to fit a map between two data distributions which preserves the image content. CycleGAN-based models \citep{CycleGAN2017} are widely used for this purpose. However, they typically have complex optimization objectives consisting of several losses \citep{amodio2019travelgan,lu2019guiding} in order to make the fitted map preserve the required attributes.

	\textbf{The main contributions of this paper are as follows:}
	\begin{enumerate}
	\item We propose an end-to-end algorithm (\wasyparagraph\ref{sec-optim}) to fit OT maps for the quadratic cost (Wasserstein-2 distance) between distributions located on the spaces of equal dimensions (\wasyparagraph\ref{sec-equal-dims}) and extend the method to unequal dimensions as well (\wasyparagraph\ref{sec-unequal}). We prove error bounds for the method (\wasyparagraph\ref{sec-error-analysis}).
		
		
		
		\item We demonstrate large-scale applications of OT maps in popular computer vision tasks. We consider image generation (\wasyparagraph\ref{sec-noise-to-data}) and unpaired image restoration (\wasyparagraph\ref{sec-enhancement}) tasks.
	\end{enumerate}
	
Our strict OT-based framework allows the theoretical analysis of the recovered transport map. The OT map obtained by our method can be directly used in large-scale computer vision problems which is in high contrast to previous related methods relying on autoencoders and OT maps in the latent space. Importantly, the performance and computational complexity of our method is comparable to OT-based generative models using OT cost as the loss.

	\textbf{Notations.} In what follows, $\mathcal{X}$ and $\mathcal{Y}$ are two complete metric spaces, $\mu(x)$ and $\nu(y)$ are probability distributions on $\mathcal{X}$ and $\mathcal{Y}$, respectively.
	For a measurable map $T:\mathcal{X}\rightarrow \mathcal{Y}$, $T_{\#}\mu$ denotes the pushforward distribution of $\mu$, i.e., the distribution for which any measurable set $E \subset \mathcal{Y}$ satisfies $T_{\#}\mu(E)=\mu(T^{-1}(E))$. For a vector $x$, $\left\| x\right\|$ denotes its Euclidean norm.
	We use $\left\langle x, y \right\rangle$ to denote the inner product of  vectors $x$ and $y$. We use $\Pi(\mu,\nu)$ to denote the set of joint probability distributions on $\mathcal{X}\times\mathcal{Y}$ whose marginals are $\mu$ and~$\nu$, respectively (couplings). For a~function $f: \mathbb{R}^D\rightarrow \mathbb{R}\cup\{\pm\infty\}$ its Legendre--Fenchel transform (the convex conjugate) is $\overline{f}(y) = \sup_{x\in\mathbb{R}^{D}}\{\langle x,y \rangle- f\left(x\right)\}$. It is convex, even if $f$ is not.

	\section{Background on Optimal Transport}
	\label{sec-prelim}
	Consider a cost of transportation, $c:\mathcal{X}\times \mathcal{Y}\rightarrow \mathbb{R}$ defined over the product space of $\mathcal{X}$ and $\mathcal{Y}$.

	\begin{wrapfigure}{r}{0.34\textwidth}\vspace{-4mm}
  \begin{center}
    \includegraphics[width=0.33\textwidth]{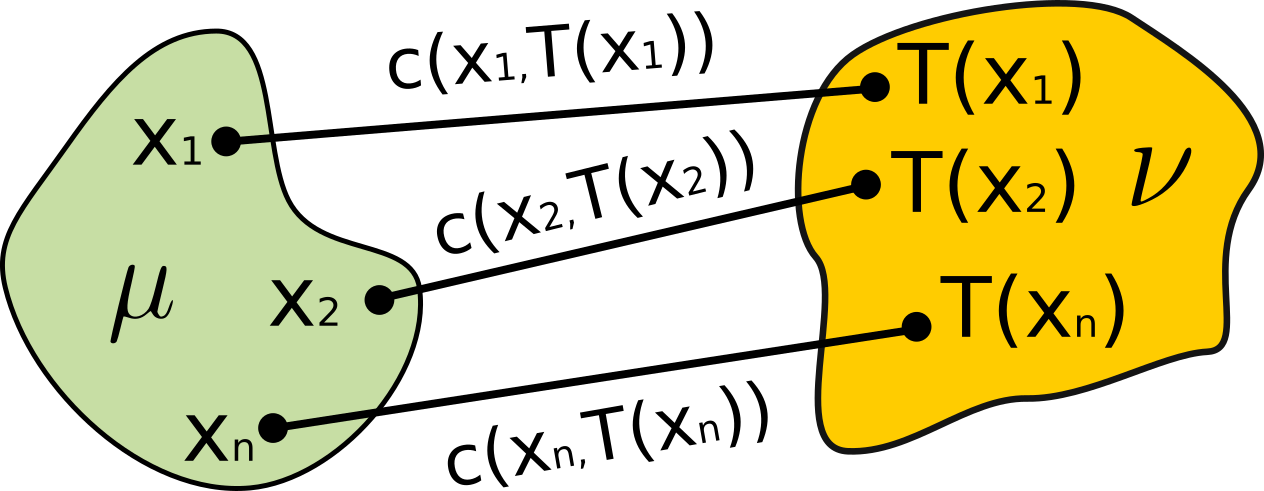}
  \end{center}
  \caption{Monge's OT.}
  \label{fig:monge}
  \vspace{-4mm}
\end{wrapfigure}\textbf{Monge's Formulation.} The \textit{optimal transport cost} between $\mu$  and $\nu$ for ground cost $c(\cdot,\cdot)$ is
		\begin{equation}
	\text{Cost}(\mu,\nu)\stackrel{\mathrm{def}}{=}\inf_{T_{\#}\mu = \nu} \int_{\mathcal{X}} c\left(x,T(x)\right) d\mu(x),
	\label{monge_eq}
	\end{equation}
	where the infimum is taken over all measurable maps ${T:\mathcal{X}\rightarrow\mathcal{Y}}$ pushing $\mu$ to $\nu$, see Figure \ref{fig:monge}. The map $T^{*}$ on which the infimum in  \eqref{monge_eq} is attained is called the \textit{optimal transport map}. Monge's formulation does not allow splitting. For example, when $\mu$ is a Dirac distribution and $\nu$ is a non-Dirac distribution, the feasible set of equation~(\ref{monge_eq}) is empty.
	
\textbf{Kantorovich's Relaxation.} Instead of asking to which particular point $y\in\mathcal{Y}$ should all the probability mass of $x$ be moved, \cite{kantorovich1948problem} asks how the mass of $x$ should be distributed among all $y\in\mathcal{Y}$. Formally, a transport coupling replaces a transport map; the OT cost is given by:
	\begin{equation}
	\text{Cost}(\mu,\nu)\stackrel{\mathrm{def}}{=}\inf_{\pi \in \Pi\left(\mu, \nu\right)} \int_{\mathcal{X}\times\mathcal{Y}} c\left(x,y\right) d\pi(x,y),
	\label{kant_eq}
	\end{equation}
	where the infimum is taken over all couplings $\pi\in\Pi(\mu,\nu)$ of $\mu$ and $\nu$.
	The coupling $\pi^{*}$ attaining the infimum of \eqref{kant_eq} is called the \textit{optimal transport plan}.	Unlike the formulation of \eqref{monge_eq}, the formulation of \eqref{kant_eq} is well-posed, and with mild assumptions on spaces $\mathcal{X},\mathcal{Y}$ and ground cost $c(\cdot,\cdot)$, the minimizer $\pi^{*}$ of \eqref{kant_eq} always exists \cite[Theorem 4.1]{villani2008optimal}. In particular, if $\pi^{*}$ is deterministic, i.e., ${\pi^{*}=[\text{id}_{\mathcal{X}},T^{*}]_{\#}\mu}$ for some $T^{*}:\mathcal{X}\rightarrow\mathcal{Y}$, then $T^{*}$ minimizes \eqref{monge_eq}.

	\textbf{Duality.} The dual form of \eqref{kant_eq} is given by \citep{kantorovich1948problem}:
	\begin{equation}
	\text{Cost}(\mu,\nu)=\sup_{\left(u,v\right)} \left\lbrace \int_{\mathcal{X}} u(x) d\mu(x) + \int_{\mathcal{Y}} v(y) d\nu(y) \colon u(x)+v(y) \leq c(x,y) \right\rbrace,
	\label{kant_dual_eq}
	\end{equation}
	with $u\in L^{1}(\mu)$, $v\in L^{1}(\nu)$ called Kantorovich \textit{potentials}. For $u:\mathcal{X}\rightarrow \mathbb{R}$ and $v: \mathcal{Y}\rightarrow \mathbb{R}$ define their $c$-transforms by ${u^c({\color{black}{y}}) = \inf_{x\in \mathcal{X}}\{c\left(x,y\right) - u\left(x\right)\}}$ and ${v^c(x) = \inf_{y\in \mathcal{Y}}\{c\left(x,y\right) - v\left(y\right)\}}$ respectively. 
	Using $c$-transform, \eqref{kant_dual_eq} is reformulated as \cite[\wasyparagraph 5]{villani2008optimal}
	\begin{equation}
	\text{Cost}(\mu,\nu)\!=\!\sup_v\!\left\lbrace \!\int_{\mathcal{X}}v^c(x) d\mu(x)\! +\! \int_{\mathcal{Y}} v(y) d\nu(y)\!\right\rbrace\!=\!\sup_u\! \left\lbrace \!\int_{\mathcal{X}}u(x) d\mu(x) \!+\! \int_{\mathcal{Y}} u^{c}(y) d\nu(y)\!\right\rbrace\!.
	\label{kant_dual_c_eq}
	\end{equation}
	
	\textbf{Primal-dual relationship.} For certain ground costs $c(\cdot,\cdot)$, the primal solution $T^{*}$ of \eqref{monge_eq} can be recovered from the dual solution $u^{*}$ of \eqref{kant_dual_eq}. For example, if $\mathcal{X}=\mathcal{Y}=\mathbb{R}^{D}$, $c(x,y)=h(x-y)$ with strictly convex $h:\mathbb{R}^{D}\rightarrow\mathbb{R}$ and $\mu$ is absolutely continuous supported on the compact set, then \begin{equation}T^{*}(x)=x-(\nabla h)^{-1}\big(\rebut{\nabla} u^{*}(x)\big),
	\label{primal-dual-relationship}
	\end{equation}
	see \cite[Theorem 1.17]{santambrogio2015optimal}. For general costs, see \cite[Theorem 10.28]{villani2008optimal}.

	\section{Optimal Transport in Generative Models}
	\label{rel_work}

	\textbf{\textsc{Optimal transport cost as the loss}} (Figure \ref{fig:pipeline-OT-cost}). Starting with the works of \cite{arjovsky2017towards,arjovsky2017wasserstein}, the usage of OT cost as the loss has become a~major way to apply OT for generative modeling. In this setting, given data distribution $\nu$ and fake distribution $\mu_{\theta}$, the goal is to minimize $\text{Cost}(\mu_{\theta},\nu)$ w.r.t.\ the parameters~$\theta$. Typically, $\mu_{\theta}$ is a pushforward distribution of some given distribution, e.g., $\mathcal{N}(0,I)$, via generator network $G_{\theta}$.
	
	The \textit{Wasserstein-1} distance ($\mathcal{W}_{1}$), i.e., the transport cost for ground cost $c(x,y)=\|x-y\|$, is the most practically prevalent example of such a loss. Models based on this loss are known as Wasserstein GANs (WGANs). They estimate $\mathcal{W}_{1}(\mu_{\theta},\nu)$ based on the dual form as given by \eqref{kant_dual_c_eq}. For $\mathcal{W}_{1}$, the optimal potentials $u^{*},v^{*}$ of \eqref{kant_dual_c_eq} satisfy $u^{*}=-v^{*}$ where $u^{*}$ is a $1$-Lipschitz function \cite[Case 5.16]{villani2008optimal}. As a result, to compute $\mathcal{W}_{1}$, one needs to optimize the following simplified form:
	\begin{equation}
	\mathcal{W}_{1}(\mu_{\theta},\nu)=\sup_{\|u\|_{L}\leq 1} \left\lbrace \int_{\mathcal{X}} u(x) d\mu_{\theta}(x) - \int_{\mathcal{Y}} u(y) d\nu(y) \right\rbrace.
	\label{kant_dual_c1_eq}
	\end{equation}
	In WGANs, the~potential $u$ is called the \textit{discriminator}. Optimization of \eqref{kant_dual_c1_eq} reduces constrained optimization of \eqref{kant_dual_c_eq} with two potentials $u,v$ to optimization of only one discriminator $u$. In practice, enforcing the Lipschitz constraint on $u$ is challenging. Most methods to do this are regularization-based, e.g., they use gradient penalty \cite[WGAN-GP]{gulrajani2017improved} and Lipschitz penalty \cite[WGAN-LP]{petzka2018regularization}. Other methods enforce Lipschitz property via incorporating certain hard restrictions on the discriminator's architecture \citep{anil2019sorting,tanielian2021approximating}.
	
	\textit{General transport costs} (other than $\mathcal{W}_{1}$) can also be used as the loss for generative models. They are less popular since they do not have a dual form reducing to a single potential function similar to \eqref{kant_dual_c1_eq} for $\mathcal{W}_{1}$. Consequently, the challenging estimation of the $c$-transform $u^{c}$ is needed. To avoid this, \cite{sanjabi2018convergence} consider the dual form of \eqref{kant_dual_eq} with two potentials $u,v$ instead form \eqref{kant_dual_c_eq} with one~$u$ and softly enforce the condition $u(x)+v(y)\leq c(x,y)$ via entropy or quadratic regularization. \cite{nhan2019threeplayer} use the dual form of \eqref{kant_dual_c_eq} and amortized optimization to compute $u^{c}$ via 
	an additional neural network. Both methods work for general $c(\cdot,\cdot)$, though the authors test them for $c(x,y)=\|x-y\|$ only, i.e., $\mathcal{W}_{1}$ distance. \cite{mallasto2019q} propose a fast way to approximate the $c$-transform and test the approach (WGAN-$(q,p)$) with several costs, in particular, the \textit{Wasserstein-2} distance ($\mathcal{W}_{2}$), i.e., the transport cost for the quadratic ground cost $c(x,y)=\frac{1}{2}\|x-y\|^{2}$. Specifically for $\mathcal{W}_{2}$, \cite{Liu_2019_ICCV} approximate the $c$-transform via a linear program (WGAN-QC).
	
	A fruitful branch of OT-based losses for generative models comes from modified versions of OT cost, such as Sinkhorn \citep{genevay2018learning}, sliced \citep{deshpande2018generative} and minibatch \citep{fatras2019learning} OT distances. They typically have lower sample complexity than usual OT and can be accurately estimated from random mini-batches without using dual forms such as \eqref{kant_dual_eq}. In practice, these approaches usually learn the ground OT cost $c(\cdot,\cdot)$. 
	
    The aforementioned methods use OT cost in the ambient space to train GANs. There also exist approaches using OT cost in the latent space. For example, \cite{tolstikhin2017wasserstein,patrini2020sinkhorn} use OT cost between encoded data and a given distribution as an additional term to reconstruction loss for training an AE. As the result, AE's latent distribution becomes close to the given one.

	
	\textbf{\textsc{Optimal transport map as the generative map}} (Figure \ref{fig:pipeline-OT-map}). Methods to compute the OT map (plan) are less common in comparison to those computing the cost. Recovering the map from the primal form \eqref{monge_eq} or \eqref{kant_eq} usually yields complex optimization objectives containing several adversarial terms \citep{xie2019scalable,liu2021learning,lu2020large}. Such procedures require careful hyperparameter choice. This needs to be addressed before using these methods in practice.
	
	Primal-dual relationship (\wasyparagraph\ref{sec-prelim}) makes it possible to recover the OT map via solving the dual form~\eqref{kant_dual_eq}. Dual-form based methods primarily consider $\mathcal{W}_{2}$ cost due to its nice theoretical properties and relation to convex functions \citep{brenier1991polar}. In the semi-discrete case ($\mu$ is continuous, $\nu$ is discrete), \cite{An2020AE-OT:} and \cite{lei2019geometric} compute the dual potential and the OT map by using the Alexandr{\color{black}ov} theory and
	convex geometry. For the continuous case, \cite{seguy2018large} use the entropy (quadratic) regularization to recover the dual potentials and extract OT map from them via the barycenteric projection. \cite{taghvaei20192}, \cite{makkuva2020optimal}, \cite{korotin2021wasserstein} employ input-convex neural networks (ICNNs, see \cite{amos2017input}) to parametrize potentials in the dual problem and recover OT maps by using their gradients.
	\begin{figure}[!h]
	\vspace{-2mm}
		\centering
		\includegraphics[width=0.6\linewidth]{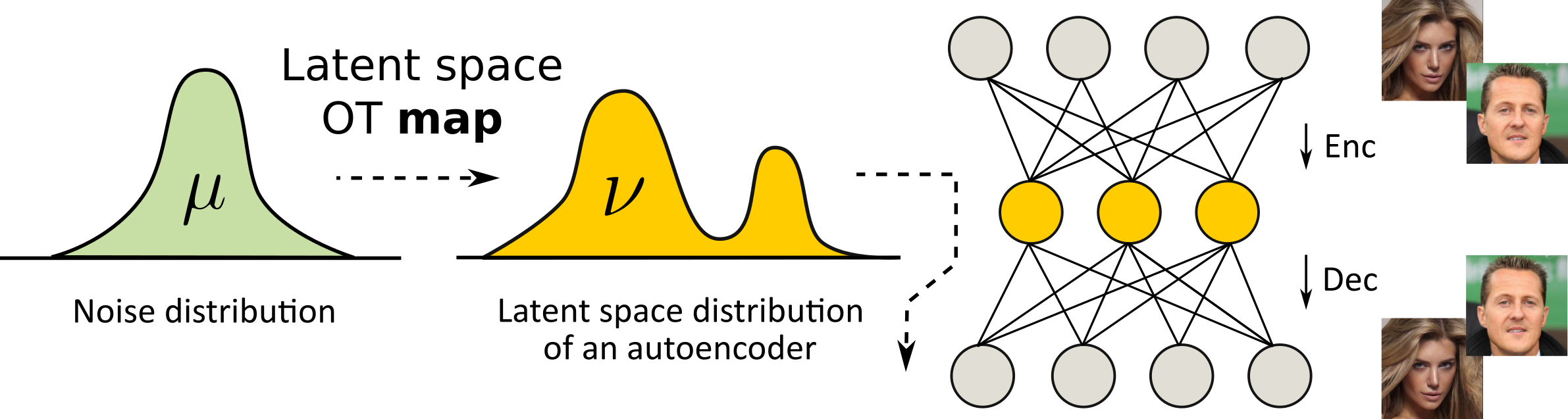}
	\vspace{-2mm}
		\caption{\centering The existing most prevalent approach to use OT maps in generative models.}
		\label{fig:pipeline-OT-latent}
				\vspace{-4mm}
	\end{figure}
	
	The aforementioned dual form methods compute OT maps in \textbf{\textsc{latent spaces}} for problems such as domain adaptation and latent space mass transport, see Figure \ref{fig:pipeline-OT-latent}. OT maps in high-dimensional ambient spaces, e.g., natural images, are usually not considered. Recent evaluation of continuous OT methods for $\mathcal{W}_{2}$ \citep{korotin2021neural} reveals their crucial limitations, which negatively affect their scalability, such as poor expressiveness of ICNN architectures or bias due to regularization.

	\section{End-to-end Solution to Learn Optimal Maps}
	\label{tdp}
	
	\subsection{Equal Dimensions of Input and Output Distributions}
	\label{sec-equal-dims}
	
	In this section, we use $\mathcal{X}=\mathcal{Y}=\mathbb{R}^{D}$ and consider the Wasserstein-2 distance ($\mathcal{W}_{2}$), i.e., the optimal transport for the quadratic ground cost $c(x,y)=\frac{1}{2}\|x-y\|^{2}$. {\color{black}We use the dual form \eqref{kant_dual_c_eq} to derive a saddle point problem the solution of which yields the OT map $T^{*}$.} We consider distributions $\mu,\nu$ with finite second moments. We assume that for distributions $\mu,\nu$ in view there \textit{exists} a \textit{unique} OT plan $\pi^{*}$ minimizing \eqref{kant_dual_eq} and it is deterministic, i.e., $\pi^{*}=[\text{id}_{\mathbb{R}^{D}},T^{*}]_{\#}\mu$. Here $T^{*}$ is an OT map which minimizes \eqref{monge_eq}. Previous related works \citep{makkuva2020optimal,korotin2021wasserstein} assumed the absolute continuity of~$\mu$, which implied the existence and uniqueness of $T^{*}$ \citep{brenier1991polar}.
	
	Let $\psi(y)\stackrel{\mathrm{def}}{=}\frac{1}{2}\|y\|^{2}-v(y)$,  where $v$ is the potential of \eqref{kant_dual_c_eq}.
	Note that
	\begin{equation}v^{c}(x)=\inf_{y\in\mathbb{R}^{D}}\left\lbrace \frac{1}{2}\|x-y\|^{2}-v(y)\right\rbrace=\frac{1}{2}\|x\|^{2}-\sup_{y\in\mathbb{R}^{D}}\left\lbrace\langle x,y\rangle-\psi(y)\right\rbrace=\frac{1}{2}\|x\|^{2}-\overline{\psi}(x).
	\end{equation}
	Therefore, \eqref{kant_dual_c_eq} is equivalent to
	\begin{eqnarray}\mathcal{W}_{2}^{2}(\mu,\nu)=\int_{\mathcal{X}}\frac{\|x\|^{2}}{2}d\mu(x)+\int_{\mathcal{Y}}\frac{\|y\|^{2}}{2}d\nu(x)+
	\sup_\psi\left\lbrace -\int_{\mathcal{X}}\overline{\psi}(x) d\mu(x) - \int_{\mathcal{Y}} \psi(y) d\nu(y)\right\rbrace=
	\label{opt-double-psi}
	\\
	\text{Const\color{black}ant}(\mu,\nu)-\inf_\psi\left\lbrace\int_{\mathcal{X}}\overline{\psi}(x) d\mu(x) + \int_{\mathcal{Y}} \psi(y) d\nu(y)\right\rbrace=
	\label{opt-double-psi-inf}
	\\
	\text{Const\color{black}ant}(\mu,\nu)-\inf_\psi\left\lbrace \int_{\mathcal{X}}\sup_{y\in\mathbb{R}^{D}}\left\lbrace\langle x,y\rangle-\psi(y)\right\rbrace d\mu(x)+\int_{\mathcal{Y}} \psi(y) d\nu(y)\right\rbrace=
	\label{opt-x}
	\\
	\text{Const\color{black}ant}(\mu,\nu)-\inf_\psi\left\lbrace \sup_{T}\int_{\mathcal{X}}\left\lbrace\langle x,T(x)\rangle-\psi\big(T(x)\big)\right\rbrace d\mu(x)+\int_{\mathcal{Y}} \psi(y) d\nu(y)\right\rbrace
	\label{opt-t}
	\end{eqnarray}
	where between lines \eqref{opt-x} and \eqref{opt-t} we replace the optimization over $y\in\mathbb{R}^{D}$ with the equivalent optimization over functions $T:\mathbb{R}^{D}\rightarrow\mathbb{R}^{D}$. {\color{black}The equivalence follows from the interchange between the integral and the supremum \cite[Theorem 3A]{rockafellar1976integral}. We also provide an independent proof of equivalence specializing Rockafellar's interchange theorem in Appendix~\ref{prf_equiv}. Thanks to the following lemma, we may solve saddle point problem \eqref{opt-t} and obtain the OT map $T^{*}$ from its solution $(\psi^{*},T^{*})$. }
	\begin{lemma}
		Let $T^{*}$ be the OT map from $\mu$ to $\nu$. Then, for every optimal potential $\psi^{*}$, 
		\begin{equation}T^{*}\in\argsup_{T}\int_{\mathcal{X}}\left\lbrace\langle x,T(x)\rangle-\psi^{*}\big(T(x)\big)\right\rbrace d\mu(x).\label{t-in-argsup}\end{equation}
		\label{lemma-eq}
	\end{lemma}
	We prove Lemma \ref{lemma-eq} in Appendix \ref{prf_lm41}. For general $\mu,\nu$ the $\argsup_{T}$ set for optimal $\psi^{*}$ might contain not only OT map $T^{*}$, but other functions as well. Working with real-world data in experiments (\wasyparagraph\ref{sec-enhancement}), we observe that despite this issue, optimization \eqref{opt-t} still recovers $T^{*}$.
	
	\textbf{Relation to previous works.} The use of  the function ${T}$ to approximate the $c$-transform was proposed by \cite{nhan2019threeplayer} to estimate the Wasserstein loss in WGANs. For $\mathcal{W}_{2}$, the fact that $T^{*}$ is an OT map was used by \cite{makkuva2020optimal,korotin2021wasserstein} who primarily assumed continuous $\mu,\nu$ and reduced \eqref{opt-t} to convex $\psi$ and $T=\nabla\phi$ for convex $\phi$. Issues with non-uniqueness of solution of \eqref{t-in-argsup} were softened, but using ICNNs to parametrize $\psi$ became necessary.
	
	\cite{korotin2021neural} demonstrated that ICNNs negatively affect practical performance of OT and tested an unconstrained formulation similar to \eqref{opt-t}. As per the evaluation, it provided the best empirical performance  \cite[\wasyparagraph 4.5]{korotin2021neural}. The method $\lfloor \text{MM:R}\rceil$ they consider parametrizes ${\frac{1}{2}\|\cdot\|^{2}-\psi(\cdot)}$ by a neural network, while we directly parametrize $\psi(\cdot)$ by a neural network (\wasyparagraph\ref{sec-optim}).
	
	Recent work by \cite{fan2021scalable} exploits formulation similar to \eqref{opt-t} for general costs $c(\cdot,\cdot)$. While their formulation leads to a max-min scheme with general costs \cite[Theorem 3]{fan2021scalable}, our approach gives rise to a min-max method for quadratic cost. In particular, we extend the formulation to learn OT maps between distributions in spaces with unequal dimensions, see the next subsection.

	\subsection{Unequal Dimensions of Input and Output Distributions}
	\label{sec-unequal}
	
	Consider the case when $\mathcal{X}=\mathbb{R}^{H}$ and $\mathcal{Y}=\mathbb{R}^{D}$ have different dimensions, i.e., $H\neq D$. In order to map the probability distribution $\mu$ to $\nu$, a straightforward solution is to embed $\mathcal{X}$ to $\mathcal{Y}$ via \textit{some} $Q:\mathcal{X}\rightarrow\mathcal{Y}$ and then to fit the OT map between $Q_{\#}\mu$ and $\nu$ for the quadratic cost on $\mathcal{Y}=\mathbb{R}^{D}$. In this case, the optimization objective becomes
	\begin{equation}
	\inf_\psi\sup_{T}\left\lbrace \int_{\mathcal{X}}\left\lbrace\langle Q(x),T\big(Q(x)\big)\rangle-\psi\big(T\big(Q(x)\big)\big)\right\rbrace d\mu(x)+\int_{\mathcal{Y}} \psi(y) d\nu(y)\right\rbrace
	\label{unequal-dim-T}
	\end{equation}
	with the optimal $T^{*}$ recovering the OT map from $Q_{\#}\mu$ to $\nu$. For equal dimensions $H=D$ and the identity embedding $Q(x)\equiv x$, expression \eqref{unequal-dim-T} reduces to optimization \eqref{opt-t} up to a constant.
	
	Instead of optimizing  \eqref{unequal-dim-T} over functions $T:Q(\mathcal{X})\rightarrow\mathcal{Y}$, we propose to consider optimization directly over generative mappings $G:\mathcal{X}\rightarrow \mathcal{Y}$:
	\begin{equation}
	\mathcal{L}(\psi,G)\stackrel{\mathrm{def}}{=}\inf_\psi\sup_{G}\left\lbrace \int_{\mathcal{X}}\left\lbrace\langle Q(x),G(x)\rangle-\psi\big(G(x)\big)\right\rbrace d\mu(x)+\int_{\mathcal{Y}} \psi(y) d\nu(y)\right\rbrace
	\label{unequal-dim-G}
	\end{equation}
	Our following lemma establishes connections between \eqref{unequal-dim-G} and OT with unequal dimensions:
	\begin{lemma}
		Assume that exists a unique OT plan between $Q_{\#}\mu$ and $\nu$ and it is deterministic, i.e., $[\textrm{id}_{\mathbb{R}^{D}},T^{*}]_{\#}(Q_{\#}\mu)$. Then $G^{*}(x)=T^{*}\big(Q(x)\big)$ is the OT map between $\mu$ and $\nu$ for the $Q$-\textbf{embedded quadratic cost} ${c(x,y)=\frac{1}{2}\|Q(x)-y\|^{2}}$. Moreover, for every optimal potential $\psi^{*}$ of problem~\eqref{unequal-dim-G},
		\begin{equation}G^{*}\in\argsup_{G}\int_{\mathcal{X}}\left\lbrace\langle Q(x),G(x)\rangle-\psi^{*}\big(G(x)\big)\right\rbrace d\mu(x).
		\label{g-in-argsup}
		\end{equation}
		\label{lemma-uneq}\end{lemma}
		
	\vspace{-2mm}\begin{wrapfigure}{r}{0.52\textwidth}
	\vspace{-5mm}
  \begin{center}
    \includegraphics[width=0.51\textwidth]{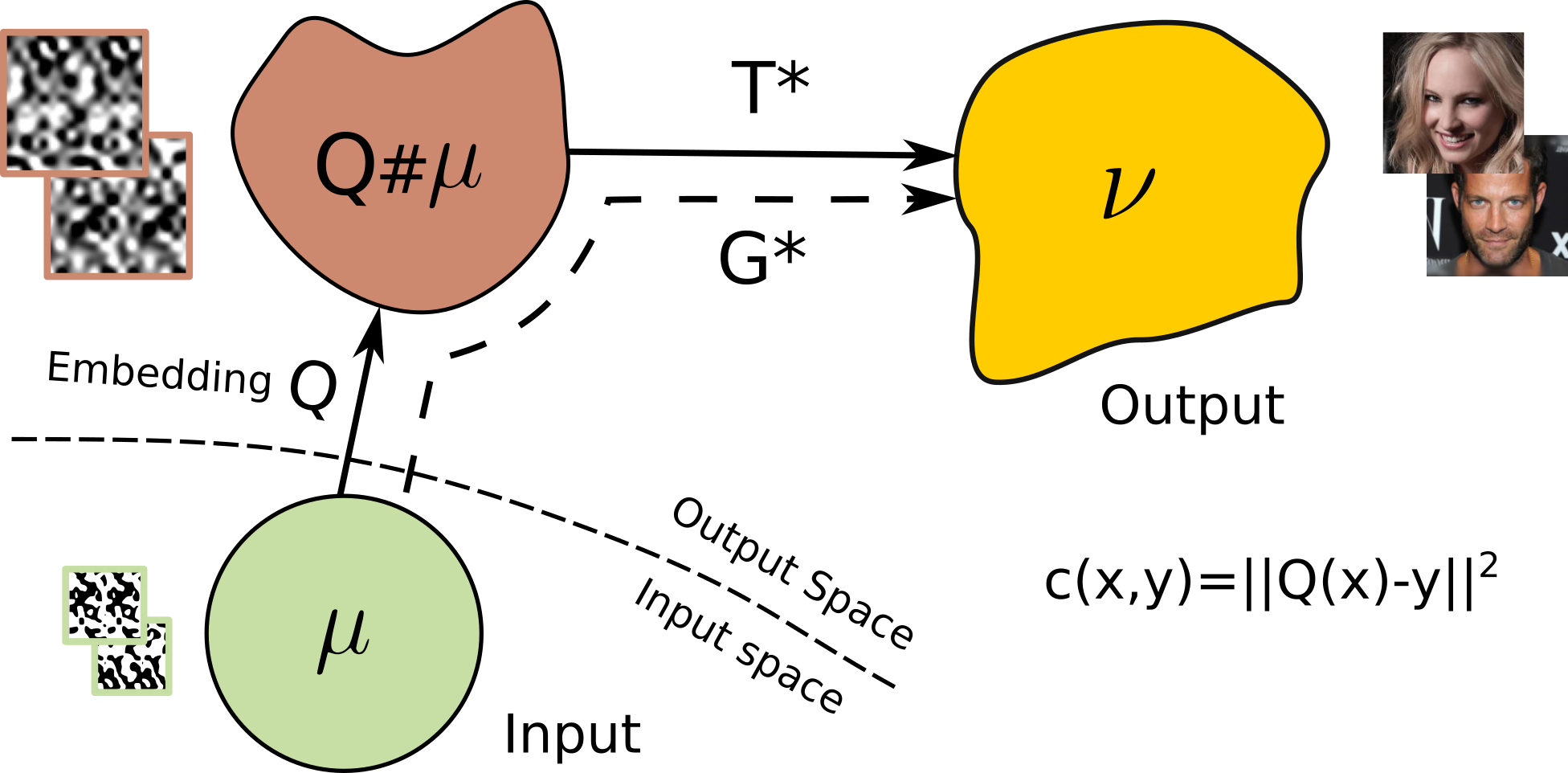}
  \end{center}
  \caption{\color{black}The scheme of our approach for learning OT maps between unequal dimensions. In the figure, the setup of \wasyparagraph\ref{sec-noise-to-data} is shown: $\mu$ is a noise, $Q$ is the bicubic upscaling, $\nu$ is a distribution of images.}
  \label{fig:pipe-q}
  \vspace{-2mm}
\end{wrapfigure}We prove Lemma \ref{lemma-uneq} in Appendix \ref{prf_lm42} {\color{black}and schematically present its idea in Figure \ref{fig:pipe-q}}. Analogously to Lemma \ref{lemma-eq}, it provides a way to compute the OT map $G^{*}$ for the $Q$-embedded quadratic cost between distributions $\mu$ and $\nu$ by  solving the saddle point problem \eqref{unequal-dim-G}. Note the situation with non-uniqueness of $\argsup_{G}$ is similar to ~\wasyparagraph\ref{sec-equal-dims}.
	
	\textbf{Relation to previous works.} In practice, learning OT maps directly between spaces of unequal dimensions was considered in the work by \cite[\wasyparagraph 5.2]{fan2021scalable} but only on toy examples. We demonstrate that our method works well in large-scale generative modeling tasks (\wasyparagraph\ref{sec-noise-to-data}). Theoretical properties of OT maps for embedded costs are studied, e.g., in \citep{pass2010regularity,mccann2020optimal}.

	\subsection{Practical Aspects and Optimization Procedure}
	\label{sec-optim}
	To optimize functional \eqref{unequal-dim-G}, we approximate ${G:\mathbb{R}^{H}\rightarrow \mathbb{R}^{D}}$ and ${\psi:\mathbb{R}^{D}\rightarrow\mathbb{R}}$ with neural networks $G_{\theta},\psi_{\omega}$ and optimize their parameters via stochastic gradient descent-ascent (SGDA) by using mini-batches from $\mu,\nu$. The practical optimization procedure is given in Algorithm \ref{algo1} below. Following the usual practice in GANs, we add a small penalty (\wasyparagraph \ref{sec-dis-gp}) on potential $\psi_{\omega}$ for better stability. The penalty is not included in Algorithm \ref{algo1} to keep it simple.

		\textbf{Relation to previous works.} WGAN by \cite{arjovsky2017towards} uses $\mathcal{W}_{1}$ as the loss to update the generator while we solve a diferent task\,---\,we fit the generator~$G$ to be the OT map for $Q$-embedded quadratic cost. Despite this, our Algorithm \ref{algo1} has similarities with WGAN's training. The update of $\psi$ (line 4) coincides with discriminator's update in WGAN. The update of generator $G$ (line 8) differs from WGAN's update by the term $-\langle Q(\cdot),G_{\theta}(\cdot)\rangle$. Besides, in WGAN the optimization is $\inf_{G}\sup_{D}$. We have $\inf_{\psi}\sup_{G}$, i.e., the generator in our case is the solution of the inner problem.
	\begin{algorithm}[!t]
		\SetAlgorithmName{Algorithm}{empty}{Empty}
		\SetKwInOut{Input}{Input}
		\SetKwInOut{Output}{Output}
		\Input{Input distribution $\mathbb{\mu}$ on $\mathcal{X}=\mathbb{R}^{H}$; output distribution $\nu$ on $\mathcal{Y}=\mathbb{R}^{D}$;\\
			generator network $G_{\theta}:\mathbb{R}^{H}\rightarrow\mathbb{R}^{D}$; potential network $\psi_{\omega}:\mathbb{R}^{D}\rightarrow\mathbb{R}$\;
			number of iterations per network: $K_{G}$, $K_{\psi}$; embedding $Q:\mathcal{X}\rightarrow\mathcal{Y}$;
		}
		\Output{Trained generator $G_{\theta}$ representing OT map from $\mu$ to $\nu$;}
		\Repeat{not converged}{
			\For{$k_{\psi}= 1$ to $K_{\psi}$}{
				Draw batch $X\sim \mu$ and $Y\sim \nu$\;
				$\mathcal{L}_{\psi}\leftarrow\frac{1}{|Y|}\sum_{y\in Y}\psi_{\omega}(y)-\frac{1}{|X|}\sum_{x\in X}\psi_{\omega}\big(G_{\theta}(x)\big)$\;
				Update $\omega$ by using $\frac{\partial \mathcal{L}_{\psi}}{\partial \omega}$ to minimize $\mathcal{L}_{\psi}$\;
			}
			\For{$k_{G} = 1$ to $K_{G}$}{
				Draw batch $X\sim \mu$\;
				$\mathcal{L}_{G}\leftarrow \frac{1}{|X|}\sum_{x\in X}\big[\psi\big(G(x)\big)-\langle Q(x),G_{\theta}(x)\rangle\big]$\;
				Update $\theta$ by using $\frac{\partial \mathcal{L}_{G}}{\partial \theta}$ to minimize $\mathcal{L}_{G}$\;
			}
		}
		\Return{$G_{\theta}$}
		\caption{Learning the optimal transport map between unequal dimensions.}
		\label{algo1}
	\end{algorithm}

	\subsection{Error Analysis}
	\label{sec-error-analysis}
	Given a pair $(\hat{\psi},\hat{G})$ approximately solving \eqref{unequal-dim-G}, a natural question to ask is how good is the recovered OT map $\hat{G}$. In this subsection, we provide a bound on the difference between $G^{*}$ and $\hat{G}$ based on the duality gaps for solving outer and inner optimization problems.
	
	In \eqref{opt-double-psi}, and, as the result, in \eqref{opt-x}, \eqref{opt-t}, \eqref{unequal-dim-T}, \eqref{unequal-dim-G}, it is enough to consider optimization over \textit{convex} functions $\psi$, see \cite[Case 5.17]{villani2008optimal}. Our theorem below assumes the convexity of $\hat{\psi}$ although it might not hold in practice since in practice $\hat{\psi}$ is a neural network.
	
	\begin{theorem}
	Assume that there exists a unique deterministic OT plan for $Q$-embedded quadratic cost between $\mu$ and $\nu$, i.e., $\pi^{*}=[\textrm{id}_{\mathbb{R}^{H}},G^{*}]_{\#}\mu$ for $G^{*}:\mathbb{R}^{H}\rightarrow\mathbb{R}^{D}$. Assume that $\hat{\psi}$ is $\beta$-strongly convex ($\beta>0$) and $\hat{G}:\mathbb{R}^{H}\rightarrow \mathbb{R}^{D}$. Define
	$$\epsilon_{1}=\sup_{G}\mathcal{L}(\hat{\psi},G)-\mathcal{L}(\hat{\psi},\hat{G})\qquad\text{and}\qquad \epsilon_{2}=\sup_{G}\mathcal{L}(\hat{\psi},G)-\inf_{\psi}\sup_{G}\mathcal{L}(\psi,G)$$
	Then the following bound holds true for the OT map $G^{*}$ from $\mu$ to $\nu$:
	\begin{equation}
 \frac{\text{\normalfont FID}(\hat{G}_{\#}\mu,\nu)}{L^{2}}\leq 2\cdot\mathcal{W}_{2}^{2}(\hat{G}_{\#}\mu,\nu)\leq\int_{\mathcal{X}} \|\hat{G}(x)-G^{*}(x)\|^{2}d\mu(x)\leq \frac{2}{\beta}(\sqrt{\epsilon_1}+\sqrt{\epsilon_2})^{2},
	\label{main-bound}
	\end{equation}
	where FID is the Fréchet inception distance \citep{heusel2017gans} and $L$ is the Lipschitz constant of the feature extractor of the pre-trained InceptionV3 neural network \citep{szegedy2016rethinking}.\label{thm-main}
	\end{theorem}
	We prove Theorem \ref{thm-main} in Appendix \ref{prf_thm43}. The duality gaps upper bound $L^{2}(\mu)$ norm between computed $\hat{G}$ and true $G^{*}$ maps, and the $\mathcal{W}_{2}^{2}$ between true $\nu$ and generated (fake) distribution $\hat{G}_{\#}\mu$. Consequently, they upper bound FID between data $\nu$ and fake (generated) $\hat{G}_{\#}\mu$ distributions.
	
	\textbf{Relation to previous works.} \cite{makkuva2020optimal, korotin2021wasserstein} prove related bounds for $\mathcal{W}_{2}$ with $\mu,\nu$ located on the spaces of the same dimension. Our result holds for different dimensions.
	
	\section{Experiments}
	\label{sec-exps}
	
	We evaluate our algorithm in generative modeling of the data distribution from a noise (\wasyparagraph\ref{sec-noise-to-data}) and unpaired image restoration task (\wasyparagraph\ref{sec-enhancement}). Technical details are given in Appendix \ref{sec-exp-details}. Additionally, in Appendix \ref{sec-addn-exps} we test our method on \rebut{toy 2D datasets and evaluate it on the Wasserstein-2 benchmark} \citep{korotin2021neural} in Appendix \ref{sec-benchmark-early}. The code is in the supplementary material.

\subsection{Modeling Data distribution from Noise Distribution}
\label{sec-noise-to-data}

In this subsection, $\mu$ is a $192$-dimensional normal noise and $\nu$ the high-dimensional data distribution. 

Let the images from $\nu$ be of size $w\times h$ with $c$ channels. As the embedding $Q:\mathcal{X}\rightarrow\mathcal{Y}$ we use a \textit{naive} upscaling of a noise. For $x\in\mathbb{R}^{192}$ we represent it as $3$-channel $8\times 8$ image and bicubically upscale it to the size $w\times h$ of data images from $\nu$. For grayscale images drawn from $\nu$, we stack $c$ copies over channel dimension.

\begin{figure*}[!t]
	\vspace{-1mm}
		\begin{subfigure}{0.33\textwidth}
			\centering
			\includegraphics[width=0.99\textwidth]{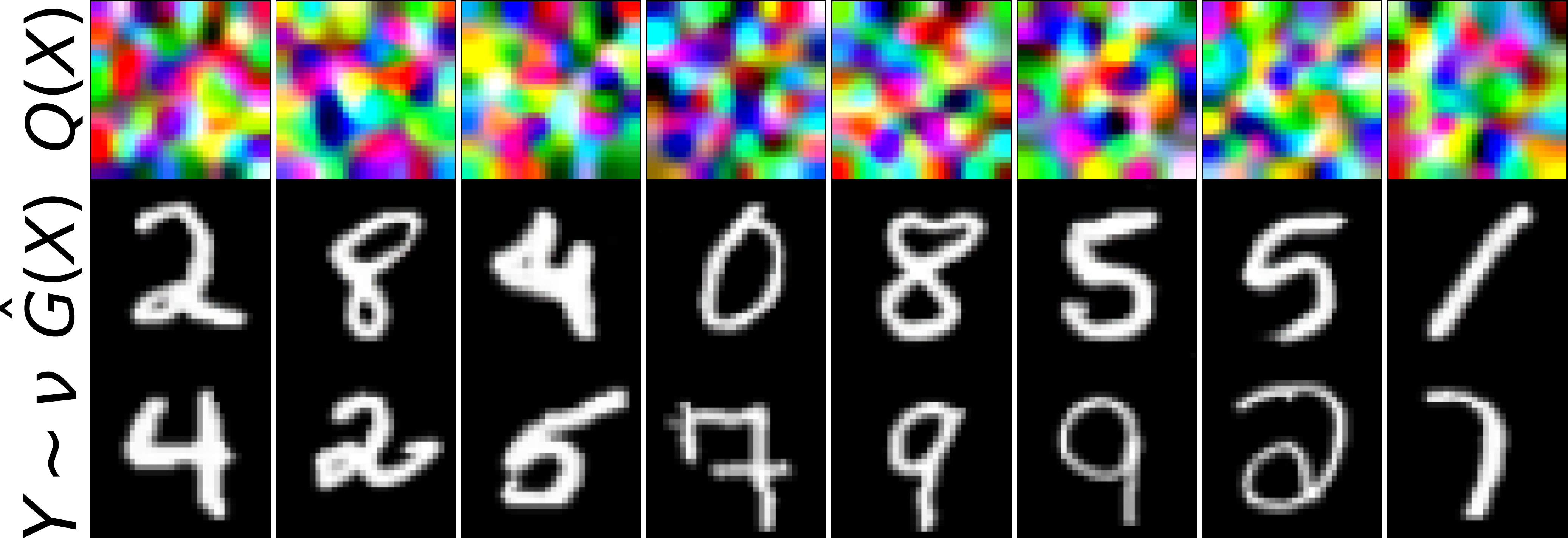}
			\caption{MNIST, $32\times 32$, grayscale}
			\label{mnist}
		\end{subfigure}  
		\begin{subfigure}{0.33\textwidth}
			\centering
			\includegraphics[width=0.99\textwidth]{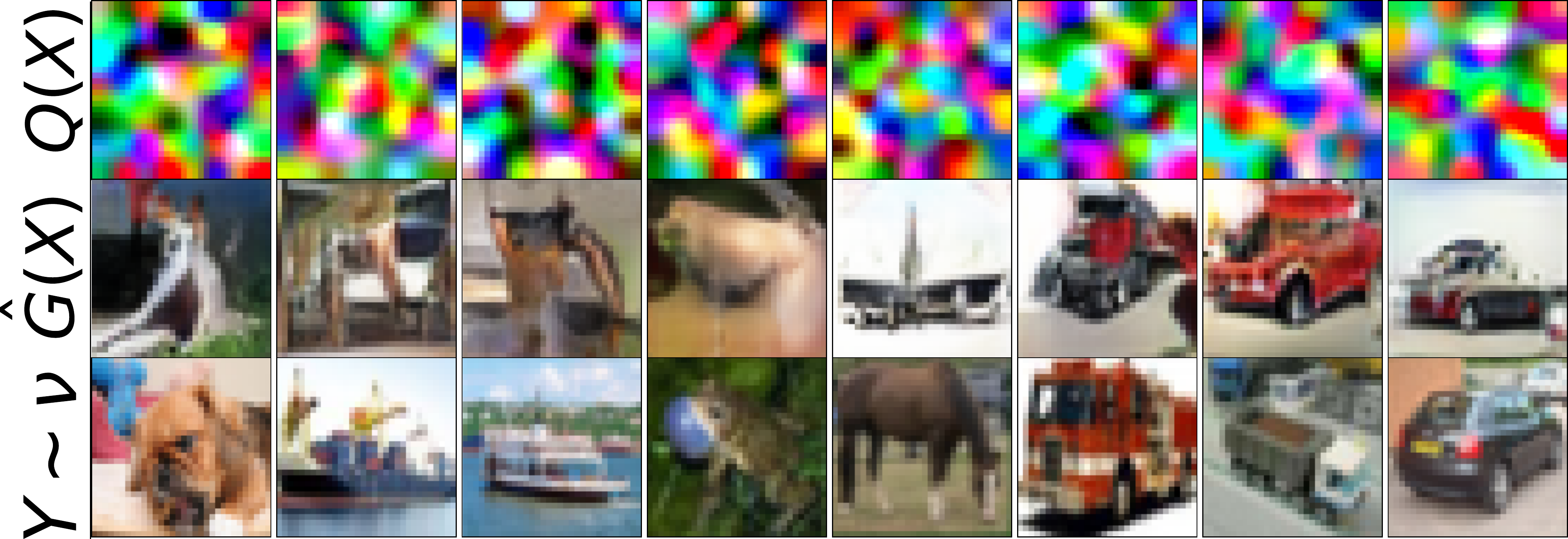}
			\caption{CIFAR10, $32\times 32$, RGB}
			\label{cifar}
		\end{subfigure}
		\begin{subfigure}{0.33\textwidth}
			\centering
			\includegraphics[width=0.99\textwidth]{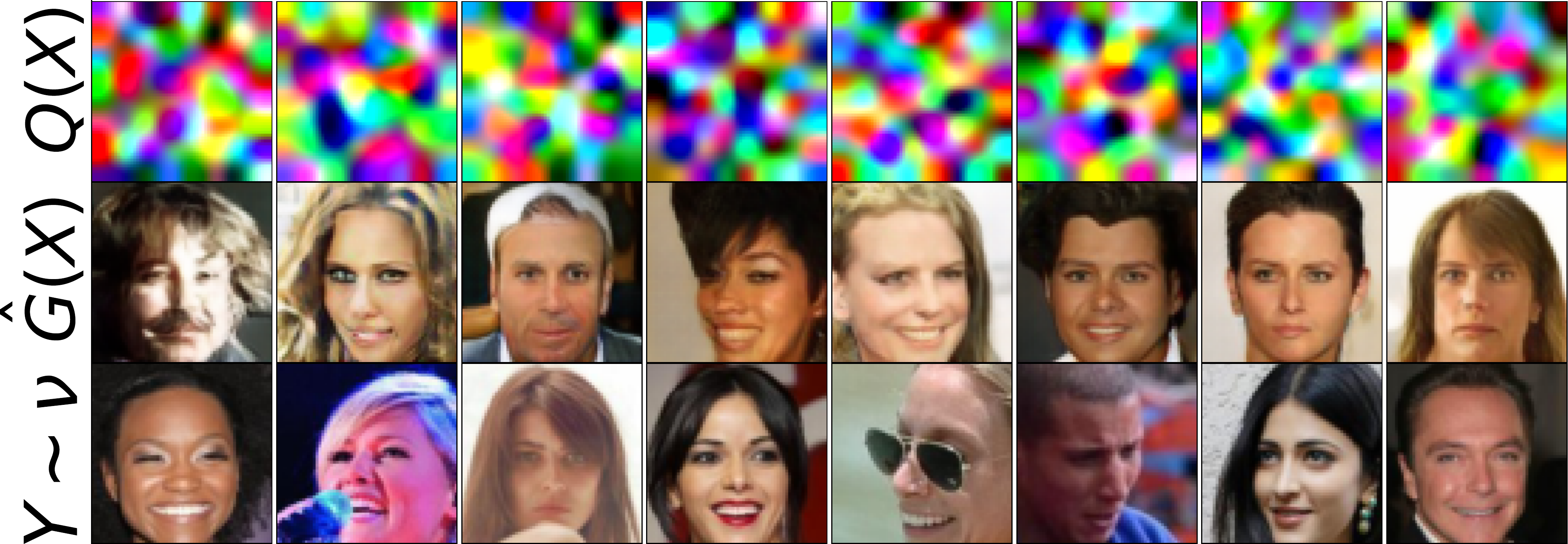}
			\caption{CelebA, $64\times 64$, RGB}
			\label{celeba}
		\end{subfigure}
	\vspace{-1mm}
		\caption{Randomly generated MNIST, CIFAR10, and CelebA samples by our method (OTM).}
		\label{fig:generated-samples}
	\end{figure*}

We test our method on MNIST $32\times 32$ \citep{lecun1998gradient}, CIFAR10 $32\times 32$ \citep{krizhevsky2009learning}, and CelebA $64\times 64$  \citep{liu2015deep} image datasets. In Figure~\ref{fig:generated-samples}, we show random samples generated by our approach, namely \textbf{Optimal Transport Modeling} (OTM). To quantify the results, in Tables \ref{table-cifar-fid} and \ref{table-celeba-fid} we give the inception \citep{salimans2016improved} and FID \citep{heusel2017gans} scores of generated samples. Similar to \cite[Appendix B.2]{song2019generative}, we compute them on 50K real and generated samples. \rebut{Additionally, in Appendix \ref{sec-addn-exps}, we test our method on $128\times 128$ CelebA faces. We provide qualitative results (images of generated faces) in Figure \ref{fig:celeba_128x128}.}


For comparison, we include the scores of existing generative models of three types: (1) OT map as the generative model; (2) OT cost as the loss; (3) not OT-based. Note that models of the first type compute OT in the \textit{latent space} of an autoencoder in contrast to our approach. According to our evaluation, the performance of our method is better or comparable to existing alternatives.
\begin{table}[!t]
	\vspace{-2mm}
		\centering
		\scriptsize
		\parbox{.45\linewidth}{
			\caption{Results on CIFAR10 dataset.}
			\label{table-cifar-fid}
			\begin{tabular}{llll}
				{\bf Model} & {\bf Related Work} & {\bf Inception $\uparrow$ } & {\bf FID $\downarrow$  } \\  \hline
				NVAE & \cite{vahdat2020nvae} & - & 51.71 \\
				PixelIQN &\cite{ostrovski2018autoregressive}   & 5.29     & 49.46 \\
				EBM & \cite{du2019implicit}  & 6.02                       & 40.58 \\
				DCGAN & \cite{radford2016unsupervised} & 6.64$\pm$0.14   & 37.70 \\
				NCSN & \cite{song2019generative} & 8.87$\pm$0.12     & 25.32 \\
				NCP-VAE & \cite{aneja2020ncp} & - & 24.08 \\ \hline
				WGAN & \cite{arjovsky2017wasserstein} & -                       & 55.2   \\
				WGAN-GP & \cite{gulrajani2017improved} & 6.49$\pm$0.09 & 39.40 \\
				3P-WGAN & \cite{nhan2019threeplayer} & 7.38 $\pm$ 0.08 & 28.8\\ \hline
				AE-OT & \cite{An2020AE-OT:} & - & 28.5  \\
				AE-OT-GAN & \cite{an2020ae} & - & 17.1  \\ \hline
				OTM  & Ours    & 7.42$\pm$0.06 & 21.78 \\
			\end{tabular}
		}
		\hspace{8mm}
		\parbox{.45\linewidth}{
			\centering
			\scriptsize
			\caption{Results on CelebA dataset.}
			\label{table-celeba-fid}
			\begin{tabular}{lll}
				{\bf Model}  & {\bf Related Work} & {\bf FID $\downarrow$ }
				\\ \hline
				DCGAN    & \cite{radford2016unsupervised} & 52.0            \\
				DRAGAN  & \cite{kodali2017convergence} & 42.3            \\ 
				BEGAN & \cite{berthelot2017began} & 38.9 \\
				NVAE & \cite{vahdat2020nvae} & 13.4\\
				NCP-VAE & \cite{aneja2020ncp} & 5.2
				\\ \hline 
				WGAN & \cite{arjovsky2017wasserstein} & 41.3  \\ 
				WGAN-GP & \cite{gulrajani2017improved} & 30.0 \\
				WGAN-QC	& \cite{Liu_2019_ICCV} & 12.9 \\ \hline 
				AE-OT & \cite{An2020AE-OT:} & 28.6 		\\
				AE-OT-GAN & \cite{an2020ae} & 7.8	\\ \hline
				OTM  & Ours & 6.5            \\
			\end{tabular}
		}
	\vspace{-3mm}
	\end{table}

\subsection{Unpaired Image Restoration}

\label{sec-enhancement}
In this subsection, we consider unpaired image restoration tasks on CelebA faces dataset. In this case, the input distribution $\mu$ consists of degraded images, while $\nu$ are clean images. In all the cases, embedding $Q$ is a straightforward identity embedding $Q(x)\equiv x$.

In image restoration, optimality of the restoration map is desired since the output (restored) image is expected to be close to the input (degraded) one minimizing the transport cost. Note that GANs do not seek for an optimal mapping. However, in practice, due to implicit inductive biases such as convolutional architectures, GANs still tend to fit low transport cost maps \citep{bezenac2021cyclegan}.

\begin{figure}[!h]
	\vspace{-3mm}
	\centering
	\includegraphics[width=0.65\linewidth]{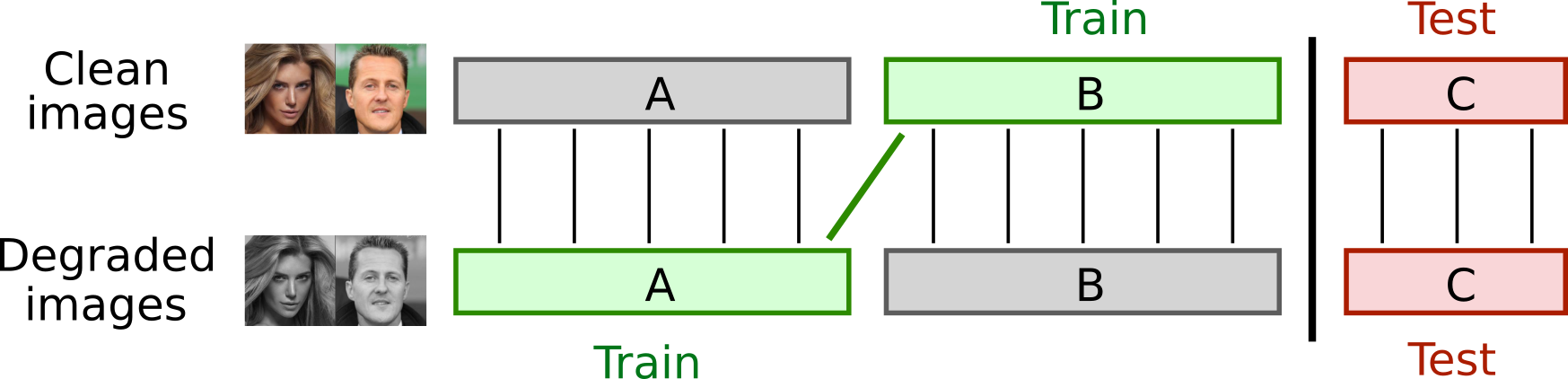}
	\vspace{-1mm}
	\caption{\centering The training/testing scheme that we use for unpaired restoration tasks.}
	\label{fig:setup-restore}
	\vspace{-3mm}
\end{figure}

The experimental setup is shown in Figure \ref{fig:setup-restore}. We split the dataset in 3 parts A, B, C containing 90K, 90K, 22K samples respectively. To each image we apply the degradation transform (decolorization, noising or occlusion) and obtain the degraded dataset containing of $3$ respective parts A, B, C. For \textit{unpaired} training we use part A of degraded and part B of clean images. For testing, we use parts C.

To quantify the results we compute FID of restored images w.r.t.\ clean images of part C. The scores for denoising, inpainting and colorization are given in Table \ref{tab:fid-restoration}, details of each experiment and qualitative results are given below.
\begin{table}[!h]
	\vspace{-3mm}
    \centering
		\small
    \begin{tabular}{llll}
		{\bf Model} & {\bf Denoising}  & {\bf Colorization} & {\bf Inpainting } \\  \hline
		Input &  166.59   & 32.12 & 47.65  \\
		WGAN-GP   & 25.49   &  7.75  &    16.51        \\ 
		OTM-GP (ours)   &  10.95    & 5.66  & 9.96            \\
		OTM (ours)   &  5.92    & 5.65  & 8.13            \\
	\end{tabular}
	\vspace{-1mm}
    \caption{FID$\downarrow$ on test part C in image restoration experiments.}
	\vspace{-5mm}
    \label{tab:fid-restoration}
\end{table}

As a baseline, we include WGAN-GP. For a~fair comparison, we fit it using \textit{exactly} the same hyperparameters as in our method OTM-GP. This is possible due to the similarities between our method and WGAN-GP's training procedure, see discussion in \wasyparagraph\ref{sec-optim}. In OTM, there is no GP (\wasyparagraph \ref{sec-dis-gp}).

\begin{figure}[!h]
	\vspace{-2mm}
	\captionsetup[subfigure]{justification=centering}
	\begin{subfigure}{0.33\textwidth}
		\centering
		\includegraphics[width=0.99\textwidth]{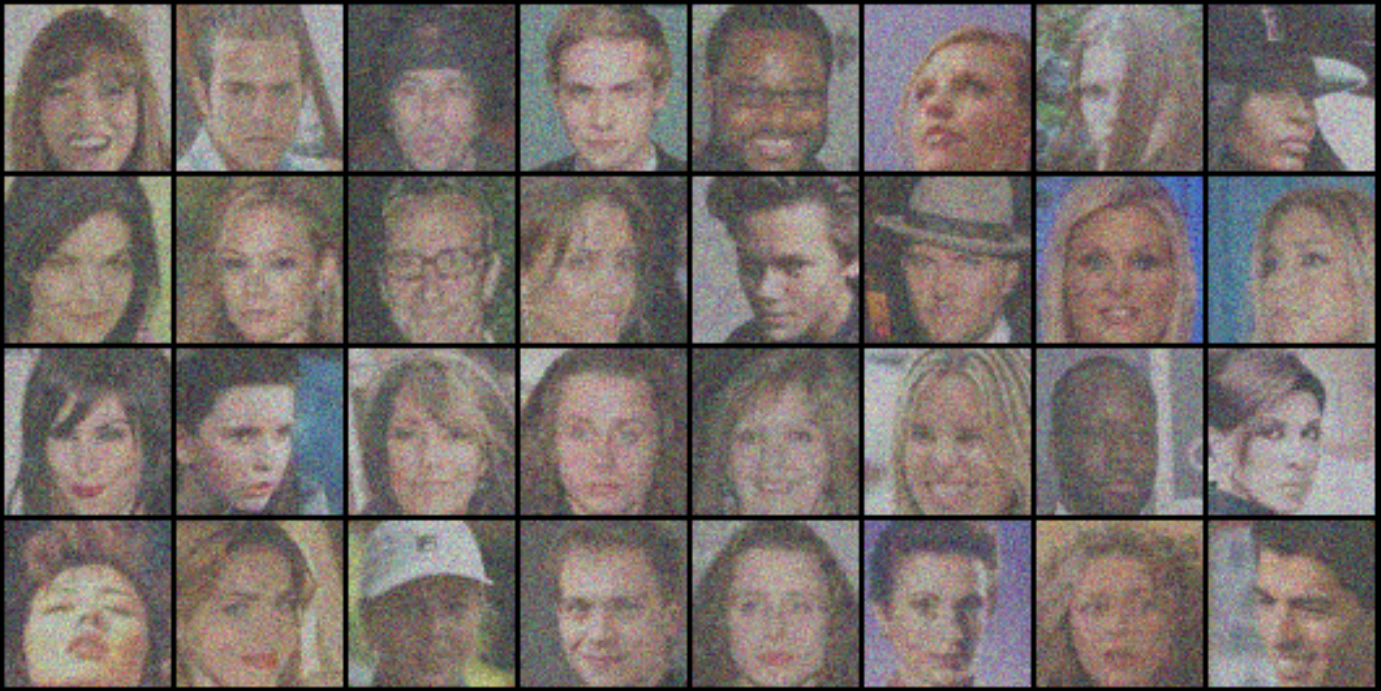}
		\caption{Noisy samples. }
		\label{denoise_ip}
	\end{subfigure}
	\begin{subfigure}{0.33\textwidth}
		\centering
		\includegraphics[width=0.99\textwidth]{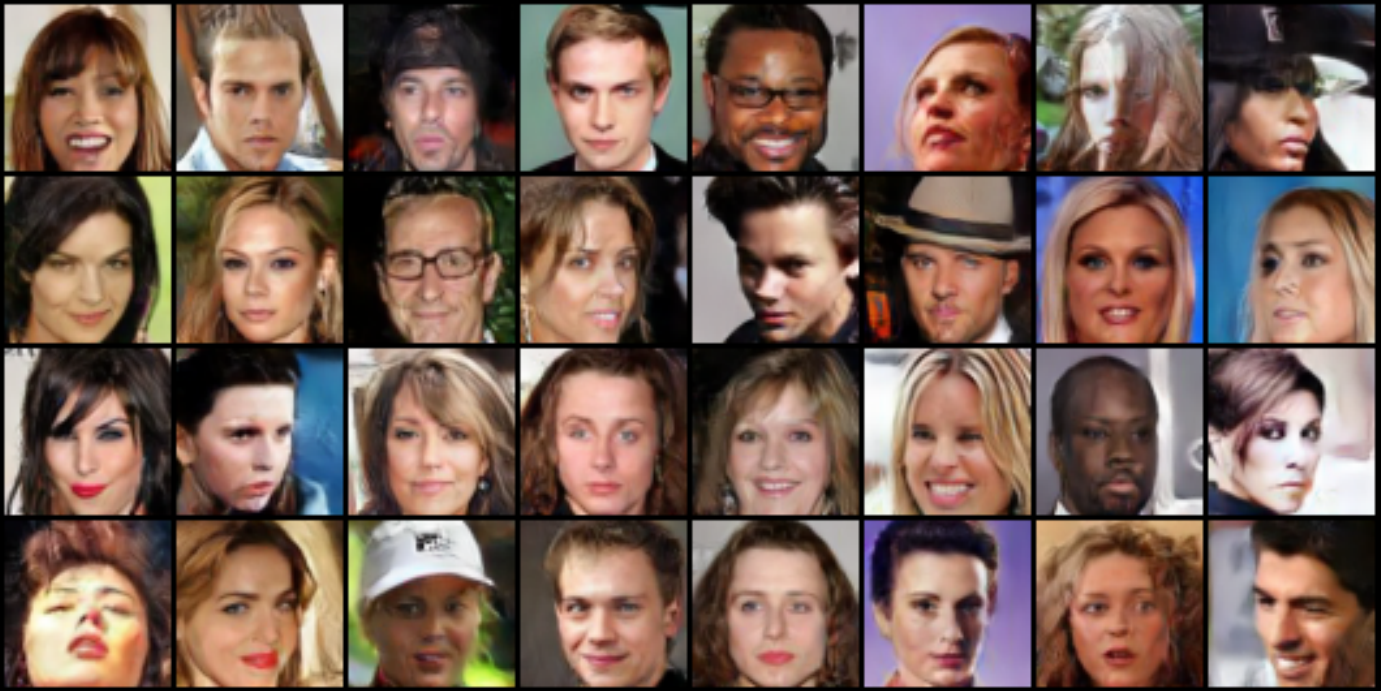}
		\caption{Pushforward samples. }
		\label{denoise_op}
	\end{subfigure}
	\begin{subfigure}{0.33\textwidth}
		\centering
		\includegraphics[width=0.99\textwidth]{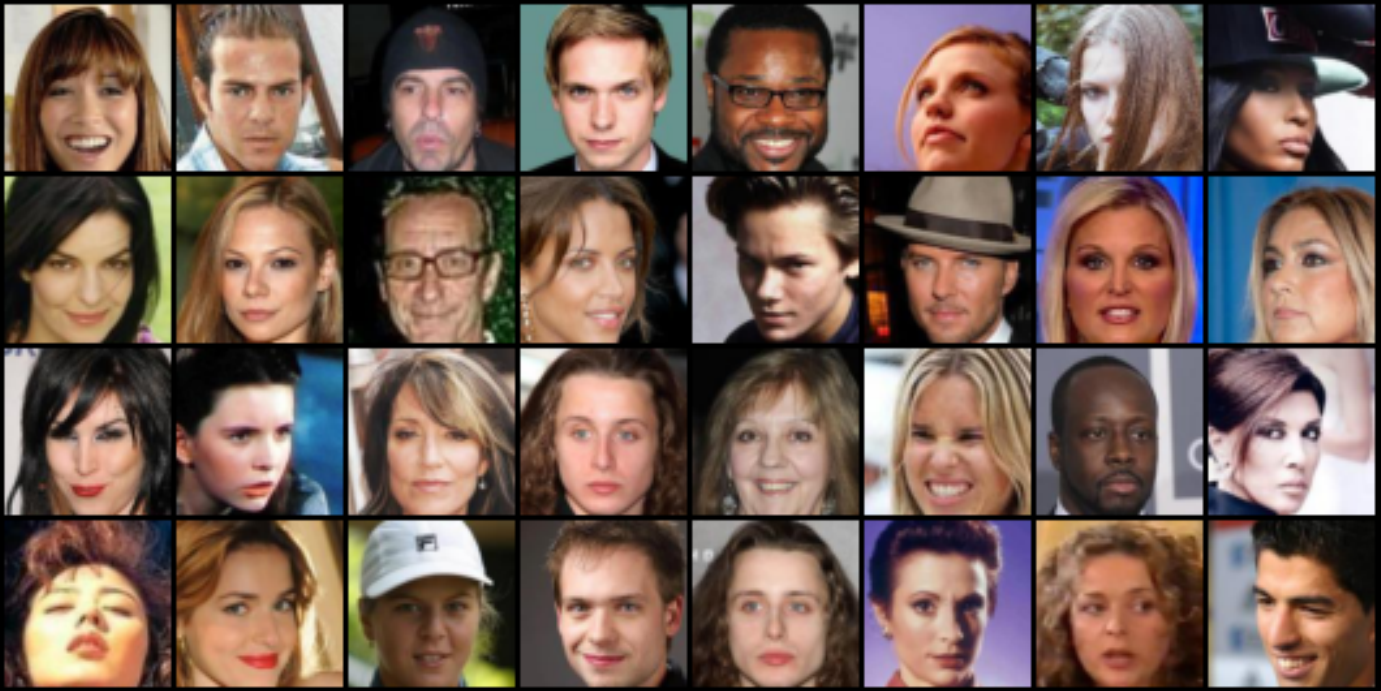}
		\caption{Original samples. }
		\label{denoise_org}
	\end{subfigure} 
	\vspace{-3mm}
	\caption{OTM for image denoising on test C part of CelebA, $64\times 64$.}
	\vspace{-3mm}
	\label{otm_denoise}
\end{figure}

	
\textbf{Denoising}. To create noisy images, we add white normal noise with $\sigma=0.3$ to each pixel. Figure~\ref{otm_denoise} illustrates image denoising using our OTM approach on the test part of the dataset. We show additional qualitative results for varying $\sigma$ in Figure \ref{otm_denoise_sigmas} of \wasyparagraph\eqref{sec-addn-exps}.

\begin{figure}[!h]
	\vspace{-3mm}
	\captionsetup[subfigure]{justification=centering}
		\begin{subfigure}{0.33\textwidth}
			\centering
			\includegraphics[width=0.99\textwidth]{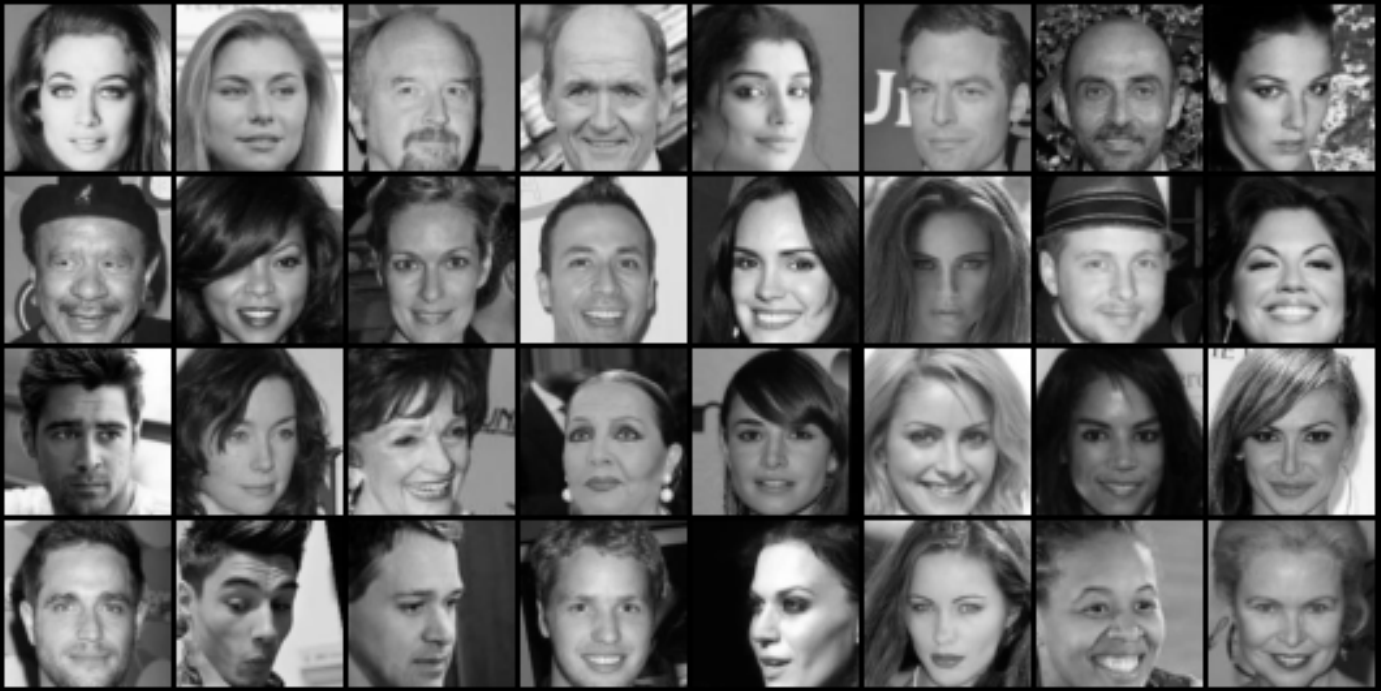}
			\caption{Grayscale samples. }
			\label{g2rgb_ip}
		\end{subfigure}
		\begin{subfigure}{0.33\textwidth}
			\centering
			\includegraphics[width=0.99\textwidth]{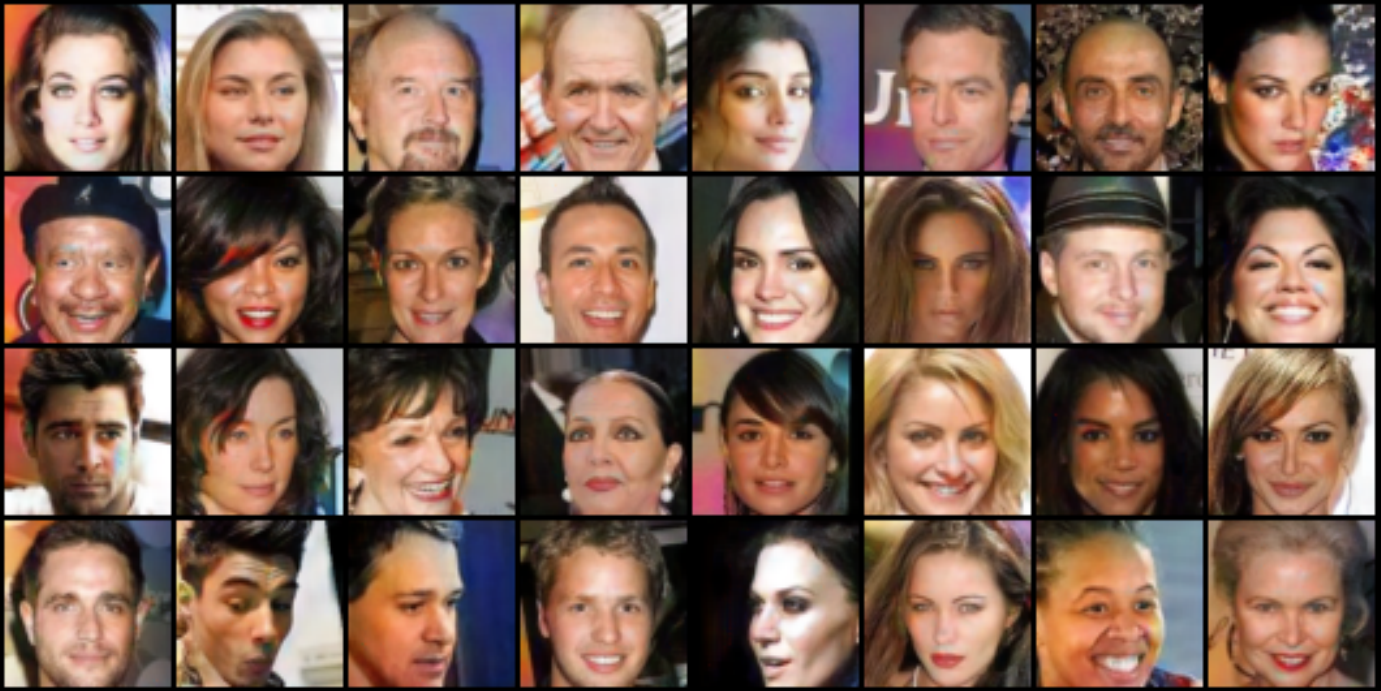}
			\caption{Pushforward samples. }
			\label{g2rgb_op}
		\end{subfigure}
	    \begin{subfigure}{0.33\textwidth}
			\centering
			\includegraphics[width=0.99\textwidth]{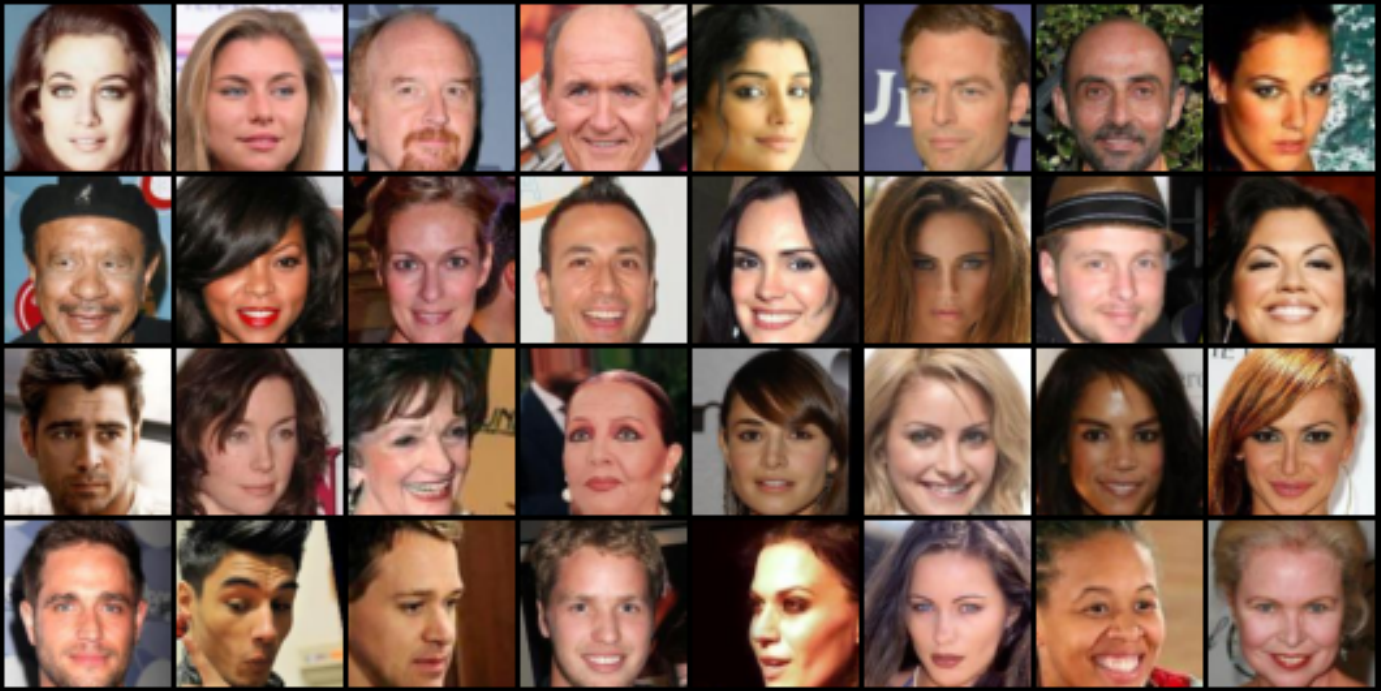}
			\caption{Original samples. }
			\label{g2rgb_org}
		\end{subfigure}  
	\vspace{-3mm}
		\caption{OTM for image colorization on test C part of  CelebA, $64\times 64$.}
		\label{otm_colorization}
	\vspace{-3mm}
\end{figure}

\textbf{Colorization.} To create grayscale images, we average the RGB values of each pixel. Figure~\ref{otm_colorization} illustrates image colorization using OTM on the test part of the dataset.

\begin{figure*}[!h]
	\vspace{-3mm}
	\captionsetup[subfigure]{justification=centering}
		\begin{subfigure}{0.33\textwidth}
			\centering
			\includegraphics[width=0.99\textwidth]{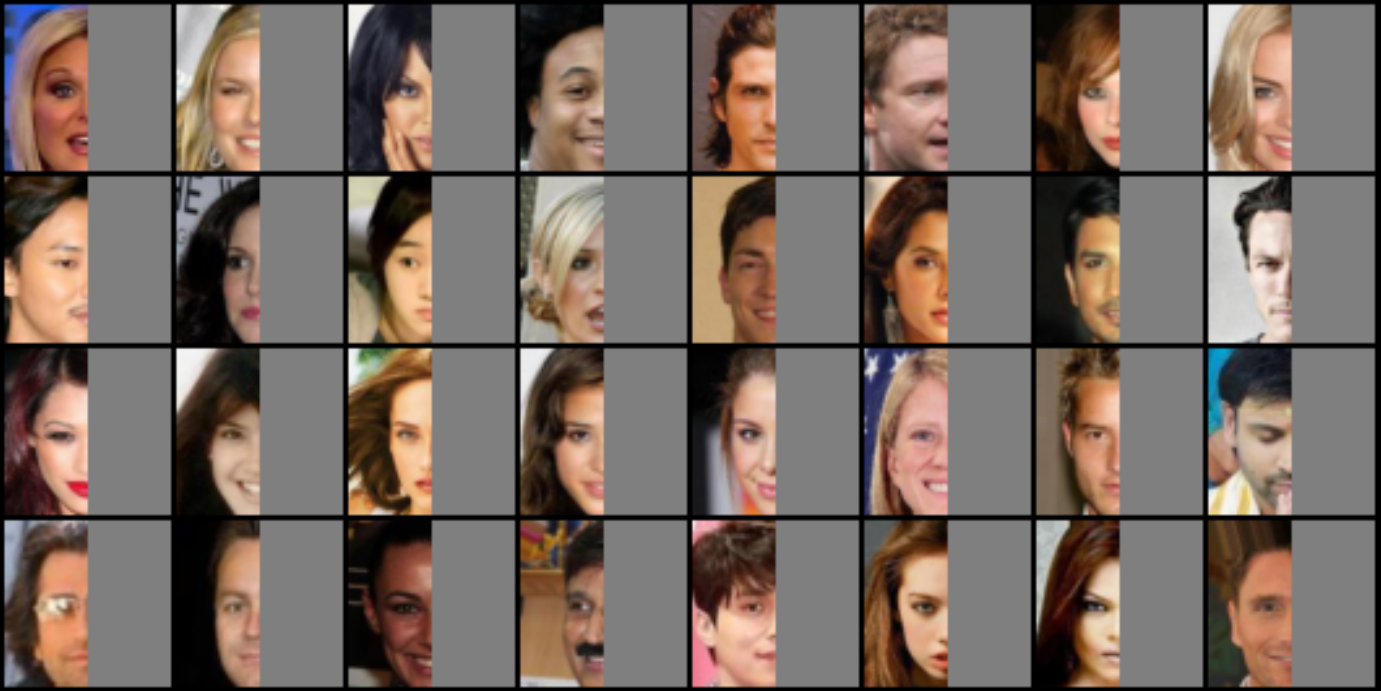}
			\caption{Occluded samples. }
			\label{inpaint_ip}
		\end{subfigure}
		\begin{subfigure}{0.33\textwidth}
			\centering
			\includegraphics[width=0.99\textwidth]{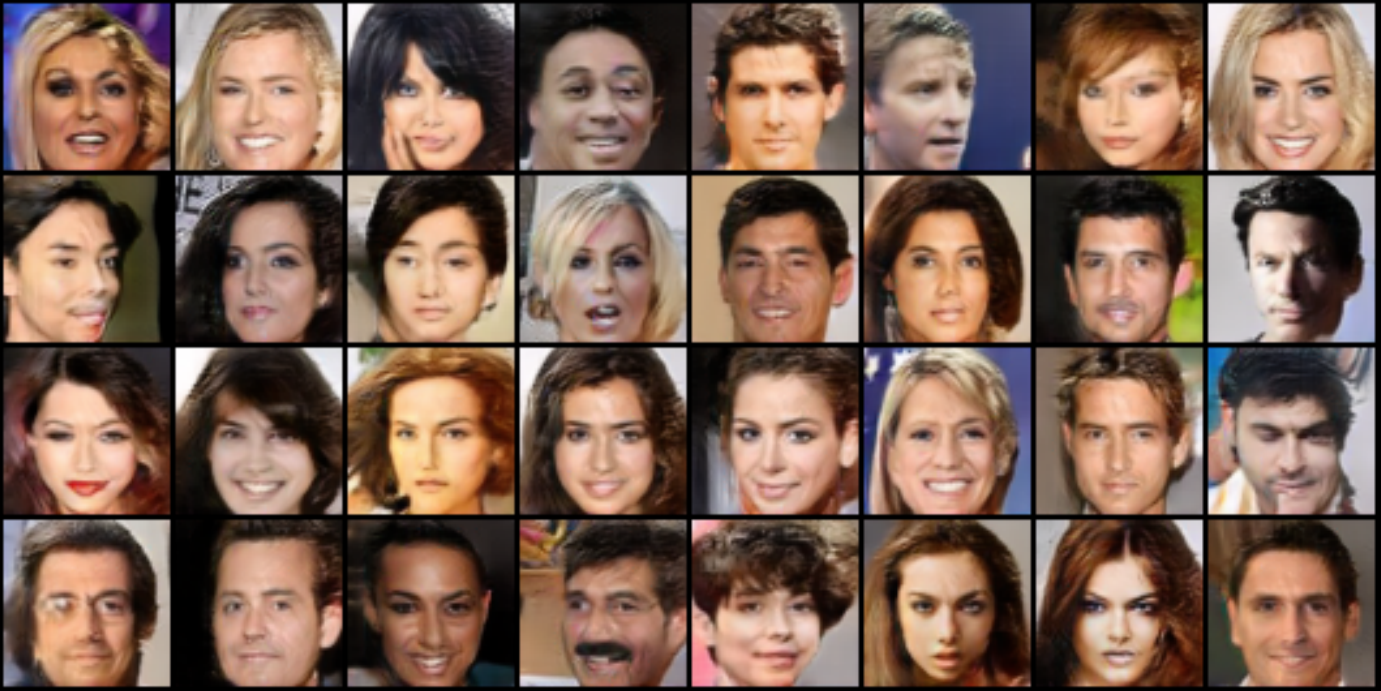}
			\caption{Pushforward samples. }
			\label{inpaint_op}
		\end{subfigure}
	    \begin{subfigure}{0.33\textwidth}
			\centering
			\includegraphics[width=0.99\textwidth]{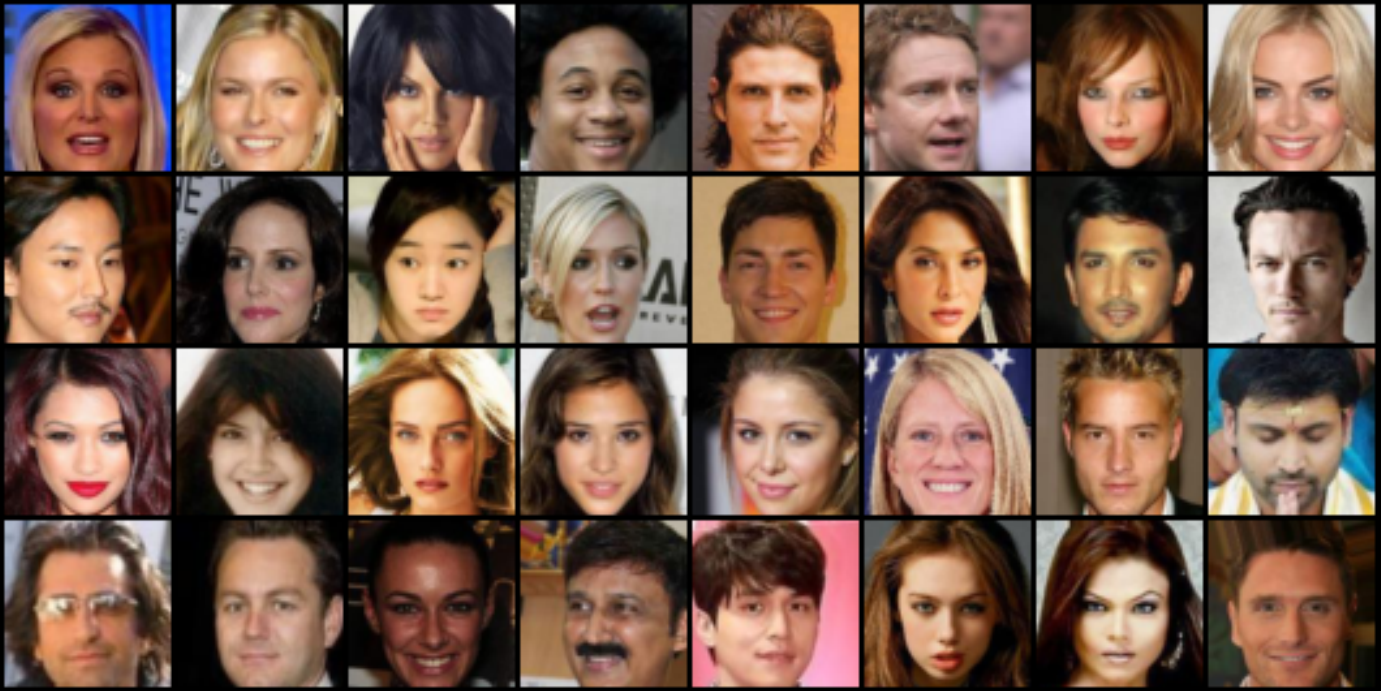}
			\caption{Original samples.}
			\label{inpaint_org}
		\end{subfigure}  
	\vspace{-3mm}
		\caption{OTM for image inpainting on test C part of CelebA, $64\times 64$.}
		\label{otm_inpaint}
	\vspace{-2mm}
	\end{figure*}

\textbf{Inpainting}. To create incomplete images, we replace the right half of each clean image with zeros. Figure~\ref{otm_inpaint} illustrates image inpainting using OTM on the test part of the dataset.

\vspace{-6mm}\section{Conclusion}
	\label{conc}
	\vspace{-3mm}
Our method fits OT maps for the embedded quadratic transport cost between probability distributions. Unlike predecessors, it scales well to high dimensions producing applications of OT maps directly in ambient spaces, such as spaces of images. The performance is comparable to other existing generative models while the complexity of training is similar to that of popular WGANs.
	
\textbf{Limitations.} For distributions $\mu,\nu$ we assume the existence of the OT map between them. In practice, this might not hold for all real-world $\mu,\nu$. Working with equal dimensions, we focus on the quadratic ground cost $\frac{1}{2}\|x-y\|^{2}$. Nevertheless, our approach extends to other costs $c(\cdot,\cdot)$, see \cite{fan2021scalable}. When the dimensions are unequal, we restrict our analysis to \textit{embedded} quadratic cost $\frac{1}{2}\|Q(x)-y\|^{2}$ where $Q$ equalizes dimensions. Choosing the embedding $Q$ might not be straightforward in some practical problems, but our evaluation (\wasyparagraph\ref{sec-noise-to-data}) shows that even naive choices of $Q$ work well.
	
\vspace{-1mm}\textbf{Potential impact and ethics.} 
	Real-world image restoration problems often do not have paired datasets limiting the application of supervised techniques. In these practical unpaired learning problems, we expect our optimal transport approach to improve the performance of the existing models. However, biases in data might lead to biases in the pushforward samples. This should be taken into account when using our method in practical problems.
	
	\textbf{Reproducibility.} The \texttt{PyTorch} source code is provided at 
	\begin{center}\url{https://github.com/LituRout/OptimalTransportModeling}\end{center}
	The instructions to use the code are included in the \texttt{README.md} file.

\section{Acknowledgment}
This research was supported by the computational resources provided by Space Applications Centre (SAC), ISRO. The first author acknowledges the funding by HRD Grant No. 0303T50FM703/SAC/ISRO. Skoltech RAIC center was supported by the RF Government (subsidy agreement 000000D730321P5Q0002, Grant No. 70-2021-00145 02.11.2021).

	\bibliography{egbib}

\begin{thebibliography}{64}
\providecommand{\natexlab}[1]{#1}
\providecommand{\url}[1]{\texttt{#1}}
\expandafter\ifx\csname urlstyle\endcsname\relax
  \providecommand{\doi}[1]{doi: #1}\else
  \providecommand{\doi}{doi: \begingroup \urlstyle{rm}\Url}\fi

\bibitem[Amodio \& Krishnaswamy(2019)Amodio and
  Krishnaswamy]{amodio2019travelgan}
Matthew Amodio and Smita Krishnaswamy.
\newblock Travelgan: Image-to-image translation by transformation vector
  learning.
\newblock In \emph{Proceedings of the IEEE/CVF Conference on Computer Vision
  and Pattern Recognition}, pp.\  8983--8992, 2019.

\bibitem[Amos et~al.(2017)Amos, Xu, and Kolter]{amos2017input}
Brandon Amos, Lei Xu, and J~Zico Kolter.
\newblock Input convex neural networks.
\newblock In \emph{International Conference on Machine Learning}, pp.\
  146--155. PMLR, 2017.

\bibitem[An et~al.(2020{\natexlab{a}})An, Guo, Lei, Luo, Yau, and
  Gu]{An2020AE-OT:}
Dongsheng An, Yang Guo, Na~Lei, Zhongxuan Luo, Shing-Tung Yau, and Xianfeng Gu.
\newblock Ae-ot: A new generative model based on extended semi-discrete optimal
  transport.
\newblock In \emph{International Conference on Learning Representations},
  2020{\natexlab{a}}.
\newblock URL \url{https://openreview.net/forum?id=HkldyTNYwH}.

\bibitem[An et~al.(2020{\natexlab{b}})An, Guo, Zhang, Qi, Lei, and
  Gu]{an2020ae}
Dongsheng An, Yang Guo, Min Zhang, Xin Qi, Na~Lei, and Xianfang Gu.
\newblock Ae-ot-gan: Training gans from data specific latent distribution.
\newblock In \emph{European Conference on Computer Vision}, pp.\  548--564.
  Springer, 2020{\natexlab{b}}.

\bibitem[Aneja et~al.(2021)Aneja, Schwing, Kautz, and Vahdat]{aneja2020ncp}
Jyoti Aneja, Alexander Schwing, Jan Kautz, and Arash Vahdat.
\newblock Ncp-vae: Variational autoencoders with noise contrastive priors.
\newblock In \emph{Advances in Neural Information Processing Systems
  Conference}, 2021.

\bibitem[Anil et~al.(2019)Anil, Lucas, and Grosse]{anil2019sorting}
Cem Anil, James Lucas, and Roger Grosse.
\newblock Sorting out lipschitz function approximation.
\newblock In \emph{International Conference on Machine Learning}, pp.\
  291--301. PMLR, 2019.

\bibitem[Arjovsky \& Bottou(2017)Arjovsky and Bottou]{arjovsky2017towards}
Martin Arjovsky and L{\'e}on Bottou.
\newblock Towards principled methods for training generative adversarial
  networks.
\newblock \emph{arXiv preprint arXiv:1701.04862}, 2017.

\bibitem[Arjovsky et~al.(2017)Arjovsky, Chintala, and
  Bottou]{arjovsky2017wasserstein}
Martin Arjovsky, Soumith Chintala, and L{\'e}on Bottou.
\newblock Wasserstein generative adversarial networks.
\newblock In \emph{International conference on machine learning}, pp.\
  214--223. PMLR, 2017.

\bibitem[Berthelot et~al.(2017)Berthelot, Schumm, and Metz]{berthelot2017began}
David Berthelot, Thomas Schumm, and Luke Metz.
\newblock Began: Boundary equilibrium generative adversarial networks.
\newblock \emph{arXiv preprint arXiv:1703.10717}, 2017.

\bibitem[B{\'e}zenac et~al.(2021)B{\'e}zenac, Ayed, and
  Gallinari]{bezenac2021cyclegan}
Emmanuel~de B{\'e}zenac, Ibrahim Ayed, and Patrick Gallinari.
\newblock Cyclegan through the lens of (dynamical) optimal transport.
\newblock In \emph{Joint European Conference on Machine Learning and Knowledge
  Discovery in Databases}, pp.\  132--147. Springer, 2021.

\bibitem[Brenier(1991)]{brenier1991polar}
Yann Brenier.
\newblock Polar factorization and monotone rearrangement of vector-valued
  functions.
\newblock \emph{Communications on pure and applied mathematics}, 44\penalty0
  (4):\penalty0 375--417, 1991.

\bibitem[Brock et~al.(2019)Brock, Donahue, and Simonyan]{brock2018large}
Andrew Brock, Jeff Donahue, and Karen Simonyan.
\newblock Large scale {GAN} training for high fidelity natural image synthesis.
\newblock In \emph{International Conference on Learning Representations}, 2019.
\newblock URL \url{https://openreview.net/forum?id=B1xsqj09Fm}.

\bibitem[Dai \& Seljak(2021)Dai and Seljak]{dai2021sliced}
Biwei Dai and Uros Seljak.
\newblock Sliced iterative normalizing flows.
\newblock In \emph{ICML Workshop on Invertible Neural Networks, Normalizing
  Flows, and Explicit Likelihood Models}, 2021.

\bibitem[Deshpande et~al.(2018)Deshpande, Zhang, and
  Schwing]{deshpande2018generative}
Ishan Deshpande, Ziyu Zhang, and Alexander~G Schwing.
\newblock Generative modeling using the sliced wasserstein distance.
\newblock In \emph{Proceedings of the IEEE conference on computer vision and
  pattern recognition}, pp.\  3483--3491, 2018.

\bibitem[Dowson \& Landau(1982)Dowson and Landau]{dowson1982frechet}
DC~Dowson and BV~Landau.
\newblock The fr{\'e}chet distance between multivariate normal distributions.
\newblock \emph{Journal of multivariate analysis}, 12\penalty0 (3):\penalty0
  450--455, 1982.

\bibitem[Du \& Mordatch(2019)Du and Mordatch]{du2019implicit}
Yilun Du and Igor Mordatch.
\newblock Implicit generation and generalization in energy-based models.
\newblock \emph{arXiv preprint arXiv:1903.08689}, 2019.

\bibitem[Fan et~al.(2021)Fan, Liu, Ma, Chen, and Zhou]{fan2021scalable}
Jiaojiao Fan, Shu Liu, Shaojun Ma, Yongxin Chen, and Haomin Zhou.
\newblock Scalable computation of monge maps with general costs.
\newblock \emph{arXiv preprint arXiv:2106.03812}, 2021.

\bibitem[Fatras et~al.(2019)Fatras, Zine, Flamary, Gribonval, and
  Courty]{fatras2019learning}
Kilian Fatras, Younes Zine, R{\'e}mi Flamary, R{\'e}mi Gribonval, and Nicolas
  Courty.
\newblock Learning with minibatch wasserstein: asymptotic and gradient
  properties.
\newblock \emph{arXiv preprint arXiv:1910.04091}, 2019.

\bibitem[Genevay et~al.(2018)Genevay, Peyr{\'e}, and
  Cuturi]{genevay2018learning}
Aude Genevay, Gabriel Peyr{\'e}, and Marco Cuturi.
\newblock Learning generative models with sinkhorn divergences.
\newblock In \emph{International Conference on Artificial Intelligence and
  Statistics}, pp.\  1608--1617. PMLR, 2018.

\bibitem[Goodfellow et~al.(2014)Goodfellow, Pouget-Abadie, Mirza, Xu,
  Warde-Farley, Ozair, Courville, and Bengio]{goodfellow2014generative}
Ian~J Goodfellow, Jean Pouget-Abadie, Mehdi Mirza, Bing Xu, David Warde-Farley,
  Sherjil Ozair, Aaron~C Courville, and Yoshua Bengio.
\newblock Generative adversarial nets.
\newblock In \emph{Advances in Neural Information Processing Systems
  Conference}, pp.\  2674--2680, 2014.

\bibitem[Gulrajani et~al.(2017)Gulrajani, Ahmed, Arjovsky, Dumoulin, and
  Courville]{gulrajani2017improved}
Ishaan Gulrajani, Faruk Ahmed, Martin Arjovsky, Vincent Dumoulin, and Aaron
  Courville.
\newblock Improved training of wasserstein gans.
\newblock In \emph{Proceedings of the 31st International Conference on Neural
  Information Processing Systems}, pp.\  5769--5779, 2017.

\bibitem[Heusel et~al.(2017)Heusel, Ramsauer, Unterthiner, Nessler, and
  Hochreiter]{heusel2017gans}
Martin Heusel, Hubert Ramsauer, Thomas Unterthiner, Bernhard Nessler, and Sepp
  Hochreiter.
\newblock Gans trained by a two time-scale update rule converge to a local nash
  equilibrium.
\newblock \emph{Advances in neural information processing systems}, 30, 2017.

\bibitem[Jacob et~al.(2018)Jacob, She, Almahairi, Rajeswar, and
  Courville]{jacob2018w2gan}
Leygonie Jacob, Jennifer She, Amjad Almahairi, Sai Rajeswar, and Aaron
  Courville.
\newblock W2gan: Recovering an optimal transport map with a gan.
\newblock 2018.

\bibitem[Kantorovich(1948)]{kantorovich1948problem}
Leonid~Vitalevich Kantorovich.
\newblock On a problem of monge.
\newblock \emph{Uspekhi Mat. Nauk}, pp.\  225--226, 1948.

\bibitem[Karras et~al.(2019)Karras, Laine, and Aila]{karras2019style}
Tero Karras, Samuli Laine, and Timo Aila.
\newblock A style-based generator architecture for generative adversarial
  networks.
\newblock In \emph{Proceedings of the IEEE/CVF Conference on Computer Vision
  and Pattern Recognition}, pp.\  4401--4410, 2019.

\bibitem[Kim et~al.(2021)Kim, Park, and Hwang]{kim2021local}
Cheolhyeong Kim, Seungtae Park, and Hyung~Ju Hwang.
\newblock Local stability of wasserstein gans with abstract gradient penalty.
\newblock \emph{IEEE Transactions on Neural Networks and Learning Systems},
  2021.

\bibitem[Kingma \& Ba(2014)Kingma and Ba]{kingma2014adam}
Diederik~P Kingma and Jimmy Ba.
\newblock Adam: A method for stochastic optimization.
\newblock \emph{arXiv preprint arXiv:1412.6980}, 2014.

\bibitem[Kingma \& Welling(2013)Kingma and Welling]{kingma2013auto}
Diederik~P Kingma and Max Welling.
\newblock Auto-encoding variational bayes.
\newblock \emph{arXiv preprint arXiv:1312.6114}, 2013.

\bibitem[Kodali et~al.(2017)Kodali, Abernethy, Hays, and
  Kira]{kodali2017convergence}
Naveen Kodali, Jacob Abernethy, James Hays, and Zsolt Kira.
\newblock On convergence and stability of gans.
\newblock \emph{arXiv preprint arXiv:1705.07215}, 2017.

\bibitem[Korotin et~al.(2021{\natexlab{a}})Korotin, Egiazarian, Asadulaev,
  Safin, and Burnaev]{korotin2021wasserstein}
Alexander Korotin, Vage Egiazarian, Arip Asadulaev, Alexander Safin, and Evgeny
  Burnaev.
\newblock Wasserstein-2 generative networks.
\newblock In \emph{International Conference on Learning Representations},
  2021{\natexlab{a}}.
\newblock URL \url{https://openreview.net/forum?id=bEoxzW_EXsa}.

\bibitem[Korotin et~al.(2021{\natexlab{b}})Korotin, Li, Genevay, Solomon,
  Filippov, and Burnaev]{korotin2021neural}
Alexander Korotin, Lingxiao Li, Aude Genevay, Justin Solomon, Alexander
  Filippov, and Evgeny Burnaev.
\newblock Do neural optimal transport solvers work? a continuous wasserstein-2
  benchmark.
\newblock \emph{arXiv preprint arXiv:2106.01954}, 2021{\natexlab{b}}.

\bibitem[Krizhevsky et~al.(2009)Krizhevsky, Hinton,
  et~al.]{krizhevsky2009learning}
Alex Krizhevsky, Geoffrey Hinton, et~al.
\newblock Learning multiple layers of features from tiny images.
\newblock 2009.

\bibitem[LeCun et~al.(1998)LeCun, Bottou, Bengio, and
  Haffner]{lecun1998gradient}
Yann LeCun, L{\'e}on Bottou, Yoshua Bengio, and Patrick Haffner.
\newblock Gradient-based learning applied to document recognition.
\newblock \emph{Proceedings of the IEEE}, 86\penalty0 (11):\penalty0
  2278--2324, 1998.

\bibitem[Lei et~al.(2019)Lei, Su, Cui, Yau, and Gu]{lei2019geometric}
Na~Lei, Kehua Su, Li~Cui, Shing-Tung Yau, and Xianfeng~David Gu.
\newblock A geometric view of optimal transportation and generative model.
\newblock \emph{Computer Aided Geometric Design}, 68:\penalty0 1--21, 2019.

\bibitem[Liu et~al.(2019)Liu, Gu, and Samaras]{Liu_2019_ICCV}
Huidong Liu, Xianfeng Gu, and Dimitris Samaras.
\newblock Wasserstein gan with quadratic transport cost.
\newblock In \emph{Proceedings of the IEEE/CVF International Conference on
  Computer Vision (ICCV)}, October 2019.

\bibitem[Liu et~al.(2021)Liu, Ma, Chen, Zha, and Zhou]{liu2021learning}
Shu Liu, Shaojun Ma, Yongxin Chen, Hongyuan Zha, and Haomin Zhou.
\newblock Learning high dimensional wasserstein geodesics.
\newblock \emph{arXiv preprint arXiv:2102.02992}, 2021.

\bibitem[Liu et~al.(2015)Liu, Luo, Wang, and Tang]{liu2015deep}
Ziwei Liu, Ping Luo, Xiaogang Wang, and Xiaoou Tang.
\newblock Deep learning face attributes in the wild.
\newblock In \emph{Proceedings of the IEEE international conference on computer
  vision}, pp.\  3730--3738, 2015.

\bibitem[Lu et~al.(2019)Lu, Zhou, Song, Ren, and Yu]{lu2019guiding}
Guansong Lu, Zhiming Zhou, Yuxuan Song, Kan Ren, and Yong Yu.
\newblock Guiding the one-to-one mapping in cyclegan via optimal transport.
\newblock In \emph{Proceedings of the AAAI Conference on Artificial
  Intelligence}, volume~33, pp.\  4432--4439, 2019.

\bibitem[Lu et~al.(2020)Lu, Zhou, Shen, Chen, Zhang, and Yu]{lu2020large}
Guansong Lu, Zhiming Zhou, Jian Shen, Cheng Chen, Weinan Zhang, and Yong Yu.
\newblock Large-scale optimal transport via adversarial training with
  cycle-consistency.
\newblock \emph{arXiv preprint arXiv:2003.06635}, 2020.

\bibitem[Makkuva et~al.(2020)Makkuva, Taghvaei, Oh, and
  Lee]{makkuva2020optimal}
Ashok Makkuva, Amirhossein Taghvaei, Sewoong Oh, and Jason Lee.
\newblock Optimal transport mapping via input convex neural networks.
\newblock In \emph{International Conference on Machine Learning}, pp.\
  6672--6681. PMLR, 2020.

\bibitem[Mallasto et~al.(2019)Mallasto, Frellsen, Boomsma, and
  Feragen]{mallasto2019q}
Anton Mallasto, Jes Frellsen, Wouter Boomsma, and Aasa Feragen.
\newblock (q, p)-wasserstein gans: Comparing ground metrics for wasserstein
  gans.
\newblock \emph{arXiv preprint arXiv:1902.03642}, 2019.

\bibitem[Mao et~al.(2017)Mao, Li, Xie, Lau, Wang, and
  Paul~Smolley]{mao2017least}
Xudong Mao, Qing Li, Haoran Xie, Raymond~YK Lau, Zhen Wang, and Stephen
  Paul~Smolley.
\newblock Least squares generative adversarial networks.
\newblock In \emph{Proceedings of the IEEE international conference on computer
  vision}, pp.\  2794--2802, 2017.

\bibitem[McCann \& Pass(2020)McCann and Pass]{mccann2020optimal}
Robert~J McCann and Brendan Pass.
\newblock Optimal transportation between unequal dimensions.
\newblock \emph{Archive for Rational Mechanics and Analysis}, 238\penalty0
  (3):\penalty0 1475--1520, 2020.

\bibitem[Miyato et~al.(2018)Miyato, Kataoka, Koyama, and
  Yoshida]{miyato2018spectral}
Takeru Miyato, Toshiki Kataoka, Masanori Koyama, and Yuichi Yoshida.
\newblock Spectral normalization for generative adversarial networks.
\newblock In \emph{International Conference on Learning Representations}, 2018.

\bibitem[Nhan~Dam et~al.(2019)Nhan~Dam, Le, Nguyen, Bui, and
  Phung]{nhan2019threeplayer}
Quan~Hoang Nhan~Dam, Trung Le, Tu~Dinh Nguyen, Hung Bui, and Dinh Phung.
\newblock Threeplayer wasserstein gan via amortised duality.
\newblock In \emph{Proc. of the 28th Int. Joint Conf. on Artificial
  Intelligence (IJCAI)}, 2019.

\bibitem[Ostrovski et~al.(2018)Ostrovski, Dabney, and
  Munos]{ostrovski2018autoregressive}
Georg Ostrovski, Will Dabney, and R{\'e}mi Munos.
\newblock Autoregressive quantile networks for generative modeling.
\newblock In \emph{International Conference on Machine Learning}, pp.\
  3936--3945. PMLR, 2018.

\bibitem[Pass(2010)]{pass2010regularity}
Brendan Pass.
\newblock Regularity of optimal transportation between spaces with different
  dimensions.
\newblock \emph{arXiv preprint arXiv:1008.1544}, 2010.

\bibitem[Patrini et~al.(2020)Patrini, van~den Berg, Forre, Carioni, Bhargav,
  Welling, Genewein, and Nielsen]{patrini2020sinkhorn}
Giorgio Patrini, Rianne van~den Berg, Patrick Forre, Marcello Carioni, Samarth
  Bhargav, Max Welling, Tim Genewein, and Frank Nielsen.
\newblock Sinkhorn autoencoders.
\newblock In \emph{Uncertainty in Artificial Intelligence}, pp.\  733--743.
  PMLR, 2020.

\bibitem[Petzka et~al.(2018)Petzka, Fischer, and
  Lukovnikov]{petzka2018regularization}
Henning Petzka, Asja Fischer, and Denis Lukovnikov.
\newblock On the regularization of wasserstein gans.
\newblock In \emph{International Conference on Learning Representations}, 2018.

\bibitem[Radford et~al.(2016)Radford, Metz, and
  Chintala]{radford2016unsupervised}
Alec Radford, Luke Metz, and Soumith Chintala.
\newblock Unsupervised representation learning with deep convolutional
  generative adversarial networks.
\newblock In \emph{4th International Conference on Learning Representations,
  {ICLR} 2016, San Juan, Puerto Rico, May 2-4, 2016, Conference Track
  Proceedings}, 2016.
\newblock URL \url{http://arxiv.org/abs/1511.06434}.

\bibitem[Rockafellar(1976)]{rockafellar1976integral}
R~Tyrrell Rockafellar.
\newblock Integral functionals, normal integrands and measurable selections.
\newblock In \emph{Nonlinear operators and the calculus of variations}, pp.\
  157--207. Springer, 1976.

\bibitem[Salimans et~al.(2016)Salimans, Goodfellow, Zaremba, Cheung, Radford,
  and Chen]{salimans2016improved}
Tim Salimans, Ian Goodfellow, Wojciech Zaremba, Vicki Cheung, Alec Radford, and
  Xi~Chen.
\newblock Improved techniques for training gans.
\newblock \emph{Advances in neural information processing systems},
  29:\penalty0 2234--2242, 2016.

\bibitem[Sanjabi et~al.(2018)Sanjabi, Razaviyayn, Ba, and
  Lee]{sanjabi2018convergence}
Maziar Sanjabi, Meisam Razaviyayn, Jimmy Ba, and Jason~D Lee.
\newblock On the convergence and robustness of training gans with regularized
  optimal transport.
\newblock \emph{Advances in Neural Information Processing Systems},
  2018:\penalty0 7091--7101, 2018.

\bibitem[Santambrogio(2015)]{santambrogio2015optimal}
Filippo Santambrogio.
\newblock Optimal transport for applied mathematicians.
\newblock \emph{Birk{\"a}user, NY}, 55\penalty0 (58-63):\penalty0 94, 2015.

\bibitem[Seguy et~al.(2018)Seguy, Damodaran, Flamary, Courty, Rolet, and
  Blondel]{seguy2018large}
Vivien Seguy, Bharath~Bhushan Damodaran, Remi Flamary, Nicolas Courty, Antoine
  Rolet, and Mathieu Blondel.
\newblock Large scale optimal transport and mapping estimation.
\newblock In \emph{International Conference on Learning Representations}, 2018.

\bibitem[Song \& Ermon(2019)Song and Ermon]{song2019generative}
Yang Song and Stefano Ermon.
\newblock Generative modeling by estimating gradients of the data distribution.
\newblock In \emph{Proceedings of the 33rd Annual Conference on Neural
  Information Processing Systems}, 2019.

\bibitem[Szegedy et~al.(2016)Szegedy, Vanhoucke, Ioffe, Shlens, and
  Wojna]{szegedy2016rethinking}
Christian Szegedy, Vincent Vanhoucke, Sergey Ioffe, Jon Shlens, and Zbigniew
  Wojna.
\newblock Rethinking the inception architecture for computer vision.
\newblock In \emph{Proceedings of the IEEE conference on computer vision and
  pattern recognition}, pp.\  2818--2826, 2016.

\bibitem[Taghvaei \& Jalali(2019)Taghvaei and Jalali]{taghvaei20192}
Amirhossein Taghvaei and Amin Jalali.
\newblock 2-wasserstein approximation via restricted convex potentials with
  application to improved training for gans.
\newblock \emph{arXiv preprint arXiv:1902.07197}, 2019.

\bibitem[Tanielian \& Biau(2021)Tanielian and Biau]{tanielian2021approximating}
Ugo Tanielian and Gerard Biau.
\newblock Approximating lipschitz continuous functions with groupsort neural
  networks.
\newblock In \emph{International Conference on Artificial Intelligence and
  Statistics}, pp.\  442--450. PMLR, 2021.

\bibitem[Tolstikhin et~al.(2017)Tolstikhin, Bousquet, Gelly, and
  Schoelkopf]{tolstikhin2017wasserstein}
Ilya Tolstikhin, Olivier Bousquet, Sylvain Gelly, and Bernhard Schoelkopf.
\newblock Wasserstein auto-encoders.
\newblock \emph{arXiv preprint arXiv:1711.01558}, 2017.

\bibitem[Vahdat \& Kautz(2020)Vahdat and Kautz]{vahdat2020nvae}
Arash Vahdat and Jan Kautz.
\newblock Nvae: A deep hierarchical variational autoencoder.
\newblock In \emph{Advances in Neural Information Processing Systems
  Conference}, 2020.

\bibitem[Villani(2008)]{villani2008optimal}
C{\'e}dric Villani.
\newblock \emph{Optimal transport: old and new}, volume 338.
\newblock Springer Science \& Business Media, 2008.

\bibitem[Xie et~al.(2019)Xie, Chen, Jiang, Zhao, and Zha]{xie2019scalable}
Yujia Xie, Minshuo Chen, Haoming Jiang, Tuo Zhao, and Hongyuan Zha.
\newblock On scalable and efficient computation of large scale optimal
  transport.
\newblock In \emph{International Conference on Machine Learning}, pp.\
  6882--6892. PMLR, 2019.

\bibitem[Zhu et~al.(2017)Zhu, Park, Isola, and Efros]{CycleGAN2017}
Jun-Yan Zhu, Taesung Park, Phillip Isola, and Alexei~A Efros.
\newblock Unpaired image-to-image translation using cycle-consistent
  adversarial networks.
\newblock In \emph{Computer Vision (ICCV), 2017 IEEE International Conference
  on}, 2017.

\end{thebibliography}
	\bibliographystyle{iclr2022_conference}
	
	\newpage
	\appendix
	\section{Proofs}
	\label{sec-proofs}
	\subsection{Proof of Equivalence: Equation \eqref{opt-x} and \eqref{opt-t}}
	\label{prf_equiv}
	\begin{proof}
	Pick any $T:\mathcal{X}\rightarrow\mathcal{Y}$. For every point $x\in\mathcal{X}$ by the definition of the supremum we have 
    $$\langle x, T(x) \rangle - \psi\left (T(x )\right)   \leq \sup_{y\in\mathcal{Y}}\left\lbrace\langle x, y \rangle - \psi(y)\right\rbrace.$$
    Integrating the expression w.r.t. $x\sim\mu$ yields
    $$\int_\mathcal{X}\lbrace\langle x, T(x) \rangle - \psi\left (T(x )\right)\rbrace d\mu(x)   \leq \int_\mathcal{X}\sup_{y\in\mathcal{Y}}\left\lbrace\langle x, y \rangle - \psi(y)\right\rbrace d\mu(x)=\mathcal{L}_{1}.$$
    Since the inequality holds for all $T:\mathcal{X}\rightarrow\mathcal{Y}$, we conclude that
    $$\mathcal{L}_{2}=\sup_{T:\mathcal{X}\rightarrow\mathcal{Y}}\int_\mathcal{X}\lbrace\langle x, T(x) \rangle - \psi\left (T(x )\right)\rbrace d\mu(x)   \leq \int_\mathcal{X}\sup_{y\in\mathcal{Y}}\left\lbrace\langle x, y \rangle - \psi(y)\right\rbrace d\mu(x)=\mathcal{L}_{1},$$
    i.e. $\mathcal{L}_{2}\leq \mathcal{L}_{1}$. Now let us prove that the sup on the left side actually equals $\mathcal{L}_{1}$. To do this, we need to show that for every $\epsilon>0$ there exists $T^{\epsilon}:\mathcal{X}\rightarrow\mathcal{Y}$ satisfying
    $$\int_\mathcal{X}\lbrace\langle x, T^{\epsilon}(x) \rangle - \psi\left (T^{\epsilon}(x)\right)\rbrace d\mu(x)\geq \mathcal{L}_{1}-\epsilon.$$
    First note that for every $x\in\mathcal{X}$ by the definition of the supremum there exists $y^{\epsilon}=y^{\epsilon}(x)$ which provides
    $$\langle x, y^{\epsilon}(x) \rangle - \psi\left (y^{\epsilon}(x)\right)   \geq \sup_{y\in\mathcal{Y}}\left\lbrace\langle x, y \rangle - \psi(y)\right\rbrace-\epsilon.$$
    We take $T^{\epsilon}(x)=y^{\epsilon}(x)$ for all $x\in\mathcal{X}$ and integrate the previous inequality w.r.t. $x\sim\mathcal{\mu}$. We obtain
    $$\int_\mathcal{X}\lbrace{\langle x, T^{\epsilon}(x) \rangle - \psi\left (T^{\epsilon}(x)\right)\rbrace}d\mu(x)   \geq \int_\mathcal{X}\sup_{y\in\mathcal{Y}}\left\lbrace\langle x, y \rangle - \psi(y)\right\rbrace d\mu(x) -\epsilon=\mathcal{L}_{1}-\epsilon,$$
    which is the desired inequality. 
    \end{proof}

	\subsection{Proof of Lemma \ref{lemma-eq}}
	\label{prf_lm41}

	\begin{proof}
	It is enough to prove that $\overline{\psi^{*}}(x)=\langle T^{*}(x),x\rangle-\psi^{*}\big(T(x)\big)$ holds $\mu$-almost everywhere, i.e., $T^{*}(x)\in\argsup\limits_{y\in\mathbb{R}^{D}}\left\lbrace \langle x,y\rangle-\psi^{*}(y)\right\rbrace$. Since ${\nu=T^{*}_{\#}\mu}$, we use \eqref{opt-double-psi-inf} with $\psi\leftarrow \psi^{*}$ to derive
\begin{eqnarray}
\mathcal{W}_{2}^{2}(\mu,\nu)-\int_{\mathcal{X}}\frac{1}{2}\|x\|^{2}d\mu(x)-\int_{\mathcal{Y}}\frac{1}{2}\|y\|^{2}d\nu(y)=
\nonumber
\\
-\int_{\mathcal{X}}\overline{\psi^{*}}(x) d\mu(x) - \int_{\mathcal{Y}} \psi^{*}(y) d\nu(y)=-\int_{\mathcal{X}}\overline{\psi^{*}}(x) d\mu(x) - \int_{\mathcal{Y}} \psi^{*}\big(T^{*}(x)\big) d\mu(x)=
\nonumber
\\
-\int_{\mathcal{X}}\big[\underbrace{\overline{\psi^{*}}(x) + \psi^{*}\big(T^{*}(x)}_{\geq\langle T^{*}(x),x\rangle}\big)\big] d\mu(x)\leq  -\int_{\mathcal{X}} \langle T^{*}(x),x\rangle d\mu(x)=
\label{conj-kant}
\\
\int_{\mathcal{X}} \frac{1}{2}\|x-T^{*}(x)\|^{2} d\mu(x)-\int_{\mathcal{X}}\frac{1}{2}\|x\|^{2}d\mu(x)-\int_{\mathcal{X}}\frac{1}{2}\|T^{*}(x)\|^{2}d\mu(x)=
\nonumber
\\
\mathcal{W}_{2}^{2}(\mu,\nu)-\int_{\mathcal{X}}\frac{1}{2}\|x\|^{2}d\mu(x)-\int_{\mathcal{Y}}\frac{1}{2}\|y\|^{2}d\nu(y).
\nonumber
\end{eqnarray}
As a~result, inequality \eqref{conj-kant} becomes the equality, in particular, $\overline{\psi^{*}}(x) + \psi^{*}\big(T^{*}(x)\big)=\langle T^{*}(x),x\rangle$ holds $\mu$-almost everywhere.
	


\end{proof}

\subsection{Proof of Lemma 4.2}
\label{prf_lm42}
\begin{proof}Let $Q\mathcal{W}_{2}^{2}$ denote the $Q$-embedded quadratic cost. We use the change of variables formula to derive
\begin{eqnarray}
Q\mathcal{W}_{2}^{2}(\mu,\nu)=\inf_{\pi \in \Pi\left(\mu, \nu\right)} \int_{\mathcal{X}\times\mathcal{Y}} \frac{1}{2}\|Q(x)-y\|^{2} d\pi(x,y)=
\nonumber
\\
\inf_{\pi' \in \Pi\left(Q_{\#}\mu, \nu\right)} \int_{\mathcal{X}\times\mathcal{Y}} \frac{1}{2}\|x-y\|^{2} d\pi'(x,y)=\mathcal{W}_{2}^{2}(Q_{\#}\mu, \nu),
\end{eqnarray}
i.e., computing the OT plan for $Q\mathcal{W}_{2}^{2}(\mu,\nu)$ boils down to computing the OT plan for $\mathcal{W}_{2}^{2}(Q_{\#}\mu, \nu)$. 
It follows that $[\text{id}_{\mathbb{R}^{H}},T^{*}\big(Q(x)\big)]_{\#}\mu=[\text{id}_{\mathbb{R}^{H}},G^{*}]_{\#}\mu$ is an OT plan for $Q\mathcal{W}_{2}^{2}(\mu,\nu)$, and $G^{*}$ is the OT map. Inclusion \eqref{g-in-argsup} now  follows from Lemma \ref{lemma-eq}.\end{proof}

\subsection{Proof of Theorem 4.3}
\label{prf_thm43}
\begin{proof}
Pick any $G'\in\argsup_{G}\mathcal{L}(\hat{\psi},G)=\argsup_{G}\int_{\mathcal{X}}\left\lbrace\langle Q(x),G(x)\rangle-\hat{\psi}\big(G(x)\big)\right\rbrace d\mu(x)$ or, equivalently, for all $x\in\mathbb{R}^{H}$, $G'(x)\in\argsup_{y}\left\lbrace\langle Q(x), y\rangle-\hat{\psi}(y)\right\rbrace$. Consequently, for all $y\in\mathbb{R}^{D}$
$$\langle Q(x), G'(x)\rangle-\hat{\psi}\big(G'(x)\big)\geq \langle Q(x), y\rangle-\hat{\psi}(y),$$
which after regrouping the terms yields
$$\hat{\psi}(y)\geq \hat{\psi}\big(G'(x)\big)+\langle Q(x), y-G'(x)\rangle.$$
This means that $Q(x)$ is contained in the subgradient $\partial\hat{\psi}$ at $G'(x)$ for a convex $\hat{\psi}$. Since $\hat{\psi}$ is $\beta$-strongly convex, for points $G(x)$, $G'(x)\in\mathbb{R}^{D}$ and $Q(x)\in \partial\hat{\psi}\big(G'(x)\big)$ we derive
$$\hat{\psi}\big(G(x)\big)\geq\hat{\psi}\big(G'(x)\big)+\langle Q(x),G(x)-G'(x)\rangle+\frac{\beta}{2}\|G'(x)-G(x)\|^{2}.$$
Regrouping the terms, this gives 
$$\big[\langle Q(x),G'(x)\rangle-\hat{\psi}\big(G'(x)\big)\big]-\big[\langle Q(x),G(x)\rangle-\hat{\psi}\big(G(x)\big)\big]\geq\frac{\beta}{2}\|G'(x)-G(x)\|^{2}.$$
Integrating w.r.t.\ $x\sim\mu$ yields
\begin{equation}\epsilon_{1}=\mathcal{L}(\hat{\psi},G')-\mathcal{L}(\hat{\psi},G)\geq \beta\int_{\mathcal{X}}\frac{1}{2}\|G'(x)-G(x)\|^{2}d\mu(x)=\frac{\beta}{2}\cdot \|G-G'\|^{2}_{L^{2}(\mu)}.
\label{eps-1-bound}
\end{equation}
Let $G^{*}$ be the OT map from $\mu$ to $\nu$. We use $G^{*}_{\#}\mu=\nu$ to derive
\begin{eqnarray}\mathcal{L}(\hat{\psi},G')= \int_{\mathcal{X}}\left\lbrace\langle Q(x),G'(x)\rangle-\hat{\psi}\big(G'(x)\big)\right\rbrace d\mu(x)+\int_{\mathcal{Y}} \hat{\psi}(y) d\nu(y)=
\nonumber
\\
\int_{\mathcal{X}}\left\lbrace\langle Q(x),G'(x)\rangle-\hat{\psi}\big(G'(x)\big)\right\rbrace d\mu(x)+\int_{\mathcal{X}} \hat{\psi}\big(G^{*}(x)\big) d\mu(x)=
\nonumber
\\
\int_{\mathcal{X}}\left\lbrace\underbrace{\langle Q(x),G'(x)\rangle-\hat{\psi}\big(G'(x)\big)+\hat{\psi}\big(G^{*}(x)\big)}_{\geq \langle Q(x),G^{*}(x)\rangle+\beta\frac{1}{2}\|G'-G^{*}\|^{2}}\right\rbrace d\mu(x)\geq
\nonumber
\\
\int_{\mathcal{X}}\langle Q(x),G^{*}(x)\rangle d\mu(x)+
\beta\int_{\mathcal{X}}\frac{1}{2}\|G'-G^{*}\|^{2}d\mu(x).
\label{bound-1}
\end{eqnarray}
Let $\psi^{*}$ be an optimal potential in \eqref{unequal-dim-G}. Thanks to Lemma \ref{lemma-uneq}, we have
\begin{eqnarray}\inf_{\psi}\sup_{G}\mathcal{L}(\psi,G)=\mathcal{L}(\psi^{*},G^{*})=
\nonumber
\\
\int_{\mathcal{X}}\left\lbrace\langle Q(x),G^{*}(x)\rangle-\psi^{*}\big(G^{*}(x)\big)\right\rbrace d\mu(x)+\int_{\mathcal{Y}} \psi^{*}(y) d\nu(y)=
\nonumber
\\
\int_{\mathcal{X}}\left\lbrace\langle Q(x),G^{*}(x)\rangle-\psi^{*}\big(G^{*}(x)\big)\right\rbrace d\mu(x)+\int_{\mathcal{X}} \psi^{*}\big(G^{*}(x)\big) d\mu(x)=
\nonumber
\\
\int_{\mathcal{X}}\langle Q(x),G^{*}(x)\rangle d\mu(x)
\label{infsup-eq-corr}
\end{eqnarray}
By combining \eqref{bound-1} with \eqref{infsup-eq-corr}, we obtain
\begin{equation}
    \epsilon_{2}=\mathcal{L}(\hat{\psi},G')-\mathcal{L}(\psi^{*},G^{*})\geq \beta\int_{\mathcal{X}}\frac{1}{2}\|G'-G^{*}\|^{2}d\mu(x)=\frac{\beta}{2}\cdot \|G'-G^{*}\|^{2}_{L^{2}(\mu)}
    \label{eps2-bound}
\end{equation}
The right-hand inequality of \eqref{main-bound} follows from the triangle inequality combined with \eqref{eps-1-bound} and \eqref{eps2-bound}. The middle inequality of \eqref{main-bound} follows from \cite[Lemma A.2]{korotin2021wasserstein} and $G^{*}_{\#}\mu=\nu$. 

Now we prove the left-hand inequality of \eqref{main-bound}. Let $\mathcal{I}$ be the feature extractor of the pre-trained InceptionV3 neural networks. FID score between generated (fake) $\hat{G}_{\#}\mu$ and data distribution $\nu$ is
\begin{equation}\text{FID}(\hat{G}_{\#}\mu,\nu)=\text{FD}(\mathcal{I}_{\#}\hat{G}_{\#}\mu,\mathcal{I}_{\#}\nu)\leq 2\cdot \mathcal{W}_{2}^{2}(\mathcal{I}_{\#}\hat{G}_{\#}\mu,\mathcal{I}_{\#}\nu),
\label{fd-upper-bound}
\end{equation}
where $\text{FD}(\cdot,\cdot)$ is the Fréchet distance which lower bounds $2\cdot \mathcal{W}_{2}^{2}$, see \citep{dowson1982frechet}. Finally, from \cite[Lemma A.1]{korotin2021wasserstein} it follows that 
\begin{equation}\mathcal{W}_{2}^{2}(\mathcal{I}_{\#}\hat{G}_{\#}\mu,\mathcal{I}_{\#}\nu)\leq L^{2}\cdot \mathcal{W}_{2}^{2}(\hat{G}_{\#}\mu,\nu).
\label{lip-upper-bound}
\end{equation}
Here $L$ is the Lipschitz constant of $\mathcal{I}$. We combine \eqref{fd-upper-bound} and \eqref{lip-upper-bound} to get the left-hand inequality in~\eqref{main-bound}.
\end{proof}
	
	\section{Experimental Details}
	\label{sec-exp-details}
    We use the PyTorch framework. All the experiments are conducted on 2$\times$V100 GPUs. We compute inception and FID scores with the official implementation from OpenAI\footnote{IS: \url{https://github.com/openai/improved-gan/tree/master/inception_score}
	} and TTUR\footnote{FID: \url{https://github.com/bioinf-jku/TTUR}}. The compared results are taken from the respective papers or publicly available source codes.
	
	\subsection{General Training Details}
    \label{sec-gen-train-details}
	\textbf{MNIST~\citep{lecun1998gradient}.} On MNIST, we use $x\in\mathbb{R}^{192}$ and $y \in \mathbb{R}^{32\times 32}$.  The batch size is $64$, learning rate $2\cdot 10^{-4}$, optimizer Adam \citep{kingma2014adam} with betas $(0,0.9)$, gradient optimality coefficient $\lambda=10$, and the number of training epochs $T=30$. We observe stable training while updating $\psi$ once in multiple $G$ updates, i.e., $k_G=2$ and $k_\psi=1$.
	
	\textbf{CIFAR10~\citep{krizhevsky2009learning}.} We use all 50000 samples while training. The latent vector $x\in\mathbb{R}^{192}$ and $y \in \mathbb{R}^{32\times 32\times 3}$, batch size 64, $\lambda = 10$, 
	$k_G=1$, $k_\psi=1$, $T=1000$, Adam optimizer with betas $\left(0,0.9\right)$, and  learning rate $2\cdot 10^{-4}$ for $G$ and $1\cdot 10^{-3}$ for $\psi$.
	
	\textbf{CelebA~\citep{liu2015deep}.} We use $x\in\mathbb{R}^{192}$ and $ y \in \mathbb{R}^{64\times 64\times 3}$. The images are first cropped at the center with size 140 and then resized to $64\times 64$. We consider all 202599 samples. We use Adam with betas $\left(0,0.9\right)$, $T=200$, $K_G = 2$, $K_\psi=1$ and learning rate $2 \cdot 10^{-4}$.
	
	\textbf{Image restoration.} In the unpaired image restoration experiments, we use Adam optimizer with betas $\left(0,0.9\right)$, $K_G = 5, K_\psi=1, \lambda = 0$, learning rate $1\cdot10^{-4}$ and train for $T=300$ epochs.

    {\color{black}\textbf{CelebA128x128~\citep{liu2015deep}.} On this dataset, we resize the cropped images as in \textbf{CelebA} to $128\times 128$, i.e. $y\in\mathbb{R}^{128\times 128 \times 3}$. Here, $K_G = 5$, $K_\psi=1$, $\lambda = 0.01$, learning rate $1\cdot 10^{-4}$ and betas=$(0.5,0.999)$. The batch size is reduced to 16 so as to fit in the GPU memory.
	

    \textbf{Anime128x128\footnote{Anime: \url{https://www.kaggle.com/reitanaka/alignedanimefaces}}.} This dataset consists of 500000 high resolution images. We resize the cropped images as in \textbf{CelebA} to $128\times 128$, i.e. $y\in\mathbb{R}^{128\times 128 \times 3}$. Here, $K_G = 5$, $K_\psi=1$, $\lambda = 0.01$, learning rate $2\cdot 10^{-4}$, batch size 16, and betas=$(0,0.9)$.

	\textbf{Toy datasets}. The dimension is $D=H=2$, total number of samples is $10000$. We use the batch size $400$, $\lambda=0.1$, $K_\psi = 1$, $K_G = 16$, and $T=100$. The optimizer is Adam with betas $\left(0.5,0.99\right)$ and learning rate $1\cdot 10^{-3}$. We use the following datasets: }Gaussian to mixture of Gaussians\footnote{\url{https://github.com/AmirTag/OT-ICNN}}{\color{black}, two moons ($\verb|sklearn.datasets.make_moons|$), circles ($\verb|sklearn.datasets.make_circles|$), gaussian to S-curve ($\verb|sklearn.datasets.make_s_curve|$), and gaussian to swiss roll ($\verb|sklearn.datasets.make_swiss_roll|$).
	
	\textbf{Wasserstein-2 benchmark (Appendix \ref{sec-benchmark-early}).} The dimension is $D=H=64\times 64\times 3$. We use batch size $64$, $\lambda=0$, $K_{\psi}=1$, $K_{G}=5$, learning rate $10^{-4}$, and Adam optimizer with default betas.
	}


	
\begin{figure*}[h]
		\begin{center}
			\includegraphics[width=0.9\linewidth]{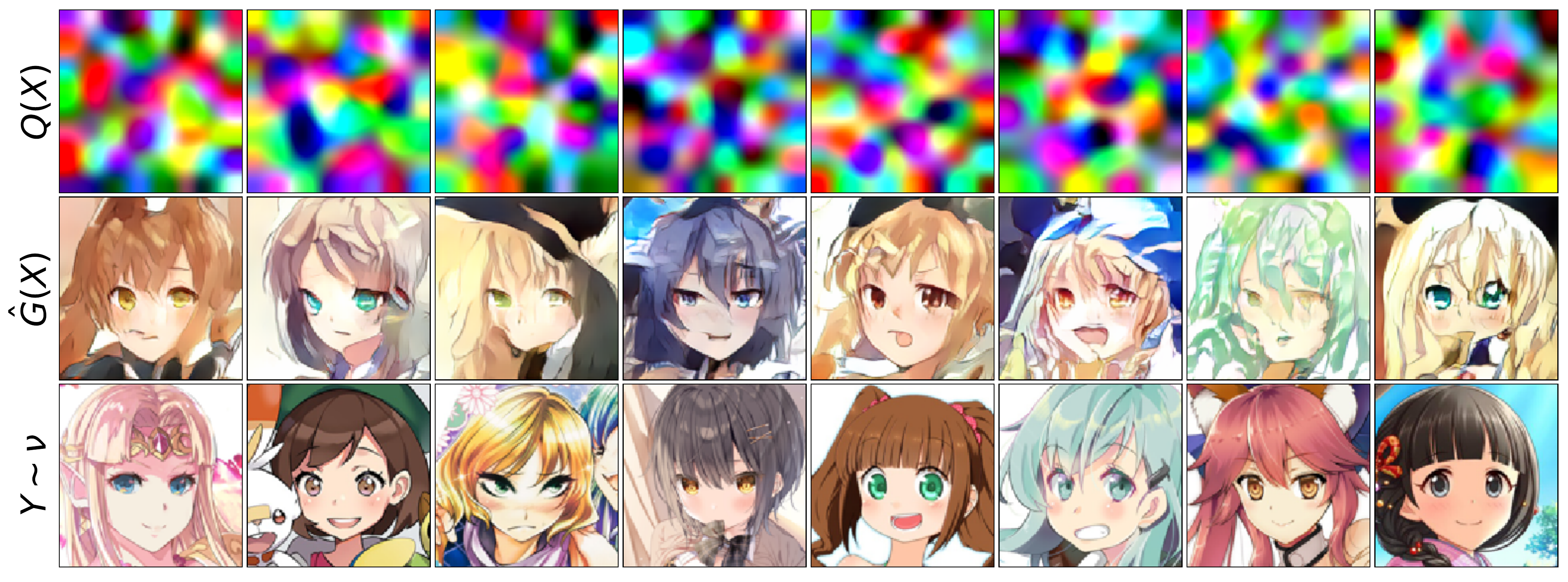}
		\end{center}
		\caption{OTM between 192-dimensional noise and Anime, $128\times 128$. The 1st line shows the \textbf{color} embedding $Q$ (bicubic upscaling of a noise, $3\times 8\times8$), the 2nd line shows corresponding generated samples, and the 3rd line shows random samples from the dataset.
		}
		\label{fig:anime_128x128}
	\end{figure*}

    \rebut{
	\subsection{Evaluation on the Continuous Wasserstein-2 Benchmark}
	\label{sec-benchmark-early}
    To empirically show that the method recovers the optimal transport maps well on equal dimensions, we evaluate it on the recent continuous Wasserstein-2 benchmark by \cite{korotin2021neural}. The benchmark provides a number of artificial test pairs ($\mu,\nu$) of continuous probability distributions with analytically known OT map $T^{*}$ between them.
    
    \begin{wraptable}{r}{4cm} 
    \vspace{-2mm}
    \centering
    \footnotesize
    \vspace{-2mm}\begin{tabular}{|c|c|}
    \hline
    Method & $\mathcal{L}^{2}$-UVP$\downarrow$  \\
    \hline
    $\lfloor \text{MM:R}\rceil$ & $1.4$\%  \\
    \hline
    OTM (ours) & $1.32$\%  \\ \hline
    \end{tabular}
    \caption{\rebut{$\mathcal{L}^{2}$-UVP metric of the recovered transport map on the "Early" images benchmark pair.}}
    \vspace{-7mm}
    \label{table-l2-uvp}
    \end{wraptable}For evaluation, we use the "Early" images benchmark pair (${D=12288}$), see \citep[\wasyparagraph 4.1]{korotin2021neural} for details. We adopt the $\mathcal{L}^{2}$-unexplained percentage metric \cite[\wasyparagraph 5.1]{korotin2021wasserstein} to quantify the recovered OT map $\hat{T}$:
    ${\mathcal{L}^{2}\text{-UVP}(\hat{T})=100\cdot\|\hat{T}-T^{*}\|^{2}/\text{Var}(\nu) \%}$. For our method the $\mathcal{L}_{2}$-UVP metric is only $\approx 1\%$, see Table \ref{table-l2-uvp}. This is comparable to the best $ \lceil \text{MM:R}\rceil$ method which the authors evaluate on their benchmark. The qualitative results are given in Figure \ref{fig:benchmark_equal}.

    \begin{figure}[!h]
	\centering
	\includegraphics[width=0.85\linewidth]{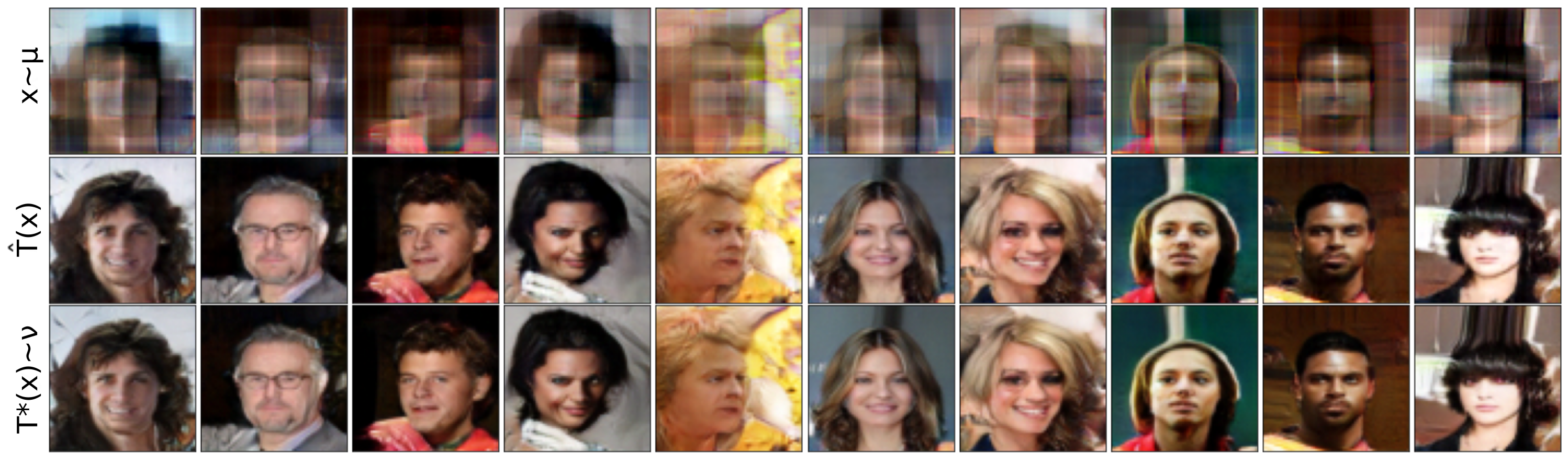}
	\caption{Qualitative results of OTM on the "Early" images  benchmark pair $(\mu,\nu)$ by \cite{korotin2021neural}. The 1st line shows samples $x\sim\mu$, the 2nd line shows fitted OT map $\hat{T}(x)$, and the 3rd line shows the corresponding optimal map $T^{*}(x)\sim \nu$.}
	\label{fig:benchmark_equal}
    \end{figure}

    \subsection{Further Discussion and Evaluation}
    \label{sec-dis-gp}
    \textbf{Generative modeling.} In the experiments, we use the gradient penalty on $\psi$ for better stability of optimization. The penalty is intended to make the gradient norm of the optimal WGAN critic equal to 1 (\citealp[Corollary 1]{gulrajani2017improved}). This condition does not necessarily hold for optimal $\psi^{*}$ in our case and consequently might introduce bias to optimization.
    
    To address this issue, we additionally tested an alternative regularization which we call the \textit{gradient optimality}. For every optimal potential $\psi^{*}$ and map $G^{*}$ of problem \eqref{unequal-dim-G}, we get from Lemma \ref{lemma-uneq}:
    \begin{eqnarray}
     \nabla_G \bigg\lbrace\mathbb{E}_{x\sim \mu} \left [ \langle Q(x), G^{*}(x) \rangle -\psi^{*}\left ( G^{*}(x)) \right ) \right  ]\bigg\rbrace = \mathbb{E}_{x\sim \mu} \left [ Q(x) \right ] - \mathbb{E}_{x\sim \mu} \left [ \nabla \psi^{*}\left ( G^{*}(x)) \right ) \right  ] = 0.
    \label{eq-go}
    \end{eqnarray}
    Since $\mu$ is normal noise distribution and $Q(x)$ is naive upscaling (\wasyparagraph \ref{sec-noise-to-data}), the above expression simplifies to $\mathbb{E}_{x\sim \mu}  \nabla \psi^{*}\left ( G^{*}(x)) \right ) = 0$. Based on this property, we establish the following regularizer $\lambda\|\mathbb{E}_{x\sim \mu}  \nabla \psi\left ( G(x)) \right )\|$ for $\lambda>0$  and add this to $\mathcal{L}_{\psi}$ in our Algorithm~\ref{algo1}.

    While gradient penalty considers expectation of norm, gradient optimality considers norm of expectation. The gradient optimality is always non-negative and vanishes at the optimal point.
    
    
    \begin{wraptable}{r}{4cm} 
    \vspace{-2mm}
    \centering
    \footnotesize
    \vspace{-2mm}\begin{tabular}{|c|c|}
    \hline
    $\lambda$ & FID$\downarrow$  \\
    \hline
    0.001 & 16.91  \\
    0.01 & 16.22 \\
    0.1 & 16.70 \\
    1.0 & 10.01\\
    10 & 6.50 \\ \hline
    \end{tabular}
    \caption{Ablation study of gradient optimality in OTM.}
    \vspace{-2mm}
    \label{table-abl-go}
    \end{wraptable} We conduct additional experiments with the gradient optimality and compare FID scores for different  $\lambda$ in Table \ref{table-abl-go}. It leads to an improvement of FID score from the earlier 7.7 with the gradient penalty to the current 6.5 with the gradient optimality on CelebA (Table \ref{table-celeba-fid}).
     
     \textbf{Unpaired restoration.} In the unpaired restoration experiments (\wasyparagraph \ref{sec-enhancement}), we test OTM with the gradient penalty to make a fair comparison with the baseline WGAN-GP. We find OTM without regularization, i.e., $\lambda = 0$ works better than OTM-GP (Table \ref{tab:fid-restoration}). In practice, more $G$ updates for a single $\psi$ update works fairly well (\wasyparagraph \ref{sec-gen-train-details}).


    \subsection{Additional Qualitative Results}
    \label{sec-addn-exps}
     \rebut{
     OTM works with both the grayscale and color embeddings of noise in the ambient space.

    \textbf{CelebA128x128.} Figure~\ref{fig:celeba_128x128} shows the grayscale embedding $Q(x)$, the recovered transport map $\hat{G}(x)$, and independently drawn real samples $y \sim \nu$. 
   
    \begin{figure*}[h]
		\begin{center}
			\includegraphics[width=0.85\linewidth]{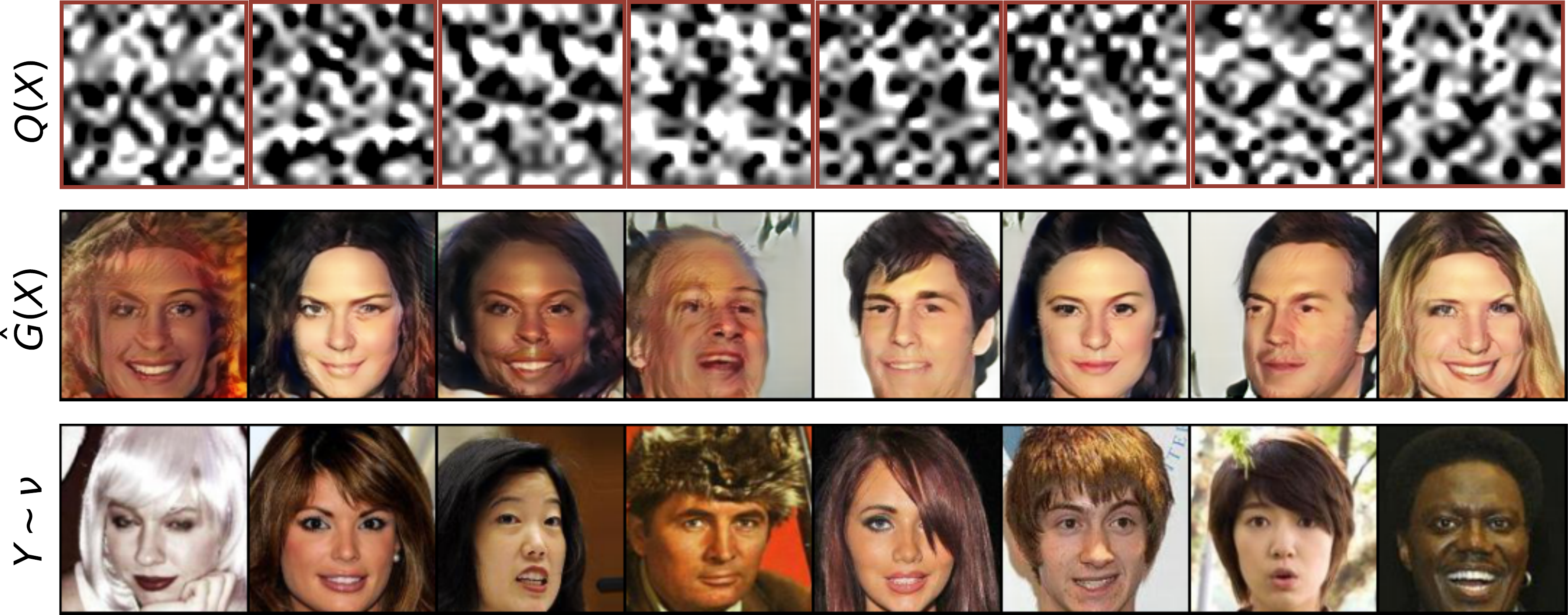}
		\end{center}
		\caption{OTM between 128-dimensional noise and CelebA, $128\times 128$. The 1st line shows the \textbf{grayscale} embedding $Q$ (repeating bicubic upscaling of a noise, $16\times8$), the 2nd line shows corresponding generated samples, and the 3rd line shows random samples from the dataset.}
		\label{fig:celeba_128x128}
	\end{figure*}
    }
    
   
   
}   
    \textbf{Anime128x128.} Figure~\ref{fig:anime_128x128} shows the color embedding $Q(x)$, the recovered transport map $\hat{G}(x)$, and independently drawn real samples $y \sim \nu$. 
       
\begin{figure*}[!h]
	\vspace{-1mm}
		\begin{subfigure}{0.33\textwidth}
			\centering
			\includegraphics[width=0.99\textwidth]{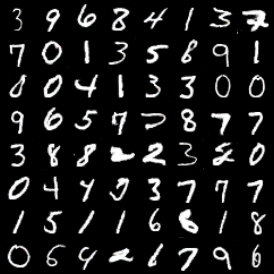}
			\caption{MNIST, $32\times 32$, grayscale}
			\label{mnist_extnd}
		\end{subfigure}  
		\begin{subfigure}{0.33\textwidth}
			\centering
			\includegraphics[width=0.99\textwidth]{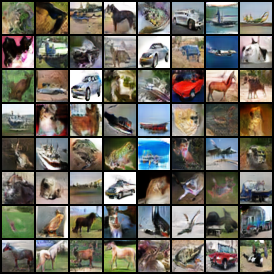}
			\caption{CIFAR10, $32\times 32$, RGB}
			\label{cifar10_extnd}
		\end{subfigure}
		\begin{subfigure}{0.33\textwidth}
			\centering
			\includegraphics[width=0.99\textwidth]{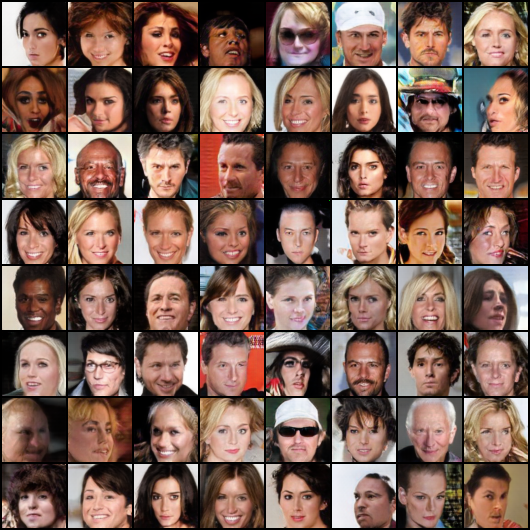}
			\caption{CelebA, $64\times 64$, RGB}
			\label{celeba_extnd}
		\end{subfigure}
	\vspace{-1mm}
		\caption{Randomly generated MNIST, CIFAR10, and CelebA samples by our method (OTM). }
		\label{fig:generated-samples-extnd}
	\end{figure*}

\begin{figure*}[!h]
		\begin{center}
			\includegraphics[width=0.9\linewidth]{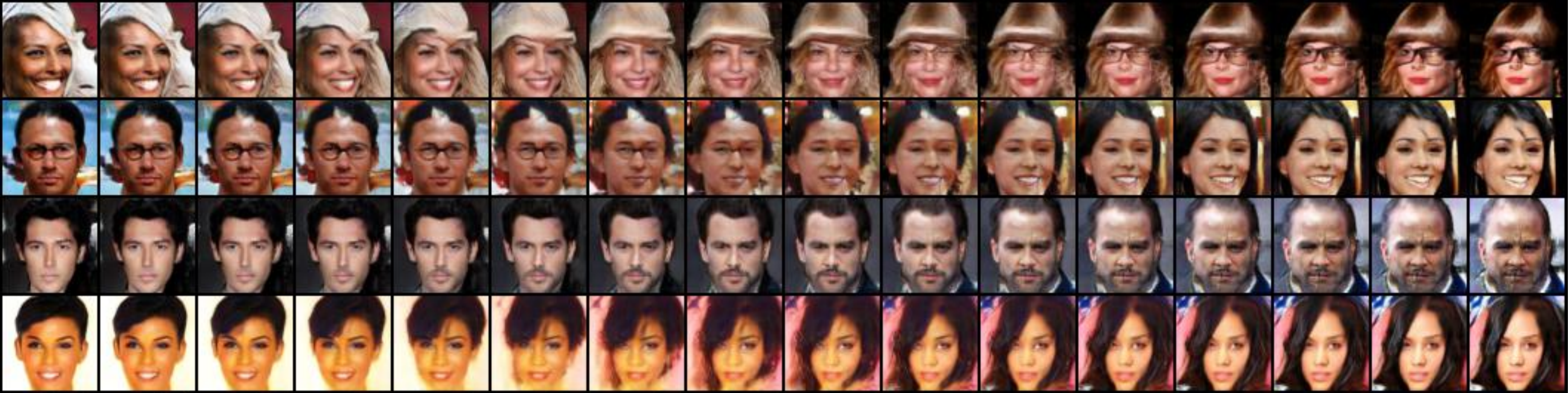}
		\end{center}
		\caption{OTM for latent space interpolation on CelebA, $64\times 64$. Extended samples. }
		\label{celeba_interp_extnd}
	\end{figure*}

\begin{figure}[!h]
    \centering
	\captionsetup[subfigure]{justification=centering}
	\begin{subfigure}{0.38\textwidth}
		\centering
		\includegraphics[width=0.99\textwidth]{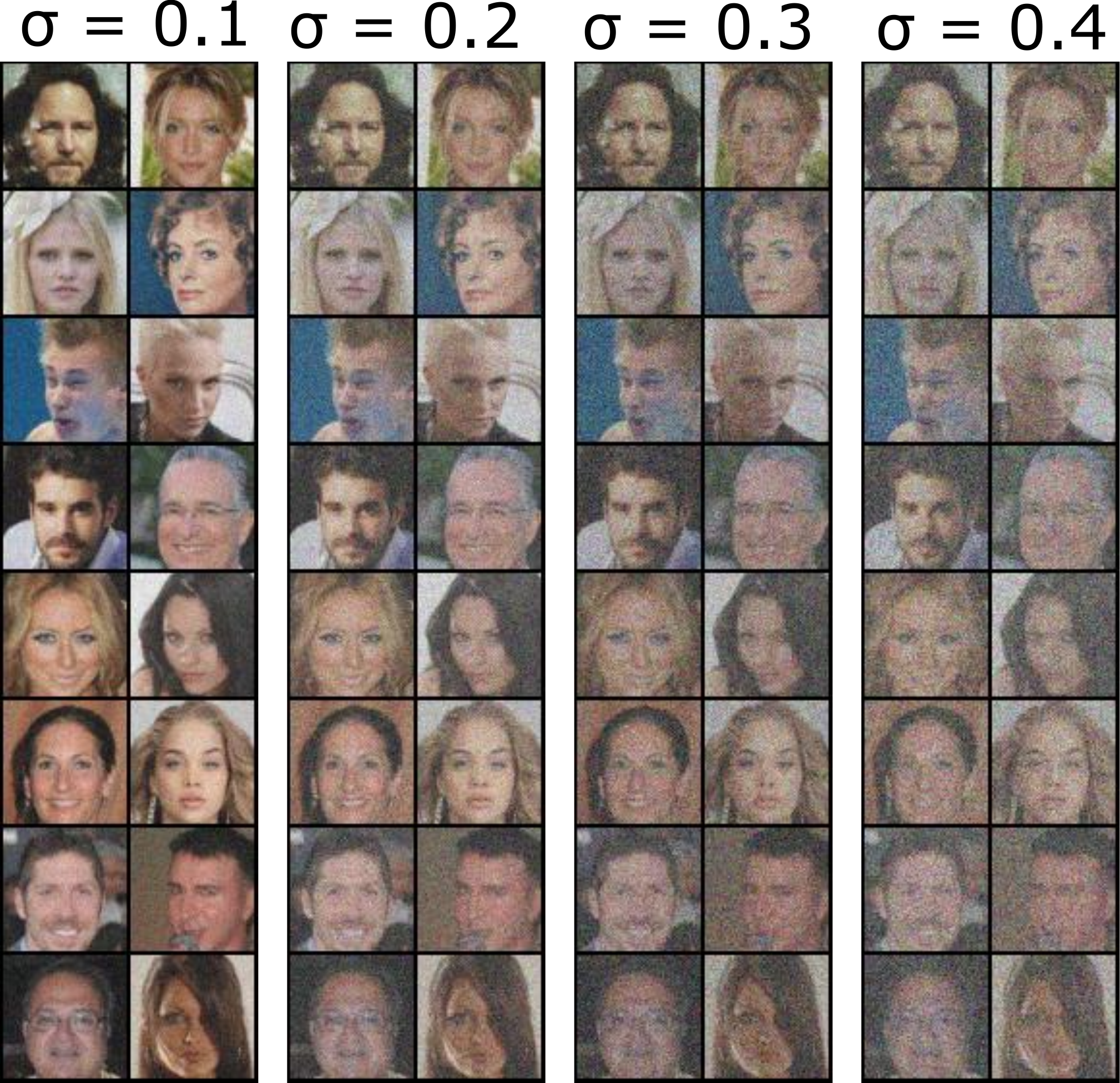}
		\caption{Noisy samples. }
		\label{denoise_sigmas_ip}
	\end{subfigure}
	\hspace{2mm}
	\begin{subfigure}{0.38\textwidth}
		\centering
		\includegraphics[width=0.99\textwidth]{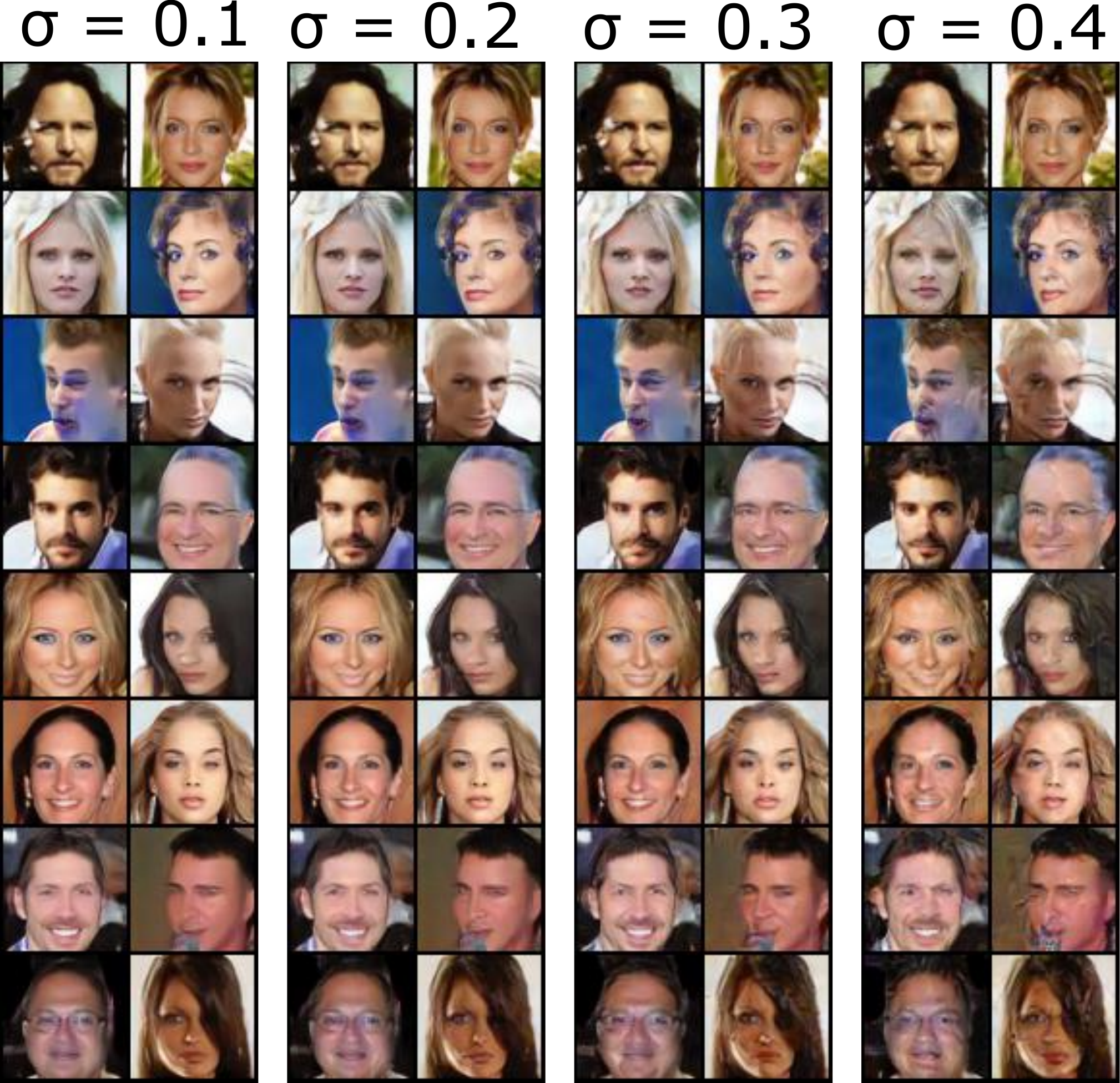}
		\caption{Pushforward samples. }
		\label{denoise_sigmas_op}
	\end{subfigure}
	\hspace{2mm}
	\begin{subfigure}{0.09\textwidth}
		\centering
		\includegraphics[width=0.99\textwidth]{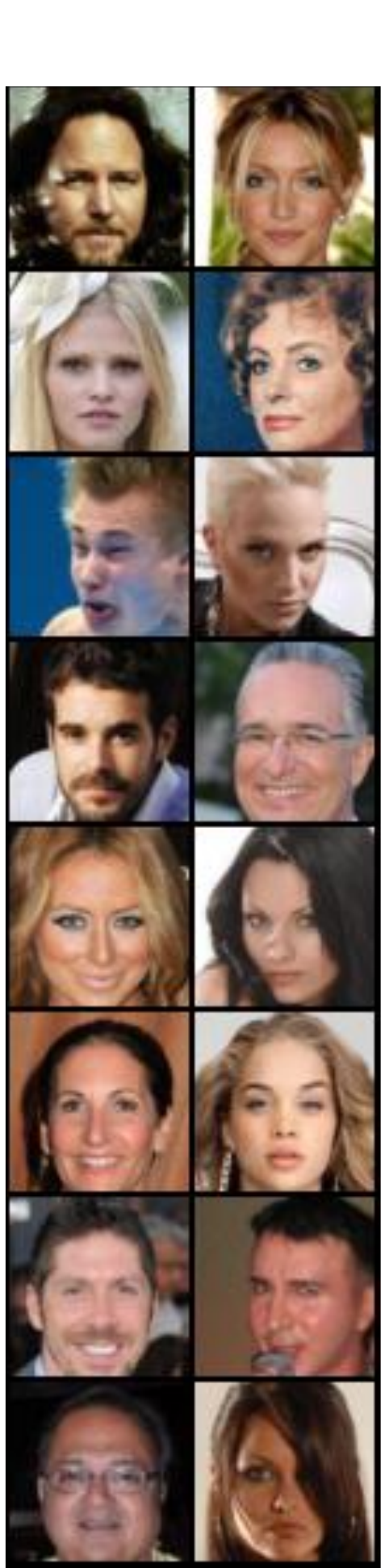}
		\caption{Clean. }
		\label{denoise_sigmas_org}
	\end{subfigure} 
	\caption{OTM for image denoising for varying levels of noise on test C part of CelebA, $64\times 64$.}
	\label{otm_denoise_sigmas}
\end{figure}

	The extended qualitative results with color embedding on \textbf{MNIST}, \textbf{CIFAR10}, and \textbf{CelebA} are shown in Figure~\ref{mnist_extnd}, Figure~\ref{cifar10_extnd}, and Figure~\ref{celeba_extnd} respectively. Table~\ref{table-mnist-fid} shows quantiative results on  MNIST. The color embedding $Q$ is bicubic upscaling of a noise in $\mathbb{R}^{3\times 8\times 8}$. The samples are generated randomly (uncurated) by fitted optimal transport maps between noise and ambient space, e.g., spaces of high-dimensional images. Figure~\ref{celeba_interp_extnd} illustrates latent space interpolation between the generated samples.
    Figure~\ref{otm_denoise_sigmas} shows denoising of images with varying levels of $\sigma = 0.1,0.2,0.3,0.4$ by the model trained with $\sigma=0.3$.

   \begin{table}[!t]
		\centering
		\scriptsize
		\parbox{.45\linewidth}{
			\caption{Results on MNIST dataset.}
			\label{table-mnist-fid}
			\begin{tabular}{llll}
				{\bf Model} & {\bf Related Work} & {\bf FID $\downarrow$  } \\  \hline
				VAE & \cite{kingma2013auto} & 23.8$\pm$0.6 \\
				LSGAN &\cite{mao2017least}   &  7.8$\pm$0.6 \\
				BEGAN   &  \cite{berthelot2017began}    &  13.1$\pm$1.0     \\ \hline
				WGAN & \cite{arjovsky2017wasserstein} &  6.7$\pm$0.4   \\
				SIG & \cite{dai2021sliced} &  4.5   \\\hline
				AE-OT  & \cite{An2020AE-OT:} &   6.2$\pm$0.2  \\
				AE-OT-GAN & \cite{an2020ae} &  3.2  \\ \hline
				OTM   & Ours    &  2.4 \\
			\end{tabular}}
	\end{table}


    \rebut{
    \textbf{Toy datasets.} Figure~\ref{fig:sota-mog} shows the results of our method and related approaches (\wasyparagraph \ref{rel_work}) on a toy 2D dataset. \rebut{Figure \ref{fig:otm_toy} shows the results of our method applied to other toy datasets.}

	\begin{figure*}[!t]
	\captionsetup[subfigure]{justification=centering}
		\begin{subfigure}{0.33\textwidth}
			\centering
			\includegraphics[width=0.99\textwidth]{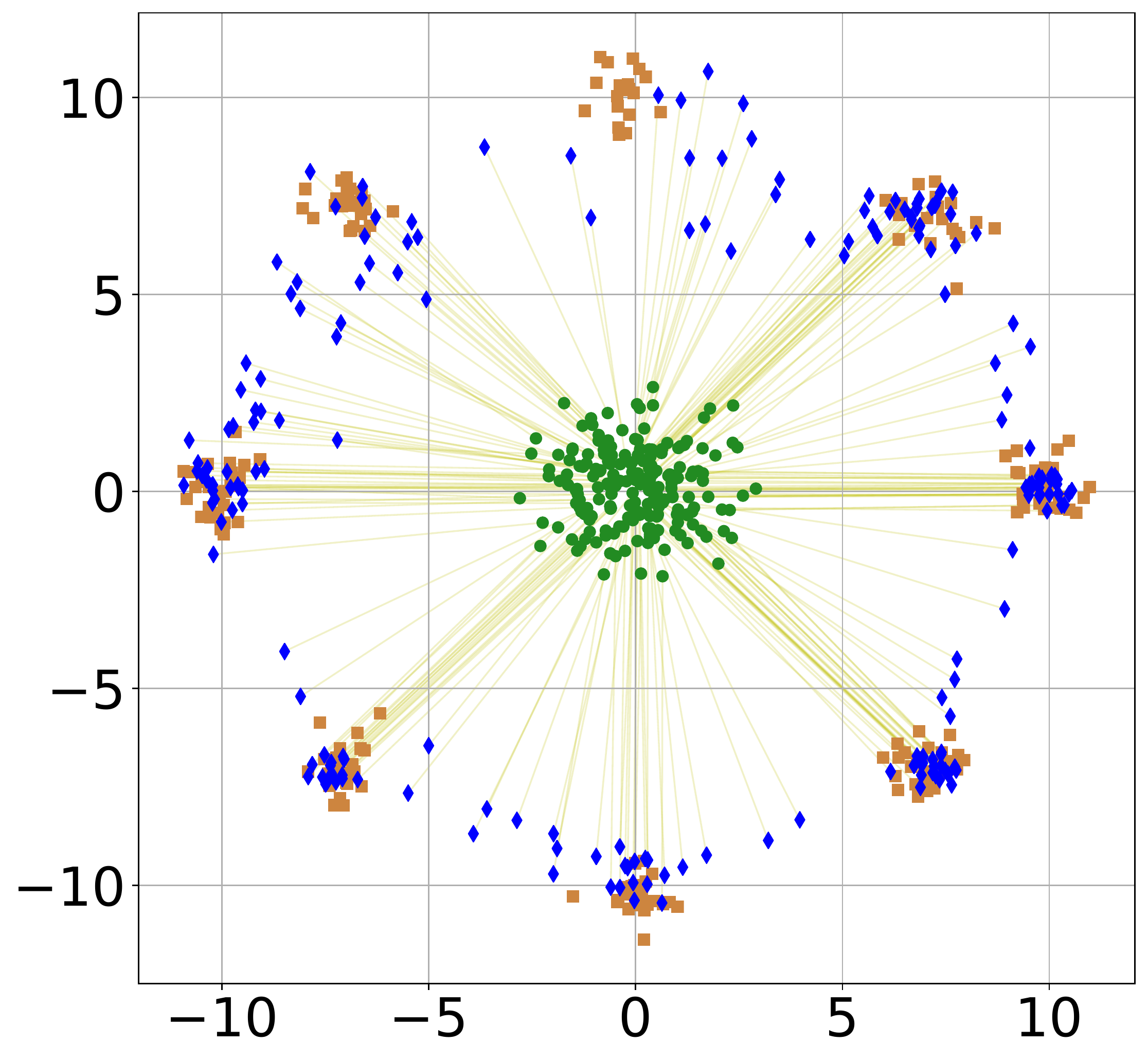}
			\caption{LSOT~\citep{seguy2018large}}
			\label{sota_comp_bary}
		\end{subfigure}
        \begin{subfigure}{0.33\textwidth}
			\centering
			\includegraphics[width=0.99\textwidth]{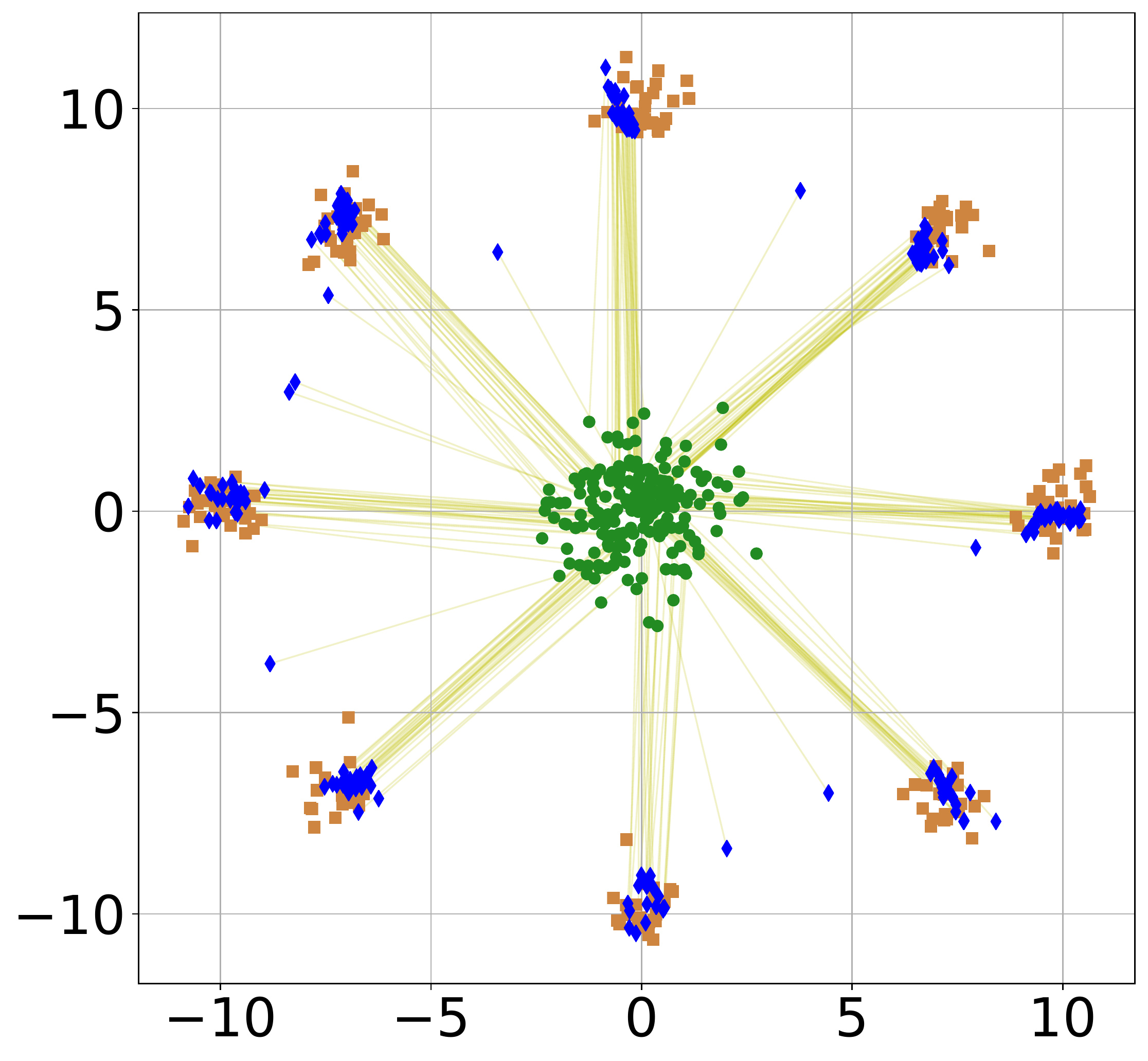}
			\caption{W2GAN~\citep{jacob2018w2gan}}
			\label{sota_comp_w2}
		\end{subfigure}
		\begin{subfigure}{0.33\textwidth}
			\centering
			\includegraphics[width=0.99\textwidth]{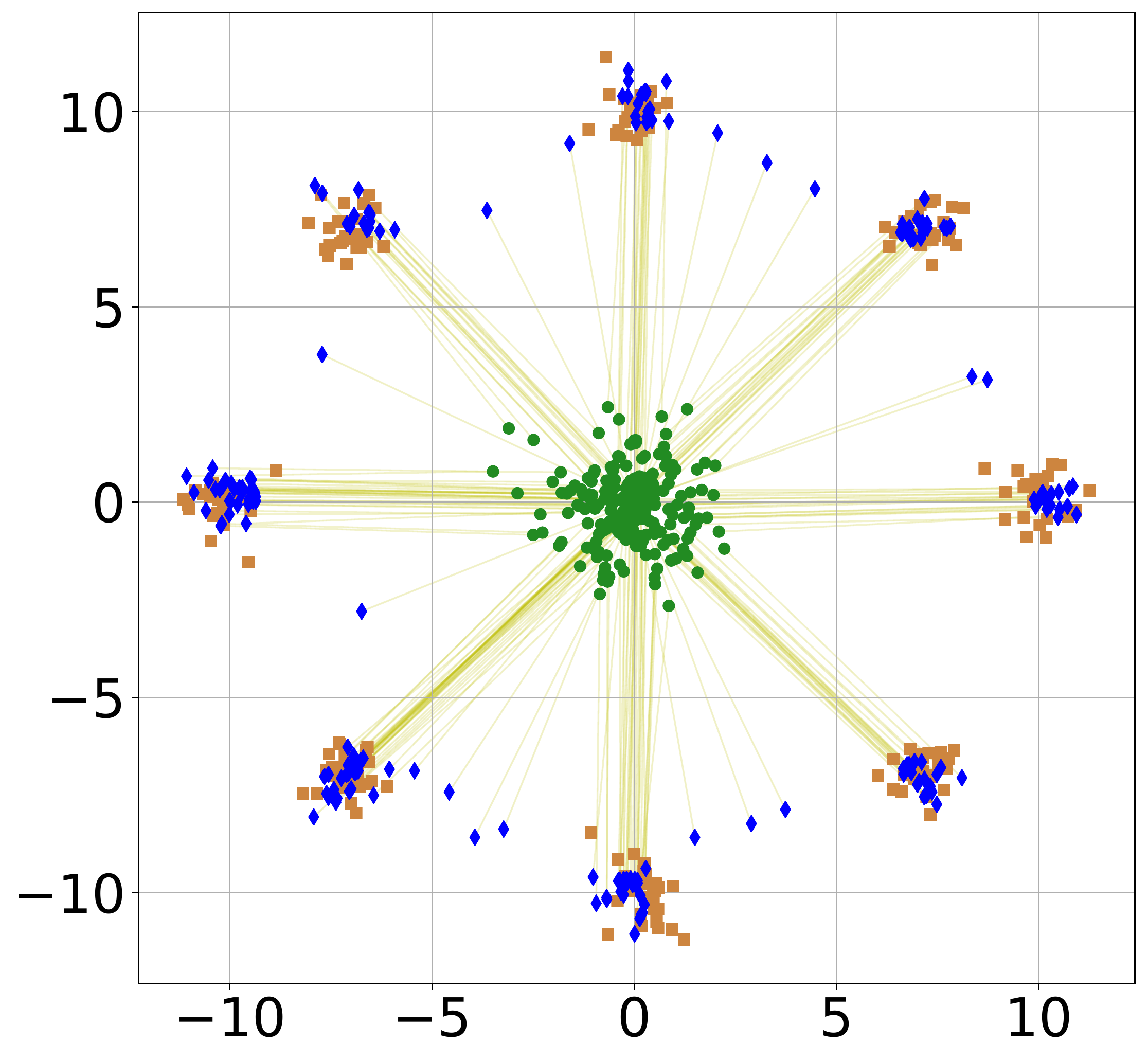}
			\caption{WGAN-LP~\citep{petzka2018regularization}}
			\label{sota_comp_w1}
		\end{subfigure}
		
		\vspace{2mm}
		\begin{subfigure}{0.33\textwidth}
			\centering
			\includegraphics[width=0.99\textwidth]{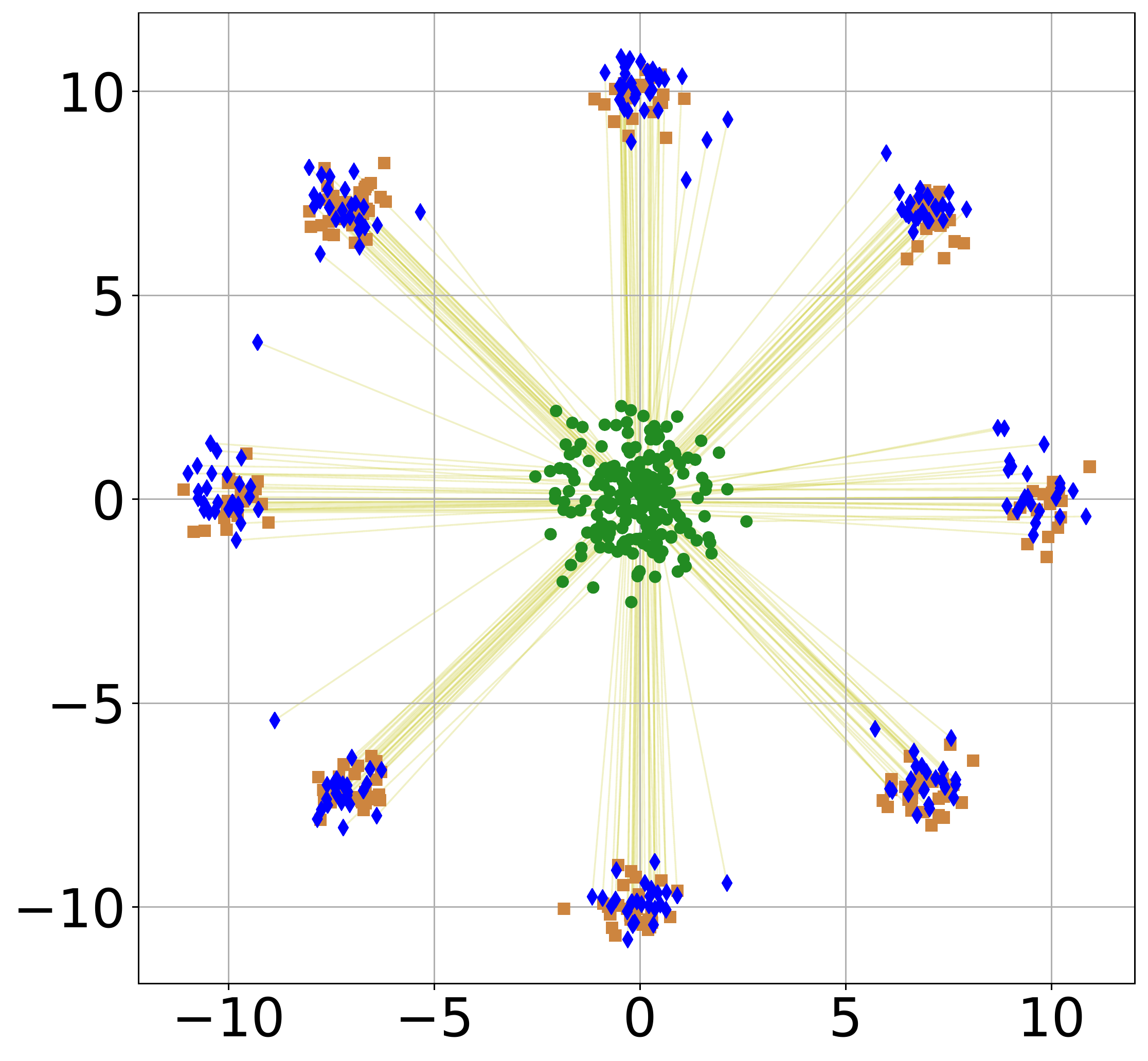}
			\caption{ICNN-OT\citep{makkuva2020optimal}}
			\label{sota_comp_icnn}
		\end{subfigure}
		\begin{subfigure}{0.33\textwidth}
			\centering
			\includegraphics[width=0.99\textwidth]{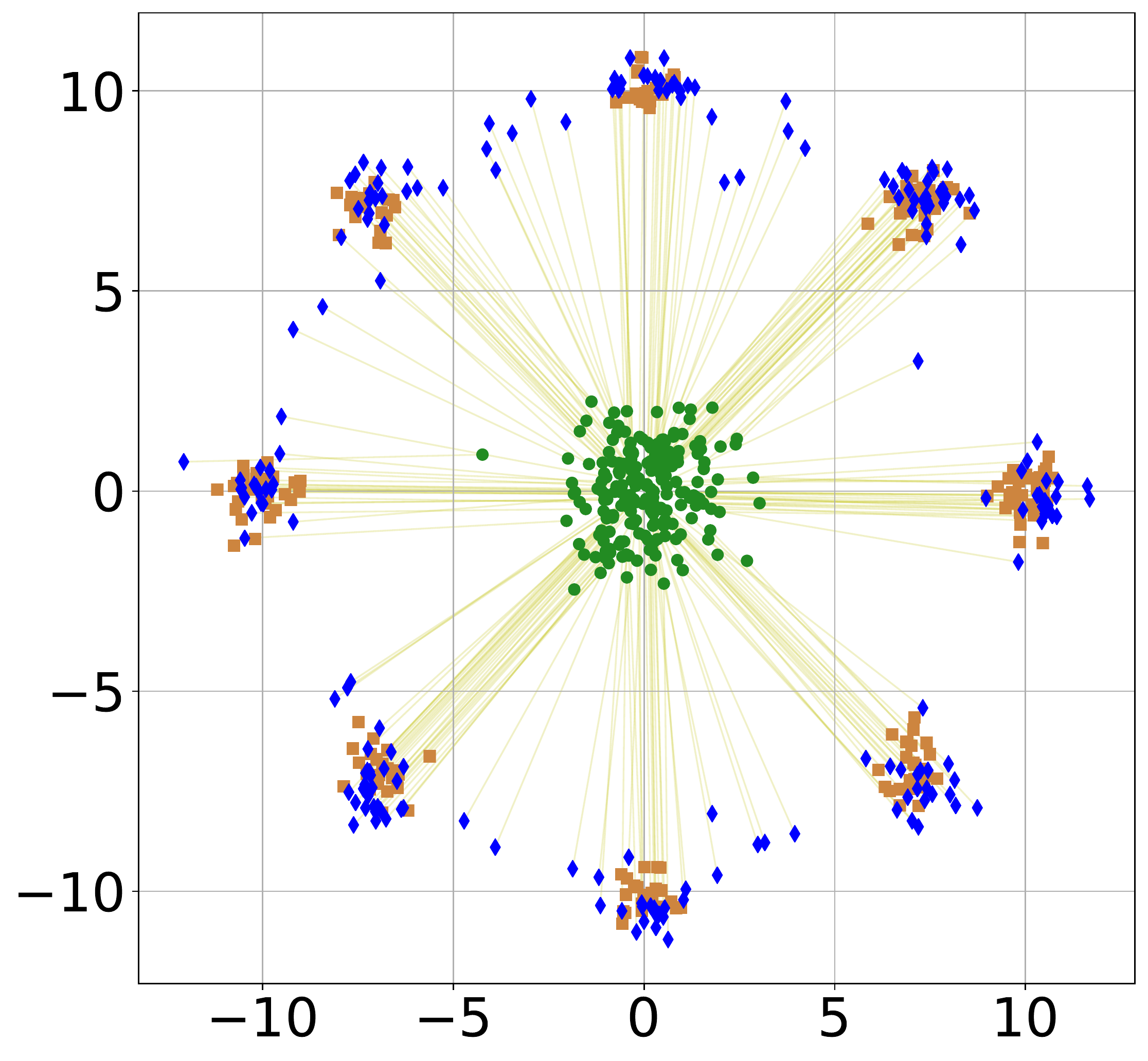}
			\caption{W2GN \citep{korotin2021wasserstein}}
			\label{sota_comp_w2gn}
		\end{subfigure} 
		\begin{subfigure}{0.33\textwidth}
			\centering
			\includegraphics[width=0.99\textwidth]{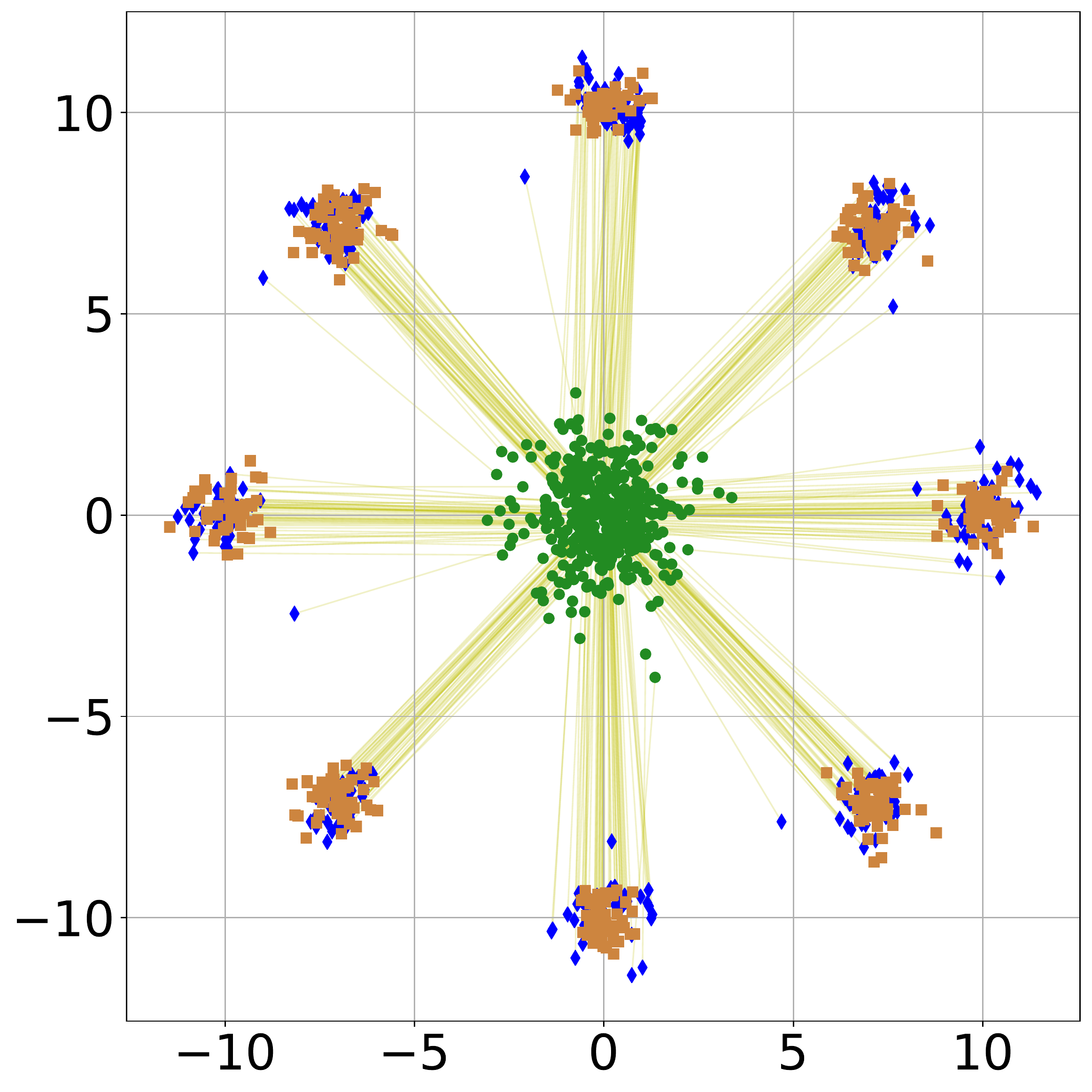}
			\caption{OTM (ours)}
			\label{sota_comp_otm}
		\end{subfigure} 
		\caption{Mapping between a Gaussian and a Mixture of 8 Gaussians in 2D by various methods. The colors green, blue, and peru represent input, pushforward, and output samples respectively.}
		\label{fig:sota-mog}
	\end{figure*}

	


      \begin{figure*}[!t]
			\centering
		\begin{subfigure}{0.37\textwidth}
			\centering
			\includegraphics[width=0.99\textwidth,height=0.99\textwidth]{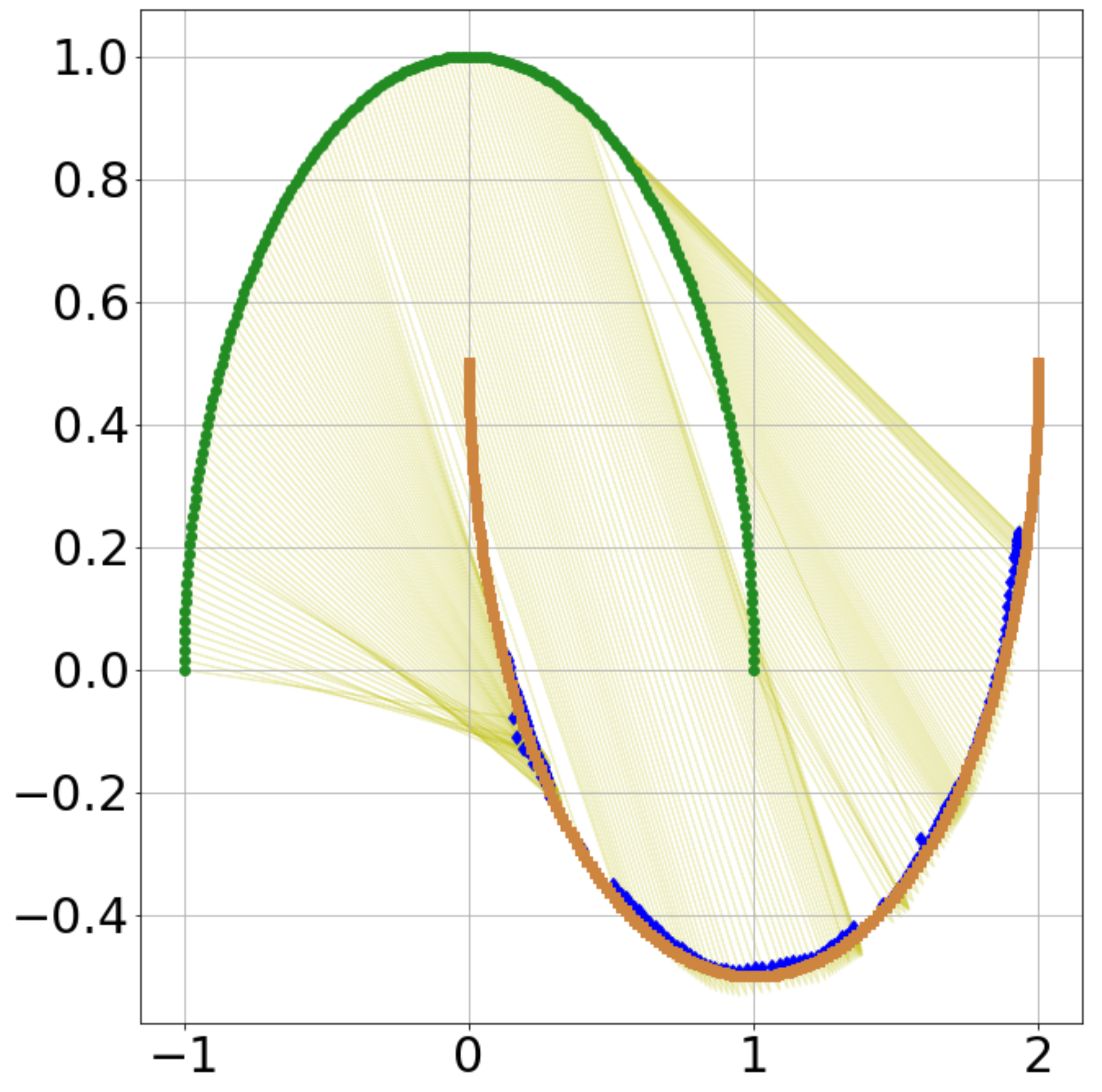}
			\caption{Two moons}
			\label{fig:otm_moon_trial1}
		\end{subfigure}  
		\begin{subfigure}{0.37\textwidth}
			\centering
			\includegraphics[width=0.99\textwidth]{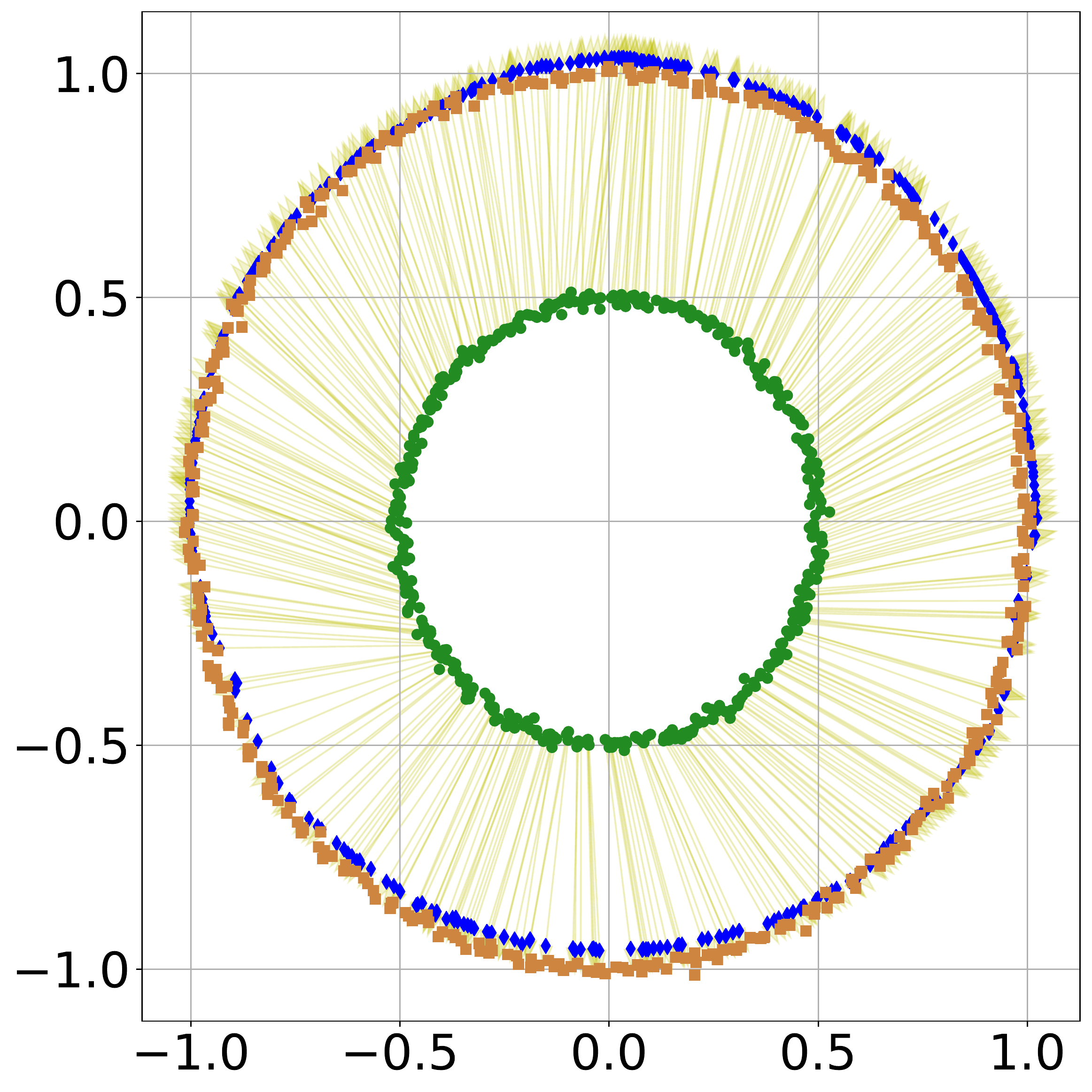}
			\caption{Circles}
			\label{fig:otm_circ}
		\end{subfigure}     
		\vspace{2mm}
		
		\begin{subfigure}{0.37\textwidth}
			\centering
			\includegraphics[width=0.99\textwidth,height=0.99\textwidth]{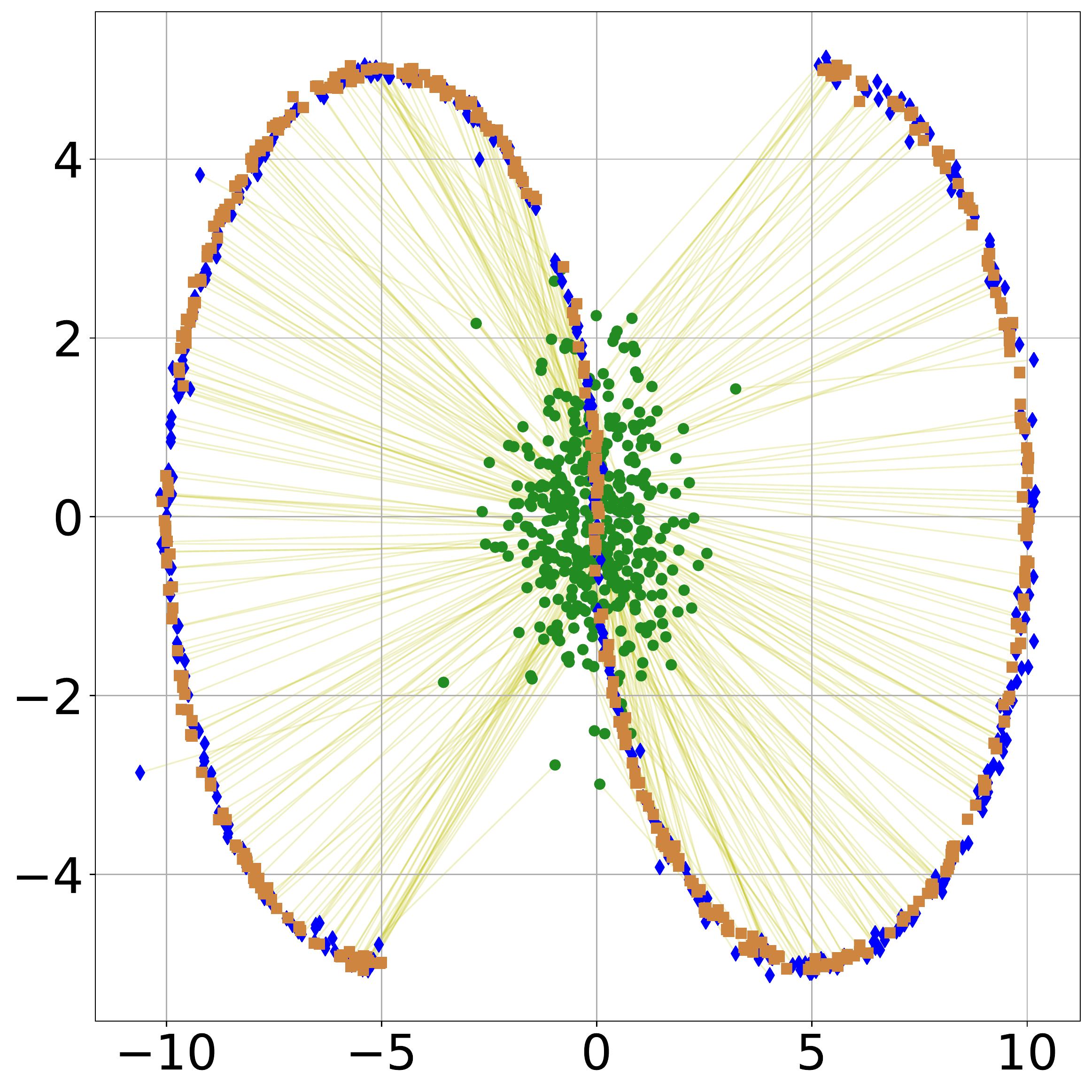}
			\caption{S Curve}
			\label{fig:otm_scurve}
		\end{subfigure}   
		\begin{subfigure}{0.37\textwidth}
			\centering
			\includegraphics[width=0.99\textwidth,height=0.99\textwidth]{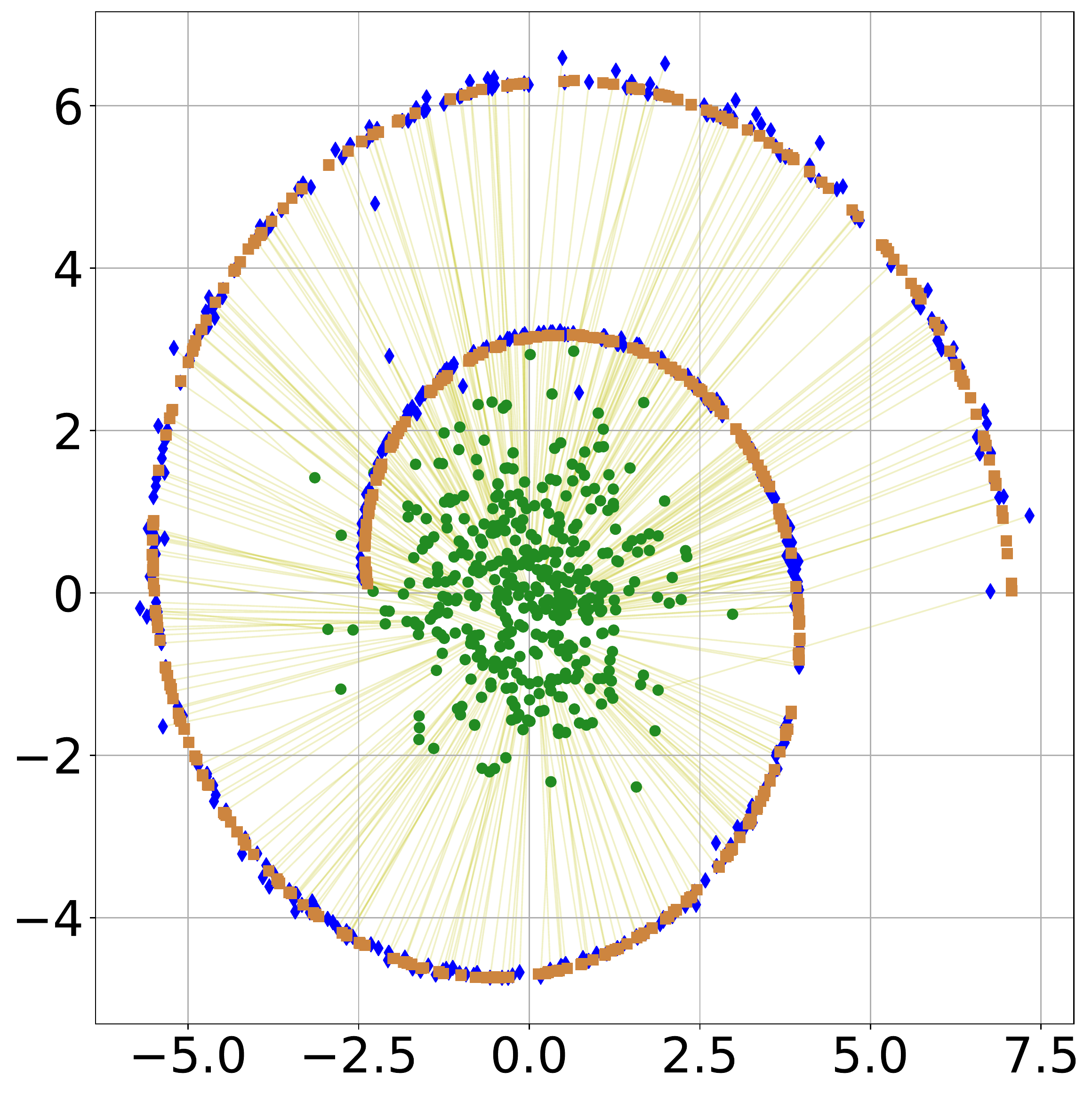}
			\caption{Swiss Roll}
			\label{fig:otm_swirl}
		\end{subfigure}  
		\caption{OTM on toy datasets, $D=2$. Here, the colors green, blue, and peru represent input, pushforward, and output samples respectively.}
		\label{fig:otm_toy}
	\end{figure*}

    }


	\subsection{Neural Network Architectures}
	\label{sec-network-arch}

    This section contains architectures on \textbf{CIFAR10} (Table~\ref{cifar_resnet}), \textbf{CelebA} (Table~\ref{celeba_conv_arch}), \rebut{\textbf{CelebA $128\times 128$} (Figure \ref{fig:celeba_128x128})} generation, \textbf{image restoration} tasks (Table~\ref{celeba_conv}), evaluation on the toy 2D datasets and \rebut{\textbf{Wasserstein-2 images benchmark} (Table \ref{table-l2-uvp})}.
    
    
    In the unpaired restoration tasks (\wasyparagraph\ref{sec-enhancement}), we use UNet architecture for transport map $G$ and convolutional architecture for potential $\psi$. 
    Similarly to \citep{song2019generative}, we use BatchNormaliation (BN) and InstanceNormalization+ (INorm+) layers. In the ResNet architectures, we use the ResidualBlock of NCSN~\citep{song2019generative}.

    In the toy 2D examples, we use a simple multi-layer perceptron with 3 hidden layers consisting of 128 neurons each and LeakyReLU activation. The final layer is linear without any activation.
    
    \rebut{The transport map $G$ and potential $\psi$ architectures on MNIST $32 \times 32$, CelebA $128\times 128$, and Anime $128 \times 128$ are the generator and discriminator architectures of WGAN-QC \cite{Liu_2019_ICCV} respectively.}

    \rebut{In the evaluation on the Wasserstein-2 benchmark, we use publicly available Unet\footnote{https://github.com/milesial/Pytorch-UNet} architecture for transport map $T$ and WGAN-QC discriminator's architecture for $\psi$ \citep{Liu_2019_ICCV}. These neural network architectures are the same as the authors of the benchmark use.
    }


	\begin{table}[t]
		\centering
		\caption{Architectures for generation task on CIFAR10, $32\times 32$.}
		\label{cifar_resnet}
		\begin{tabular}{c}
			\hline
			$G(.)$    \\ \hline
			Noise: $x\in \mathbb{R}^{128} $\\
			Linear, Reshape, output shape: $[128\times 4\times 4]$ \\
			ResidualBlock Up, output shape: $[128\times 8\times 8]$ \\ 
			ResidualBlock Up, output shape: $[128\times 16\times 16]$ \\ 
			ResidualBlock Up, output shape: $[128\times 32\times 32]$ \\ 
			Conv, Tanh, output shape: $[3\times 32\times 32]$          \\     \\
			\hline
			$\psi(.)$ \\
			\hline
			Target: $y\in \mathbb{R}^{3\times 32 \times 32} $ \\
			ResidualBlock Down, output shape: $[128\times 16\times 16]$ \\ 
			ResidualBlock Down, output shape: $[128\times 8\times 8]$ \\ 
			ResidualBlock, output shape: $[128\times 8\times 8]$ \\
			ResidualBlock, output shape: $[128\times 8\times 8]$ \\
			ReLU, Global sum pooling, output shape: $[128\times 1\times 1]$     \\
			Linear, output shape: $[1]$ \\  
		\end{tabular}
	\end{table}
	
		\begin{table}[t]
		\centering
		\caption{Architectures for generation task on Celeba, $64\times 64$.}
		\label{celeba_conv_arch}
		\begin{tabular}{c}
			\hline
			$G(.)$    \\ \hline
			Noise: $x\in \mathbb{R}^{128}$\\
			ConvTranspose, BN, LeakyReLU, output shape: $[256\times 1\times 1]$ \\ 
			ConvTranspose, BN, LeakyReLU, output shape: $[512\times 4\times 4]$  \\ 
			Conv, PixelShuffle, BN, LeakyReLU, output shape: $[512\times 8\times 8]$   \\
			Conv, PixelShuffle, BN, LeakyReLU, output shape: $[512\times 16\times 16]$ \\ 
			Conv, PixelShuffle, BN, LeakyReLU, output shape: $[512\times 32\times 32]$ \\ 
			ConvTranspose, Tanh, output shape: $[3\times 64\times 64]$         \\ \\
			\hline
			$\psi(.)$ \\
			\hline
			Target: $y\in \mathbb{R}^{3\times 64 \times 64} $ \\
			Conv, output shape: $[128\times 64\times 64]$ \\
			ResidualBlock Down, output shape: $[256\times 32\times 32]$ \\
			ResidualBlock Down, output shape: $[256\times 16\times 16]$ \\
			ResidualBlock Down, output shape: $[256\times 8\times 8]$ \\
			ResidualBlock Down, output shape: $[128\times 4\times 4]$ \\
			Conv, output shape: $[1]$ \\                           
		\end{tabular}
	\end{table}

		\begin{table}[t]
		\centering
		\caption{Architectures for restoration tasks on CelebA, $64\times 64$.}
		\label{celeba_conv}
		\begin{tabular}{c}
			\hline
			$G(.)$    \\ \hline
			Input: $x\in \mathbb{R}^{3\times 64 \times 64} $\\
			Conv, BN, LeakyReLU, output shape: $[256 \times 64\times 64]$ \\ 
			Conv, LeakyReLU, AvgPool, output shape: $[256 \times 32\times 32]$, x1  \\ 
			Conv, LeakyReLU, AvgPool, output shape: $[256 \times 16\times 16]$, x2  \\ 
			Conv, LeakyReLU, AvgPool, output shape: $[256 \times 8\times 8]$, x3  \\ 
			Conv, LeakyReLU, AvgPool, output shape: $[256 \times 4\times 4]$, x4  \\ 
			\hline
            Nearest Neighbour Upsample, Conv, BN, ReLU, output shape: $[256\times 8 \times 8]$, y3 \\
            Add (y3, x3), output shape: $[256\times 8 \times 8]$, y3 \\ 
			Nearest Neighbour Upsample, Conv, BN, ReLU, output shape: $[256\times 16 \times 16]$, y2 \\
            Add (y2, x2), output shape: $[256\times 16 \times 16]$, y2 \\ 
            Nearest Neighbour Upsample, Conv, BN, ReLU, output shape: $[256\times 32 \times 32]$, y1 \\
            Add (y1, x1), output shape: $[256\times 32 \times 32]$, y1 \\
            Nearest Neighbour Upsample, Conv, BN, ReLU, output shape: $[256\times 64 \times 64]$, y \\
            Add (y, x), output shape: $[256\times 64 \times 64]$, y \\ 
			ConvTranspose, Tanh, output shape: $[3\times 64\times 64]$         \\ \\
			\hline
			$\psi(.)$ \\
			\hline
			Target: $y\in \mathbb{R}^{3\times 64 \times 64}$ \\
			Conv, LeakyReLU, AvgPool, output shape: $[256 \times 32\times 32]$  \\ 
			Conv, LeakyReLU, AvgPool, output shape: $[256 \times 16\times 16]$  \\ 
			Conv, LeakyReLU, AvgPool, output shape: $[256 \times 8\times 8]$  \\ 
			Conv, LeakyReLU, AvgPool, output shape: $[256 \times 4\times 4]$  \\ 
			Linear, output shape: $[1]$ \\                           
		\end{tabular}
	\end{table}

\end{document}